\documentclass{article}

\usepackage[preprint]{neurips_2021}

\usepackage[utf8]{inputenc} 
\usepackage[T1]{fontenc}    
\usepackage{hyperref}       
\usepackage{url}            
\usepackage{booktabs}       
\usepackage{amsfonts}       
\usepackage{nicefrac}       
\usepackage{microtype}      
\usepackage{xcolor}         
\usepackage{multirow}
\usepackage{subcaption,graphicx}
\usepackage[inline]{enumitem}

\usepackage{amsmath,amssymb,amsfonts,dsfont,amsthm}
\newtheorem{theorem}{Theorem}
\newtheorem{definition}{Definition}
\newtheorem{lemma}{Lemma}
\newtheorem{remark}{Remark}
\newtheorem{corollary}{Corollary}
\usepackage[nohints]{minitoc}

\usepackage{graphicx}
\usepackage{algorithm}
\usepackage{algorithmic}
\usepackage{amssymb}

\newcommand\argmax{\mathop{\rm arg\,max}}

\newcommand \domain {\ensuremath{\mathcal{X}}}
\newcommand \tree {\ensuremath{\mathcal{T}}}

\newtheorem{assumption}{Assumption}

\newcommand \reg {\mathrm{Reg}}
\newcommand{\ball}{\ensuremath{\mathcal{B}}}

\newcommand \prob {\mathbb{P}}
\newcommand \expect {\mathbb{E}}
\newcommand \real {\mathbb{R}}

\DeclareMathAlphabet{\mathpzc}{OT1}{pzc}{m}{it}

\if 1

\else

\fi

\makeatletter
\newtheorem*{rep@theorem}{\rep@title}
\newcommand{\newreptheorem}[2]{%
	\newenvironment{rep#1}[1]{%
		\def\rep@title{\textbf{#2} \ref{##1}}%
		\begin{rep@theorem}}%
		{\end{rep@theorem}}}
\makeatother
\newreptheorem{theorem}{Theorem}
\newreptheorem{lemma}{Lemma}
\newreptheorem{proposition}{Proposition}
\newreptheorem{assumption}{Assumption}
\newreptheorem{corollary}{Corollary}

\newcommand{\setof}[1]{\{#1\}}

\newcommand{\framework}{{\color{blue}\ensuremath{\mathsf{PCTS}}}}
\newcommand{\frameworkucb}{{\color{blue}\ensuremath{\mathsf{PCTS+DUCB1}}}}
\newcommand{\frameworkucbs}{{\color{blue}\ensuremath{\mathsf{PCTS+DUCB1}\sigma}}}
\newcommand{\frameworkucbv}{{\color{blue}\ensuremath{\mathsf{PCTS+DUCBV}}}}
\newcommand{\bandit}{{\color{red}\ensuremath{\mathsf{BANDIT}}}}
\newcommand{\ducb}{{\color{red}\ensuremath{\mathsf{DUCB1}}}}
\newcommand{\ducbs}{{\color{red}\ensuremath{\mathsf{DUCB1}\sigma}}}
\newcommand{\ducbv}{{\color{red}\ensuremath{\mathsf{DUCBV}}}}
\usepackage{mathrsfs}
\DeclareRobustCommand{\bigO}{%
  \text{\usefont{OMS}{cmsy}{m}{n}O}%
}

\newcommand*\mean[1]{\bar{#1}}
\allowdisplaybreaks%
\newif\ifdoublecol
\doublecolfalse

\title{Procrastinated Tree Search: Black-box Optimization with Delayed, Noisy, and Multi-fidelity Feedback}

\author{%
  Junxiong Wang \\
  Dept. of Computer Science\\
    Cornell University\\
  Ithaca, NY, USA 14850\\
   \And
   Debabrota Basu \\
   \'Equipe Scool, Inria\\UMR 9189 - CRIStAL, CNRS\\Univ. Lille, Centrale Lille\\ Lille, France 59000
   \And
   Immanuel Trummer \\
   Dept. of Computer Science\\
   Cornell University\\
  Ithaca, NY, USA 14850\\
}

\begin{document}

\maketitle
\doparttoc 
\faketableofcontents 
\begin{abstract}
  In black-box optimization problems, we aim to maximize an unknown objective function, where the function is only accessible through feedbacks of an evaluation or simulation oracle. In real-life, the feedbacks of such oracles are often noisy and available after some unknown delay that may depend on the computation time of the oracle. Additionally, if the exact evaluations are expensive but coarse approximations are available at a lower cost, the feedbacks can have multi-fidelity. In order to address this problem, we propose a generic extension of hierarchical optimistic tree search (HOO), called ProCrastinated Tree Search (\framework{}), that flexibly accommodates a delay and noise-tolerant bandit algorithm. We provide a generic proof technique to quantify regret of \framework{} under delayed, noisy, and multi-fidelity feedbacks. Specifically, we derive regret bounds of \framework{} enabled with delayed-UCB1 (\ducb) and delayed-UCB-V (\ducbv) algorithms. Given a horizon $T$, \framework{} retains the regret bound of non-delayed HOO for expected delay of $\bigO(\log T)$ and worsens by $\bigO(T^{\frac{1-\alpha}{d+2}})$ for expected delays of $\bigO(T^{1-\alpha})$ for $\alpha \in (0,1]$. We experimentally validate on multiple synthetic functions and hyperparameter tuning problems that \framework{} outperforms the state-of-the-art black-box optimization methods for feedbacks with different noise levels, delays, and fidelity.
\end{abstract}
\section{Introduction}
Black-box optimization~\citep{mcts,sen19_MFHOO}, alternatively known as zeroth-order optimization~\citep{xu2020zeroth} or continuous-arm multi-armed bandit~\citep{bubeck2011a}, is a widely studied problem and has been successfully applied in reinforcement learning~\citep{mcts,grill2020monte}, neural architecture search~\citep{wang2019alphax}, large-scale database tuning~\citep{pavlo2017self,Wang2021}, robotics~\citep{martinez2017bayesian}, AutoML~\citep{fischer2015machines}, material science~\citep{xue2016accelerated,kajita2020autonomous}, and many other domains.
In black-box optimization, we aim to maximize an unknown function $f:\domain \to \real$, i.e. to find\vspace*{-.5em}
\begin{equation}
    x^* \triangleq \argmax_{x\in \domain} f(x).
\end{equation}
In this setting, the optimizer does not have access to the derivatives of $f$, rather can access $f$ only by sequentially querying a simulation or evaluation oracle~\citep{jamieson2012query}.
The goal is to minimize the expected error in optimization, i.e. $\mathbb{E}[f(x^*)-f(x_T)]$, after $T$ queries~\citep{mcts}, or to reach a fixed error threshold with as few queries as possible~\citep{jamieson2012query}.
\ifdoublecol
We adopt the first approach in this paper.
\else
We adopt the first approach of analysis in this paper.
\fi

\textbf{Approaches to Black-box Optimization.} \cite{jamieson2012query} have shown that black-box optimization for convex functions is in general efficient. For convex functions, typically a Zeroth-order (ZO) Gradient Descent (GD) framework is used that replaces the gradient with the difference between functional evaluations~\citep{jamieson2012query,kumagai2017regret,liu2018zeroth}. This approach requires double evaluation queries per-step and also multiple problem-specific hyperparameters to be tuned to obtain reasonable performance. Still, these methods are less robust to noise and stochastic delay~\citep{li2019bandit} than the next two other approaches, i.e.\ Bayesian Optimization (BO) and Hierarchical Tree Search.

\begin{table*}[ht!]
\centering
\caption{Comparison of existing tree search, BO, and zeroth-order GD optimizers.}
\label{tab:related}
\resizebox{\textwidth}{!}{%
\begin{tabular}{l|l|l|l|l|l}
\toprule
Algorithm   & Expected Simple Regret                            & Delay   & Noise & Fidelity  & Assumptions  \\
\midrule
$\framework{}$ & $T^{-1/(d+2)}(\log T+\frac{\expect[Delay]}{\sigma^2 +2b})^{1/(d+2)}$ & Stochastic & Unknown & Yes & Local Lip. \\
HOO~\citep{bubeck2011a}    & $T^{-1/(d+2)}(\log T)^{1/(d+2)}$ & x       & Known & MF-HOO~\citep{sen19_MFHOO}    & Local Lip.    \\
GP-UCB~\citep{gpucb} & ${T}^{-1/2}~\mathrm{InfoGain}(T)$                  & x       & Known & MF-GP-UCB~\citep{mfgpucb} & GP surrogate \\
GP-EI~\citep{gpei,nguyen2019filtering}   & ${T}^{-1/2}O((\log T)^{d/2})$
& x & Known & x         & GP surrogate   \\
DBGD~\citep{li2019bandit}   & $\sqrt{T+\sum {Delay}}/T$                      & Bounded & x     & x         & Convex   \\
\hline
\end{tabular}%
}\vspace*{-1em}
\end{table*}
For an objective function with no known structure except local smoothness, solving the black-box optimization problem is equivalent to estimating $f$ almost everywhere in its domain $\domain$~\citep{goldstein1977optimization}. This can lead to an exponential complexity in the dimensionality of the domain~\citep{chen1988lower,wang2019optimization}. Thus, one approach for this problem is to learn a surrogate $\hat{f}$ of the actual function $f$, such that $\hat{f}$ is a close approximation of $f$ and $\hat{f}$ can be learned and optimized with fewer samples. This has led to research in Bayesian Optimization (BO) and its variants~\citep{gpucb,gpei,sko,mfgpucb}, where specific surrogate regressors are fitted to the Bayesian posterior of $f$. However, if $f$ is highly nonlinear or high dimensional, the Bayesian surrogate, namely Gaussian Process (GP)~\citep{gpucb} or Bayesian Neural Network (BNN)~\citep{NIPS2016_a96d3afe}, requires many samples to fit and generalize well. Also, there are two other issues. Firstly, often myopic acquisition used in BO algorithms leads to excessive exploration of the boundary of the search domain~\citep{oh2018bock}. Secondly, the error bounds of BO algorithms include the information gain term ($\mathrm{InfoGain}(T)$)  that often increases with $T$~\citep{gpucb}.

Instead of fixing on to such specific surrogate modelling, the alternative is to use \textit{hierarchical tree search} methods which have drawn significant attention and success in the recent past\ifdoublecol~\citep{mcts,bubeck2011a,poo_first,Shang2018AdaptiveBO,poo,sen18_mfdoo,sen19_MFHOO}.\else~\citep{mcts,bubeck2011a,10.1145/1374376.1374475,poo_first,Shang2018AdaptiveBO,poo,sen18_mfdoo,sen19_MFHOO}.\fi
The tree search approach explores the space using a hierarchical binary tree with nodes representing subdomains of the function domain $\domain$. Then, it leverages a bandit algorithm to balance the exploration of the domain and fast convergence towards the subdomains with optimal values of $f$. This approach does not demand more than local smoothness assumption with respect to the hierarchical partition~\citep[Assumption 1]{Shang2018AdaptiveBO} and an asymptotically consistent bandit algorithm~\citep{bubeck2011a}. The generic nature of hierarchical tree search motivated us to extend it to black-box optimization with delayed, noisy, and multi-fidelity feedbacks.


\textbf{Imperfect Oracle: Delay, Noise, and Multi-fidelity (DNF).} In real-life, the feedbacks of the evaluation oracle can be received after a delay due to the computation time to complete the simulation or evaluation~\citep{weinberger2002delayed}, or to complete the communication between servers~\citep{agarwal2012distributed,sra2015adadelay}. Such delayed feedback is natural in different optimization problems, including the white-box settings~\citep{Wang2021,li2019bandit,joulani2016delay,langford2009slow}. In some other problems, introducing artificial delays while performing tree search, may create opportunities for work sharing between consecutive evaluations, thereby reducing computation time~\citep{Wang2021}. This motivated us to look into the delayed feedback for black-box optimization.
Additionally, the feedbacks of the oracle can be noisy or even the objective function itself can be noisy, for example simulation oracles for physical processes~\citep{kajita2020autonomous} and evaluation oracles for hyperparameter tuning of classifiers~\citep{sen19_MFHOO} and computer systems~\citep{fischer2015machines,Wang2021}. 
On the other hand, the oracle may invoke a multi-fidelity framework. Specially, if there is a fixed computational or time budget for the optimization, the optimizer may choose to access coarse but cheaper evaluations of $f$ than the exact and costlier evaluations~\citep{sen18_mfdoo,sen19_MFHOO,mfgpucb}.
Both the noisy functions and multi-fidelity frameworks are studied separately in tree search regime while assuming a known upper bound on the noise variance~\citep{sen19_MFHOO} or known range of noise~\citep{xu2020zeroth}.
We propose to extend tree search to a setting where all three imperfections, delay, noise, and multi-fidelity (\textit{DNF}), are encountered concurrently. Additionally, we remove the requirement that the noise is either known or bounded.


\textbf{Our Contributions.} The main contributions of this paper are as follows:

1. \textit{Algorithmic:} We show that the hierarchical tree search (HOO) framework is extendable to delayed, noisy and multi-fidelity (DNF) feedback through deployment of the upper confidence bounds of a bandit algorithm that is immune to the corresponding type of feedback. This reduces the tree search design problem to designing compatible bandit algorithms. In Section~\ref{sec:pcts}, we describe this generic framework, and refer to it as the \textit{Procrastinated Tree Search} (\framework{}).

2. \textit{Theoretical:} We leverage the generalities of the regret analysis of tree search and incorporate delay and noise-tolerant bandit algorithms to show the expected simple regret bounds for expected delay $\tau = O(\log T)$ and $O(T^{1-\alpha})$ for $\alpha \in (0,1)$. We instantiate the analysis for delayed versions of UCB1-$\sigma$~\citep{Auer2002} and UCB-V~\citep{ucbv}. This requires analysing a delayed version of UCB1-$\sigma$ and extending UCB-V to the delayed setting. We show that we have constant loss and $T^{(1-\alpha)/(d+2)}$ loss compared to non-delayed HOO in case of the two delay models (Sec.~\ref{sec:delay}). We also extend the analysis to unknown noise variance (Sec.~\ref{sec:noise}) and multi-fidelity (Sec.~\ref{sec:mf}). To the best of our knowledge, we are the first to consider DNF-feedback in hierarchical tree search, and our regret bound is more general than the existing ones for black-box optimization with either delay or known noise or multi-fidelity (Table~\ref{tab:related}).

3. \textit{Experimental:} We experimentally and comparatively evaluate performance of \framework{} on multiple synthetic and real-world hyperparameter optimization problems against the state-of-the-art black-box optimization algorithms (Sec.~\ref{sec:experiments})\footnote{Link to our code: \url{https://github.com/jxiw/PCTS}}. We evaluate for different delays, noise variances (known and unknown), and fidelities. In all the experiments, \framework{} with delayed-UCB1-$\sigma$ (\ducbs) and delayed-UCB-V (\ducbv) outperform the competing tree search, BO, and zeroth-order GD optimizers.

\section{Background and Problem Formulation}\label{sec:back}
We aim to maximize an objective function $f:\domain \rightarrow \real$, where the domain $\domain \subseteq \real^D$. At each iteration, the algorithm queries $f$ at a chosen point $x_t \in \domain$ and gets back an evaluation $y=f(x_t) + \epsilon$, such that $\expect[\epsilon] = 0$ and $\mathbb{V}[\epsilon]=\sigma^2$~\citep{jamieson2012query}. We consider both, the case where $\sigma^2$ is known and where it is unknown to the algorithm. We denote $x^*$ as the optimum and $f^* \triangleq f(x^*)$ as the optimal value.

\textbf{Structures of the Objective Function.}
In order to prove convergence of \framework{} to the global optimum $f^*$, we need to assume that the domain $\domain$ of $f$ has at least a semi-metric $\ell$ defined on it~\citep{mcts}. This allows us to define an $\ell$-ball of radius $\rho$ with $\ball_\rho \triangleq \{ x| \max_y \ell(x,y)\leq \rho~\forall~x, y \in \ball_\rho \subseteq \domain\}$.
Now, we aim to define the near-optimality dimension of the function $f$, given semi-metric $\ell$. The near-optimality dimension quantifies the inherent complexity of globally optimising a function using tree search type algorithms.
Near-optimality dimension quantifies the $\epsilon$-dependent growth in the number of $\ell$-balls needed to pack this set of $\epsilon$-optimal states: $\domain_\epsilon \triangleq \{ x \in \domain | f(x) \geq f^* - \epsilon\}$.
\begin{definition}[$c$-near-optimality dimension~\citep{bubeck2011a}]\label{def:NOpt_d}
$c$-near-optimality dimension is the smallest $d\geq 0$, such that for all $\epsilon > 0$, the maximal number of disjoint $\ell$-balls of radius $c\epsilon$ whose centers can be accommodated in $\domain_\epsilon$ is $\bigO(\epsilon^{-d})$.
\end{definition}
This is a joint property of $f$ and the dissimilarity measure $\ell$. $d$ is independent of the algorithm of choice and can be defined for any $f$ and $\domain$ with semi-metric $\ell$.
Additionally, we need $f$ to be smooth around the optimum $x^*$, i.e. to be weak Lipschitz continuous, for the tree search to converge.
\begin{assumption}[Weak Lipschitzness of $f$~\citep{bubeck2011a}]~\label{ass:lipschitz}
For all $x, y \in \domain$, $f$ satisfies $f^* - f(y) \leq f^* - f(x) + \max\{ f^* - f(x), \ell(x,y)\}$.
\end{assumption}
Weak Lipschitzness implies that there is no sudden drop or jump in $f$ around the optimum $x^*$.
Weak Lipschitzness can hold even for discontinuous functions.
Thus, it widens applicability of hierarchical tree search methodology and corresponding analysis to more general performance metrics and domain spaces in comparison with algorithms that explicitly need gradients or smoothness in stricter forms.
In Appendix, we show that we can relax this assumption proposed in Hierarchical Optimistic Optimization (HOO)~\cite{bubeck2011a} to more local assumptions like~\citep{Shang2018AdaptiveBO}. As we develop \framework{} using the HOO framework, we keep this assumption here to directly compare the effects of DNF feedbacks.

\textbf{Structure: Non-increasing Hierarchical Partition.} The Hierarchical Tree Search or $\domain$-armed bandit family of algorithms~\citep{bubeck2011a,poo,sen19_MFHOO} grow a tree $\tree \subseteq \cup \{(h,l)\}_{h,l=0,1}^{H, 2^h}$ of depth $H$, such that each node $(h,l)$ represents a subdomain $\domain_{(h,l)}$ of $\domain$,\footnote{Here, $(h,l)$ represents the $l$-th node at depth $h$.} and the corresponding upper confidence intervals partition the domain of the performance metric $f$. Then, it uses a UCB-type bandit algorithm to assign optimistic upper confidence values to each partition. Using these values, it chooses a node to evaluate and expand at every time step.
As the tree grows deeper, we obtain a more granular hierarchical partition of the domain.
As we want the confidence intervals to shrink with increase in their depth, we need to ensure certain regularity of such hierarchical partition. 
Though we state the hierarchical partition as an assumption, it can be considered as an artifact of the tree search algorithm.
\begin{assumption}[Hierarchical Partition with Decreasing Diameter and Shape~\citep{mcts}]\label{ass:partition}\

1. \emph{Decreasing diameters.} There exists a decreasing sequence $\delta(h) > 0$ and constant $\nu_1 > 0$ such that
$\mathrm{diam}(X_{h,l}) \triangleq \max_{x \in X_{h,l}} \ell(x_{h,l}, x) \leq \nu_1 \delta(h)$,
for any depth $h \geq 0$, for any interval $X_{h,l}$, and for all $i=1,\ldots,2^h.$ For simplicity, we consider that $\delta(h) = \rho^h$ for $\rho \in (0,1)$.

2. \emph{Regularity of the intervals.} There exists a constant $\nu_2>0$ such that for any depth $h\geq 0$, every interval $X_{h,l}$ contains at least a ball $\ball_{h,l}$ of radius $\nu_2 \rho^h$ and center $x_{h,l}$ in it. Since the tree creates a partition at any given depth $h$, $\ball_{h,l} \cap \ball_{h,l'} = \emptyset$ for all $1\leq l < l' \leq 2^h$.
\end{assumption}

\textbf{Simple Regret: Performance Metric.} While analyzing iterative or sequential algorithms, regret $\reg_T \triangleq \sum_{t=1}^T[f(x^*)-f(x_t)]$ is widely used as the performance measure~\citep{mcts}. For optimization algorithms, another relevant performance metric is expected error or expected simple regret incurred at time $T$: 
$\epsilon_T= \expect[r_T] = \expect[f(x^*) - f(x_T)] = \frac{1}{T}\mathbb{E}[\reg_T]$.
Since the last equality holds for tree search~\citep{mcts}, we state only the expected simple regret results in the main paper.
The algorithm performance is better if the expected simple regret is lower.
If the upper bound on expected simple regret grows sublinearly with horizon $T$, the corresponding algorithm asymptotically converges to the optimum. 
Given the aforementioned assumptions and definitions, and choosing simple regret as the performance measure, we state the expected error bound of HOO~\citep[Theorem 6]{bubeck2011a} (using UCB1).
\begin{theorem}[Regret of HOO]~\label{thm:HOO}
Assume that the expected objective function $f$ satisfies Assumption~\ref{ass:lipschitz}, and its $4\nu_1/\nu_2$-near-optimality dimension is $d>0$. Then, under Assumption~\ref{ass:partition} and for any $d'>d$, expected simple regret of HOO
\begin{equation}
    \epsilon_T = \mathbb{E}[r_T] = \bigO\left(T^{-\frac{1}{d'+2}}(\log T)^{\frac{1}{d'+2}}\right)
\end{equation}
\ifdoublecol
for $T > 1$, and $4\nu_1/\nu_2$-near-optimality dimension $d$ of $f$.
\else
for a horizon $T > 1$, and $4\nu_1/\nu_2$-near-optimality dimension $d$ of $f$.\fi 
\end{theorem}\vspace*{-1em}
\section{\framework{}: Procrastinated Tree Search}
In this section, we first provide a generic template for our framework. Following that, we incrementally show expected error bounds under delayed, noisy with known variance, noisy with unknown variance, and multi-fidelity feedbacks.
\begin{algorithm}[t!]
	\caption{\framework{} under DNF feedback and with a compatible \bandit{} algorithm }
	\label{alg:pcts}
	\begin{algorithmic}[1]
		\STATE{\textbf{Input:} } Total cost budget $\Lambda$, Bias function $\zeta$, Cost function $\lambda$, Smoothness parameters $(\nu_1, \rho)$.
		\STATE{\textbf{Initialization:} } $\tree_1 = \{(0,0)\}$ (root), $B_{(1,1)}^{\min}=B_{(1,2)}^{\min}=\infty$, $t=0$ (iteration), $C=0$ (cost)
		\WHILE{$C \leq \Lambda$}
		\STATE Compute $B^{\min}$ values for each node in $\tree_t$ using a UCB-type algorithm \bandit (Eq.~\eqref{eq:bmin}) \label{line:update}
		\STATE Select a leaf node $(h_t, l_t)$ by following an ``optimistic path" from root such that each selected node in the path has the highest $B^{\min}$ value among its sibling nodes
		\STATE Sample a point $x_t$ uniformly at random in the subdomain of node $(h_t, l_t)$
		\STATE Query the evaluation oracle with $x_t$ and at fidelity $z_{h_t}$
		\STATE Observe the delayed and noisy feedbacks $\mathcal{O}_t \triangleq \{f_{s|t}(x_{h_s,l_s}|z_{h_s}) + \epsilon_s:s+\tau_s = t\}$ with the timestamps of invoking these queries $\{s:s+\tau_s = t\}$
		\STATE Expand node $(h_t, l_t)$ and add its children to $\tree_t$ to form $\tree_{t+1}$
		\ENDWHILE
	\end{algorithmic}
\end{algorithm}\vspace*{-.5em}
\subsection{Algorithmic Framework}\label{sec:pcts}
\framework{} adapts the HOO algorithm~\citep{bubeck2011a} to delayed, noisy, and multi-fidelity feedbacks. We illustrate the pseudocode in Algorithm~\ref{alg:pcts}. Thus, in \framework{}, we first assign optimistic $B^{\min}$ values to each node $(h,l)$ in the hierarchical tree $\tree_t$. Then, we incrementally select an `optimistic path' from the root such that the path corresponds to one node at every depth and every chosen node has larger $B^{\min}$ value than its sibling nodes. Sibling nodes are the nodes that share the same parent. Following that, we sample a point $x_t$ randomly from the subdomain $X_{(h_t,l_t)}$ that the leaf node $(h_t,l_t)$ of the optimistic path represents. We expand this leaf node and add its children to $\tree_{t+1}$. Lines 4-6 and 9 essentially come from the HOO algorithm. The difference is in mainly three steps. In Line 7, we query the evaluation oracle with the point $x_t$ and fidelity $z_{h_t}$ due to multi-fidelity evaluator.
In Line 8, we observe a delayed set of noisy feedbacks $\mathcal{O}_t \triangleq \{f_{s|t}(x_{h_s,l_s}|z_{h_s}) + \epsilon_s:s+\tau_s = t\}$ that arrives with corresponding timestamps when the queries were invoked. Here, $f_{s|t}(x_{h_s,l_s}|z_{h_s})$ is the multi-fidelity feedback lower bounded by $f(x_{h_s,l_s}) - \zeta(z_{h_s}$), and $\epsilon_s$ is a noise with zero mean and bounded variance.
Such DNF feedback constrains us to use an asymptotically optimal bandit algorithm, \bandit{} (Line 4), that allows us to get an upper confidence bound $B_{(h,l),s,t}$, which would be immune to DNF. In the following sections, we incrementally design such \bandit{} confidence bounds  and derive corresponding error bounds for \framework{}.
Though we describe the algorithm and the analysis for given smoothness parameters $(\nu_1, \rho)$, we describe in Appendix B the details of how to extend \framework{} to unknown smoothness parameters.
\subsection{Adapting to Delayed Feedbacks}\label{sec:delay}

\paragraph{Observable Stochastic Delay Model.} We consider the stochastic delay setting~\citep{Joulani2013a,joulani2016delay}. This means that the feedback $f(x_s)$ of the evaluation oracle invoked at time $s \in [0,T]$ arrives with a delay $\tau_s \in \mathbb{R}^{\geq 0}$, such that $\{\tau_s\}_{s=0}^T$ are random variables invoked by an underlying but unknown stochastic process $\mathcal{D}$.
Here, the delays are independent of the algorithm's actions.
\begin{assumption}[Bounded Mean Delay]\label{ass:delay}
Delays are generated i.i.d from an unknown delay distribution $\mathcal{D}$. The expectation of delays $\tau \triangleq \expect[\tau_s:s\geq 0]$ is bounded and observable to the algorithm.
\end{assumption}
We observe that constant or deterministic delay with $\tau_{const} < \infty$ is a special case of this delay model.

\textbf{From \framework{} to Delayed Bandits.} Due to the delayed setting, we observe feedback of a query invoked at time $s$ at time $t \geq s$. Let us denote such feedback as $f_{s|t}(x_{s})$. Thus, at time $t$, \framework{} does not have access to all the invoked queries but a delayed subset of it: $\cup_{t'=1}^t\mathcal{O}_{t'} = \cup_{t=1}^t \{f_{s|t'}(x_{s}): s + \tau_s = t' \}$. At time $t$, \framework{} uses $\mathcal{O}_t$ to decide which node to extend next.
Thus, making \framework{} immune to unknown stochastic delays reduces to deployment of a \bandit{} algorithm that can handle such stochastic delayed feedback.

Multi-armed bandits with delayed feedback is an active research area~\citep{eick1988two,Joulani2013a,vernade2017stochastic,gael2020stochastic,pikeburke2018bandits}, where researchers have incrementally studied the known constant delay, the unknown observable stochastic delay, and the unknown anonymous stochastic delay settings.
In this paper, we operate in the second setting, where a delayed feedback comes with the timestamp of the query.
Under delayed feedback, designing an UCB-type bandit algorithm requires defining an optimistic confidence interval around the expected value of a given node $i$ that will consider both $T_i(t)$ and $S_i(t)$. $T_i(t)$ and $S_i(t)$ are the number of times a node $i$ is evaluated and the number of evaluation feedbacks observed until time $t$.
Given such delayed statistics, any UCB-like bandit algorithm computes $B_{i,s,t}$, i.e. the optimistic upper confidence bound for action $i$ at time $t$ (Table 2), and chooses the one with maximum $B_{i,s,t}$:
\begin{align*}
    i_t = \argmax_{i \in \mathcal{A}} B_{i,s,t}.
\end{align*}
We show three such confidence bounds in Table~\ref{tab:ucb}. Here, $\hat{\mu}_{i,s}$, $\hat{\sigma}^2_{i,s}$, and $\sigma^2$ are sample mean, sample variance, and predefined variance respectively. For the non-delayed setting, $s= T_i(t-1)$, and for delayed setting, $s=S_i(t-1)$.
\begin{table*}[t!]
\centering
\caption{Confidence Bounds for different bandit algorithms with delayed/non-delayed feedback.}\label{tab:ucb}\vspace*{-.5em}
\resizebox{0.9\textwidth}{!}{%
\begin{tabular}{c|c|c|c}
\toprule
\bandit{}   & \ducb{}~\citep{joulani2016delay}                            & \ducbs{}   & \ducbv{} \\
\midrule
$B_{i,s,t}$ & $\hat{\mu}_{i,s} + \sqrt{\frac{2 \log t}{s}}$ & $\hat{\mu}_{i,s} + \sqrt{\frac{2 \sigma^2 \log t}{s}}$ & $\hat{\mu}_{i,s} + \sqrt{\frac{2 \hat{\sigma}_{i,s}^2 \log t}{s}} + \frac{3b\log t}{s}$ \\
\bottomrule
\end{tabular}%
}\vspace*{-1em}
\end{table*}
Representing the modifications of UCB1~\citep{Auer2002}, UCB1-$\sigma$~\citep{Auer2002}, and UCB-V~\citep{ucbv} in such a general form allows us to extend them for delayed settings and incorporate them for node selection in \framework{}.
Thus, given an aforementioned UCB-like optimistic bandit algorithm, the leaf node $(h_t, l_t)$ selected by \framework{} at time $t$ is
\begin{align}\label{eq:bmin}
    &(h_t, l_t) \triangleq \argmax_{(h,l) \in \tree_t} B^{\min}_{(h,l)}(t)\notag\\ 
    &\triangleq \argmax_{(h,l) \in \tree_t} \min\lbrace B_{(h,l),s,t} + \nu_1\rho^h, \max_{(h',l')\in C(h,l)} B^{\min}_{(h,l)}(t)\rbrace.
\end{align}
Here, $\tree_t$ is the tree constructed at time $t$, and $C(h,l)$ is the set of children nodes of the node $(h,l)$.
Equation~\eqref{eq:bmin} is as same as that of HOO except that $B_{(h,l),s,t}$ is replaced by bounds in Table~\ref{tab:ucb}. Under these modified confidence bounds for delays, we derive the bound on expected regret of \frameworkucb{} that extends the regret analysis of bandits with delayed feedback to HOO~\citep{mcts}.
\begin{theorem}[Regret of \frameworkucb{} under Stochastic Delays]~\label{thm:regret_delay}
Under the same assumptions as Theorem~\ref{thm:HOO} and upper bound on expected delay $\tau$, \framework{} using Delayed-UCB1 (\ducb) achieves 
expected simple regret
\begin{align}
    \epsilon_T
    &= \bigO\left(\left(\frac{\ln T}{T}\right)^{\frac{1}{d'+2}} \left(1+ \frac{\tau}{\ln T}\right)^{\frac{1}{d'+2}}\right).
\end{align}
\end{theorem}
\begin{corollary}[Regret of \frameworkucb{} under Constant Delay]\label{cor:const_delay}
If the assumptions of Theorem~\ref{thm:HOO} hold, and the delay is constant, i.e. $\tau_{const} > 0$, the expected simple regret of \frameworkucb{} is
\begin{align}
    \epsilon_T
    &= \bigO\left(\left(\frac{\ln T}{T}\right)^{\frac{1}{d'+2}} \left(1+ \frac{\tau_{const}}{\ln T}\right)^{\frac{1}{d+2}}\right).
\end{align}
\end{corollary}
\textbf{Consequences of Theorem~\ref{thm:regret_delay}.} The bound of Theorem~\ref{thm:regret_delay} provides us with a few interesting insights.\\ 
1. \textit{Degradation due to delay:} we observe that the expected error of $\frameworkucb{}$ worsens by a factor of $\left(1+ \frac{\tau}{\ln T}\right)^{\frac{1}{d+2}}$ compared to HOO, which uses the non-delayed UCB1~\citep{Auer2002}. This is still significantly better than the other global optimization algorithm that can handle delay, such as Delayed Bandit Gradient Descent (DBGD)~\citep{li2019bandit} that achieves expected error bound $\sqrt{\frac{1}{T}+\frac{D}{T}}$. Here $D$ is the total delay.
Also, appearance of delay as an additive term in our analysis resonates with the proven results in bandits with delayed feedback,\ifdoublecol where an additive term appears due to delay. \else where an additive term appears in regret bounds due to delay. \fi For $d=0$, our bound matches in terms of $T$ and $\tau$ with the problem-independent lower bound of bandits with finite $K$-arms and constant delay, i.e. $\sqrt{(K/T + \tau/T)}$~\cite[Cor. 11]{cesa2016delay}, up to logarithmic factors.

2. \textit{Wait-and-act vs. \frameworkucb{}.} A na\"ive way to handle \textit{known constant delay} is to wait for the next $\tau_{const}$ time steps and to collect all the feedbacks in that interval to update the algorithm. In that case, the effective horizon becomes $\frac{T}{\tau_{const}}$. Thus, the corresponding error bound will be $\bigO\left(T^{-\frac{1}{d'+2}} \left(\tau_{const} \ln T\right)^{\frac{1}{d+2}}\right)$. This is still higher than our error bound in Equation 5 for \textit{unknown constant delay} $\tau_{const}>1$ and $T\geq 3$. 

3. \textit{Deeper trees.} While proving Theorem~\ref{thm:regret_delay}, we observe that the depth $H>0$ achieved by \frameworkucb{} till time $T$ is such that $\rho^{-H(d'+2)} \geq \frac{T}{\tau+\ln T}$. This implies that for a fixed horizon $T$, the achieved depth should be $H \geq \frac{1}{d'+2}\frac{\tau+\ln T}{\ln (1/\rho)} = \Omega(\tau+\ln T)$. In contrast, HOO grows a tree of depth $H = \Omega(\ln(T/\tau))$. This shows that \frameworkucb{} constructs a deeper tree than HOO.

4. \textit{Benign and adversarial delays.} If the expected delay is $\bigO(\ln T)$ , the expected simple regret is practically of the same order as that of non-delayed feedbacks. Thus, in cases of applications where introducing artificial delays helps in improving the computational cost~\citep{Wang2021}, we can tune the delays to $\bigO(\ln T)$ for a given horizon $T$ without harming the accuracy. We refer to this range of delays as \textit{benign delay}.
In contrast, one can consider delay distributions that have tails with $\alpha$-polynomial decay, i.e. the expected delay is $\bigO(T^{1-\alpha})$ for $\alpha \in(0,1)$. 
In that case, the expected error is at least $\tilde{\bigO}(T^{-\frac{\alpha}{d+2}})$. Thus, it worsens the HOO bound by a factor of $T^{\frac{1-\alpha}{d+2}}$. This observation in error bound resonates with the impossibility result of~\citep{gael2020stochastic} that, in case of delays with $\alpha$-polynomial tails, a delayed bandit algorithm cannot achieve total expected regret lower than $(T^{1-\alpha})$. Thus, it is unexpected that any hierarchical tree search with such delays achieves expected error better than $\bigO(T^{-\frac{\alpha}{d+2}})$.

\begin{table*}[t!]
\centering
\caption{Per-step cost $\lambda(Z_h)$ and total number of iterations $H(\Lambda)$  for different Fidelity models.}\label{tab:fidelity}\vspace*{-.5em}
\resizebox{\textwidth}{!}{%
\begin{tabular}{c|c|c|c|c}
\toprule
Fidel. Model   & Linear Growth                           & Constant   & Polynomial Decay & Exponential Decay   \\
\midrule
$\lambda(Z_h)$ & $\min\{\beta h, \lambda(1)\}$, $\beta>0$ & $\min\{\beta, \lambda(1)\}$, $\beta>0$ & $\min\{h^{-\beta}, \lambda(1)\}$, $\beta>0, \neq 1$ & $\min\{\beta^{-h}, \lambda(1)\}$, $\beta \in (0,1]$  \\
$H(\Lambda)$    & $\sqrt{2(2\Lambda - \lambda(1))/\beta}$ & ${2(2\Lambda - \lambda(1))/\beta}$ & $(1+(1-\beta)(2\Lambda-\lambda(1)))^{1/(1-\beta)}$ & $\log_{1/\beta}\left(1+(1-\beta)(2\Lambda - \lambda(1))\right)$       \\
\bottomrule
\end{tabular}%
}
\end{table*}
\subsection{Adapting to Delayed and Noisy Feedback}\label{sec:noise}
Typically, when we evaluate the objective function at any time step $t$, we obtain a noisy version of the function as feedback such that $\Tilde{f}(X_t) = f(X_t) + \epsilon_t$. Here, $\epsilon_t$ is a noise sample independently generated from a noise distribution $\mathcal{N}$ with mean $0$. Till now, we did not explicitly consider the noise for the action selection step. In this section, we provide analysis for both known and unknown variance cases.
In both cases, we assume that the noise has bounded variance $\sigma^2$, i.e. sub-Gaussian. In general, this assumption can be imposed in the present setup as any noisy evaluation can be clipped in the range of the evaluations where we optimize the objective function. It is known that a bounded random variable is sub-Gaussian with bounded mean and variance.

\textbf{Case 1: Known Variance.} Let us assume that the variance of the noise is known, say $\sigma^2 > 0$. In this case, the optimistic $B$-values can be computed using a simple variant of delayed-UCB1, i.e. delayed-UCB-$\sigma$ (\ducbs), where
\ifdoublecol
\begin{align}\label{eq:noisy_ducb1}\hspace*{-1.1em}
    B_{(h,i),S_{(h,i)}(t-1),t} &\triangleq \hat{\mu}_{(h,i),S_{(h,i)}(t-1)} + \sqrt{\frac{2 \sigma^2 \log t}{S_{(h,i)}(t-1)}}.
\end{align}
\else
\begin{align}\label{eq:noisy_ducb1}
    B_{(h,i),S_{(h,i)}(t-1),t} &\triangleq \hat{\mu}_{(h,i),S_{(h,i)}(t-1)} + \sqrt{\frac{2 \sigma^2 \log t}{S_{(h,i)}(t-1)}}.
\end{align}
\fi
Here, $\hat{\mu}_{(h,i),S_{(h,i)}(t-1)}$ is the empirical mean computed using noisy evaluations obtained till time $t$, i.e. $\cup_{t'=0}^t\mathcal{O}_{t}$, and for node $(h,i)$. In multiple works, this known noise setup and $UCB-\sigma^2$ algorithm has been considered in tree search algorithms without delays.
\begin{theorem}\label{thm:regret_delay_noise_known}
Let us assume that the variance of the noise in evaluations is $\sigma^2$ and is available to the algorithm. Then, under the same assumptions as Theorem~\ref{thm:HOO} and upper bound on expected delay $\tau$, \framework{} using \ducbs{} for node selection achieves
expected simple regret
\begin{align}
    \epsilon_T
    &= \bigO\left(T^{-\frac{1}{d'+2}} \left((\sigma/\nu_1)^2 \ln T+ {\tau}{}\right)^{\frac{1}{d+2}}\right).
\end{align}
\end{theorem}
\noindent\textbf{Effect of Known Noise.} We observe that even with no delay, i.e. $\tau=0$, noisy feedback with known variance $\sigma^2$ worsens the bound of HOO with noiseless evaluations by $\sigma^{2/(d+2)}$. 

\textbf{Case 2: Unknown Variance.} If the variance of the noise is unknown, we have to estimate the noise variance empirically from evaluations. Given the evaluations $\{\Tilde{f}(X_1)\}_{t=0}^T$ and the delayed statistics $S_{(h,i)}(t-1)$ of node $(h,i)$, the empirical noise variance at time $t$ is
{\small{$\hat{\sigma}_{(h,i),S_{(h,i)}(t-1)}^2 \triangleq$ $\frac{1}{S_{(h,i)}(t-1)} \sum_{j=1}^{S_{(h,i)}(t-1)} (\Tilde{f}(X_j)\mathds{1}[(H_j, I_j)=(h,i)]-\hat{\mu}_{(h,i),S_{(h,i)}(t-1)})^2$,} }
where empirical mean $\hat{\mu}_{(h,i),S_{(h,i)}(t-1)} \triangleq \frac{1}{S_{(h,i)}(t-1)} \sum_{j=1}^{S_{(h,i)}(t-1)} \Tilde{f}(X_j)\mathds{1}[(H_j, I_j)=(h,i)]$.
Using the empirical mean and variance of functional evaluations for each node $(h,i)$, we now define a delayed-UCBV (\ducbv)
confidence bound for selecting next node:
\begin{align}\label{eq:ducbv}
    &B_{(h,i),S_{(h,i)}(t-1),t} \triangleq \hat{\mu}_{(h,i),S_{(h,i)}(t-1)}\notag\\ &+ \sqrt{\frac{2 \hat{\sigma}_{(h,i),S_{(h,i)}(t-1)}^2 \log t}{S_{(h,i)}(t-1)}} + \frac{3b \log t}{S_{(h,i)}(t-1)}.
\end{align} 
In practice, we do not need an exact value of $b$. We can use a large proxy value such that the feedback is bounded by it.
\begin{theorem}\label{thm:regret_delay_noise}
Let us assume that the upper bound on variance of the noise in evaluations is $\sigma^2$, which is unknown to the algorithm. If $[0,b]$ is the range of $f$, under the same assumptions as of Theorem~\ref{thm:HOO}, \framework{} using \ducbv{} achieves 
expected simple regret
\begin{align*}\hspace*{-1em}
    \epsilon_T
    &= \bigO\left(T^{-\frac{1}{d'+2}} \left(((\sigma/\nu_1)^2 +2b/\nu_1)\ln T+ {\tau}\right)^{\frac{1}{d'+2}}\right).
\end{align*}
\end{theorem}
\textbf{Effect of Unknown Noise.} \textit{Adapting UCB-V in the stochastic delay setting and using the corresponding bound in \framework{} allows us to extend hierarchical tree search for unknown noise both in presence and absence of delays.} To the best of our knowledge, this paper is the first to extend HOO framework for unknown noise, and also UCB-V to stochastic delays. This adaptation to unknown noise comes at a cost of $(((\sigma/\nu_1)^2+2b/\nu_1)\ln T)^{1/(d+2)}$ in expected error, whereas for known noise variance, it is $((\sigma/\nu_1)^2 \ln T)^{1/(d+2)}$.
\subsection{Adapting to Delayed, Noisy, and Multi-fidelity (DNF) Feedback}\label{sec:mf}
Now, let us consider that we do not only have a delayed and noisy functional evaluator at each step but also an evaluator with different fidelity at each level $h$ of the tree. This setup of multi-fidelity HOO without unknown noise and delay was first considered in~\citep{sen19_MFHOO}. We extend their schematic to the version with delayed and noisy feedback with unknown delays and noise.
Following the multi-fidelity formulation of~\citep{sen18_mfdoo,sen19_MFHOO}, we consider the mean of the multi-fidelity query, $f_z(x)$, as biased, and progressively smaller bias can be obtained but with varying costs. The cost of selecting a new node at level $h > 0$ is $\lambda(Z_h) \in \real^+$ and the bias added in the decision due to the limited evaluation is $\zeta(Z_h) \in \real^+$. Here, the bias function is monotonically decreasing, and $Z_h \in \mathcal{Z}$ is the state of fidelity of the multi-fidelity evaluator, which influences both the cost of and the bias in evaluation. Thus, the evaluation at $x_s$ is $\tilde{f}(x_{(h_s, l_s)}|z_{h,s}) \triangleq f(x_{(h_s,l_s)}) + \epsilon_s + \zeta(z_{h_s})$.
Hence, under DNF feedback, the \ducbv{} selection rule becomes
\ifdoublecol
\begin{align*}\label{eq:ducbv-fidelity1}\hspace*{-1.em}
        &B_{(h,i),S_{(h,i)}(t-1),t} \triangleq \hat{\mu}_{(h,i),S_{(h,i)}(t-1)}\notag\\ &+ \sqrt{\frac{2 \hat{\sigma}_{(h,i),S_{(h,i)}(t-1)}^2 \log t}{S_{(h,i)}(t-1)}} + \frac{3b \log t}{S_{(h,i)}(t-1)} + \zeta(Z_h).
\end{align*}
\else
\begin{align}\label{eq:ducbv-fidelity1}
        B_{(h,i),S_{(h,i)}(t-1),t} &\triangleq \hat{\mu}_{(h,i),S_{(h,i)}(t-1)} + \sqrt{\frac{2 \hat{\sigma}_{(h,i),S_{(h,i)}(t-1)}^2 \log t}{S_{(h,i)}(t-1)}} + \frac{3b \log t}{S_{(h,i)}(t-1)} + \zeta(Z_h).
\end{align}
\fi
Here, the empirical mean and variance are computed using the multi-fidelity and delayed feedbacks.
We do not need to know $\zeta$ for the algorithm but we assume it to be known for the analysis. Given this update rule and the multi-fidelity model, we observe that the Lemma 1 of~\citep{sen19_MFHOO} holds. Given a total budget $\Lambda$ and the multi-fidelity selection rule,  
the total number of iterations that the algorithms runs for is $T(\Lambda) \geq H(\Lambda) + 1,~\text{where}~H(\Lambda) \triangleq \max \lbrace H: \sum_{h=1}^H \lambda(Z_h) \leq \Lambda \rbrace$
Thus, we can retain the previously derived bounds of Theorem~\ref{thm:regret_delay} and~\ref{thm:regret_delay_noise} by substituting $T= H(\Lambda)$.
\begin{corollary}[\frameworkucbv{} under DNF Feedback]\label{thm:regret_multifidel}
If the function under evaluation has $h$-dependent fidelity such that $ H(\Lambda) \triangleq \max \lbrace H: \sum_{h=1}^H \lambda(Z_h) \leq \Lambda \rbrace$ and the induced bias $\zeta(Z_h) = \nu_1 \rho^h$, then under the same assumptions as of Theorem~\ref{thm:HOO}, \framework{} using \ducb{} achieves
\begin{align*}
    \epsilon_\Lambda
    &= \bigO\left((H(\Lambda))^{-\frac{1}{d'+2}} \left(\ln H(\Lambda)+ {\tau}\right)^{\frac{1}{d'+2}}\right),
\end{align*}
and \framework{} using \ducbv{} achieves expected simple regret
\begin{align*}
    &\epsilon_\Lambda=\notag\\
    &\bigO\left((H(\Lambda))^{-\frac{1}{d'+2}} \left(\ln H(\Lambda)+ \frac{\tau}{(\sigma/\nu_1)^2+2b/\nu_1}\right)^{\frac{1}{d'+2}}\right).
\end{align*}
\end{corollary}

\textbf{Models of Multi-fidelity.} Depending on the evaluation problem, we may have different cost functions. In Table~\ref{tab:fidelity}, we instantiate the cost model, bias model, and total number of iterations for four multi-fidelity models with linear growth, constant, polynomially decaying, and exponentially decaying costs of evaluations. The linear growth, polynomial decays, and exponential decays are observed in the cases of hyperparameter tuning of deep-learning models, database optimization, and tuning learning rates of convex optimization respectively. Further details are in Appendix C.

\begin{figure*}[t!]
    \centering
    \includegraphics[clip, width=\textwidth]{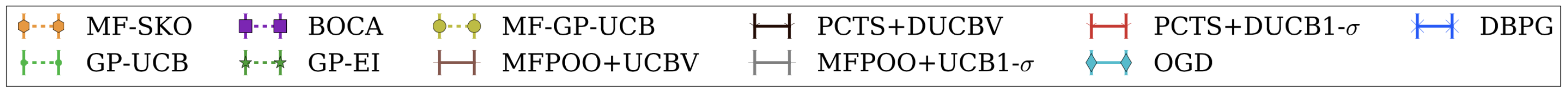}
    \includegraphics[clip, width=\textwidth]{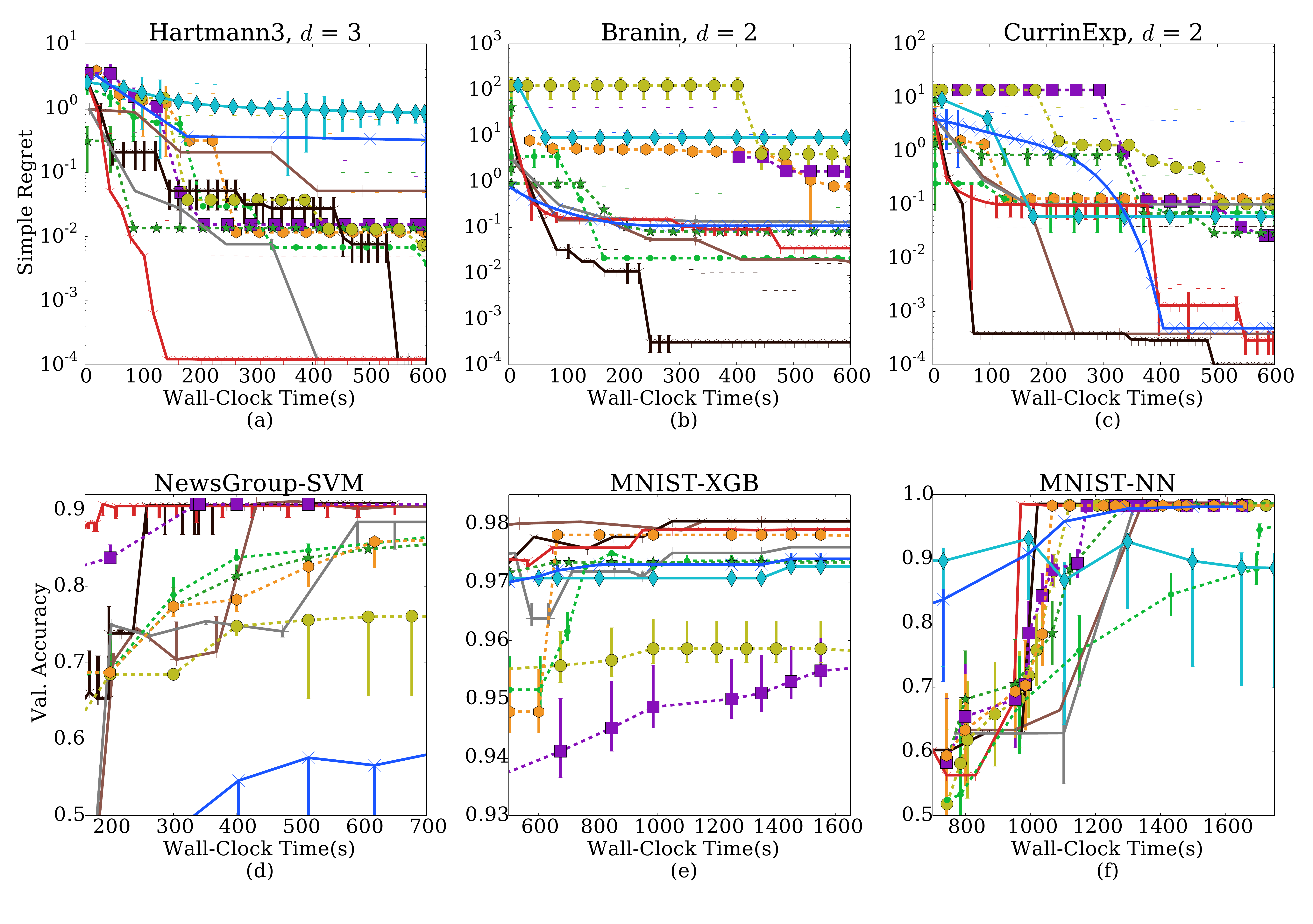}\vspace*{-1.6em}
    \caption{Figures (a) to (c) show simple regret (median of 10 runs) of different algorithms on synthetic functions with DNF feedbacks. Figures (d) to (f) show the cross-validation accuracy (median of 5 runs) achieved on the hyperparameter tuning of classifiers on datasets with DNF feedbacks.}\label{fig:experiement}
    \vspace*{-1em}
\end{figure*}

\section{Experimental Analysis}\label{sec:experiments}
\textbf{Experimental Setup.}
Similar to prior work on tree search with multi-fidelity and known noise~\citep{sen18_mfdoo, sen19_MFHOO}, we evaluate performance of \framework{} on both synthetic functions and machine learning models operating on real-data but under delayed, noisy (unknown), and multi-fidelity (DNF) feedback. We compare the performance of \framework{}  with: BO algorithms (BOCA \citep{boca}, GP-UCB \citep{gpucb}, MF-GP-UCB \citep{mfgpucb}, GP-EI \citep{gpei}, MF-SKO \citep{sko})\footnote{We use the implementations in \url{https://github.com/rajatsen91/MFTreeSearchCV} for baselines except OGD, DBGD, and MFPOO-UCBV. For BO algorithms, this implementation chooses the best among the polynomial kernel, coordinate-wise product kernel and squared exponential kernel for each problem.}, tree search algorithms (MFPOO \citep{sen18_mfdoo}, MFPOO with UCB-V~\citep{ucbv}), zeroth-order GD algorithms (OGD, DBGD \citep{li2019bandit}).

In our experiments, cost is the time required for a function evaluation. Delay is the time required to communicate the result to the \framework{} algorithm. The cost-function and delay can be general but in our experiments, we focus on time-efficiency. Thus, we plot the convergence of the simple regret of competing algorithms with respect to the wall-clock time.
In our experiments, we do not assume the smoothness parameters, the bias function, and the cost function to be known. The smoothness parameters are computed in a similar manner as POO and MFPOO. For comparison, we keep the delay constant and use wait-and-act versions of delay-insensitive baselines. The experiments with stochastic delay are elaborated in Appendix, where \framework{} variants perform even better in comparison with the constant delay setups.

\textbf{Synthetic Functions.}
We illustrate results for three different synthetic functions, Hartmann3~\citep{hartmann3}, Branin~\citep{hartmann3}, and CurrinExp~\citep{currin} with noise variances $\sigma^2=0.01$, $0.05$, and $0.05$ respectively. We follow the fidelity setup of~\citep{sen18_mfdoo, sen19_MFHOO}, that modifies the synthetic functions to incorporate the fidelity space $\mathcal{Z} = [0, 1]$. The delay time $\tau$ for all synthetic functions is set to four seconds. We choose to add noise from Gaussian distributions with variance $\sigma^2$. Note that the noise can be added from any distribution with variance $\leq \sigma^2$. This $\sigma$ is passed to UCB1-$\sigma$ and DUCB1-$\sigma$ in MFPOO and \framework{} as it assumes the noise variance is known~\citep{sen19_MFHOO}.
We implement all  baselines in Python (version 2.7). We run each experiment ten times for 600s on a MacBook Pro with a 6-core Intel(R) Xeon(R)@2.60GHz CPU and plot the median value of simple regret, i.e. $l_1$ distance between the value of current best point and optimal value, for each algorithm. 







\textbf{Real Data: Hyperparameter Tuning.}
We evaluate the aforementioned algorithms on a 32-core Intel(R) Xeon(R)@2.3 GHz server for hyperparameter tuning of SVM on News Group dataset, and XGB and Neural Network on MNIST datasets. We use corresponding scikit-learn modules~\citep{sklearn} for training all the classifiers. For each tuning task, we plot the median value of cross-validation accuracy in five runs for 700s, 1700s, and 1800s respectively. We set $\sigma^2=0.02$ for algorithms where $\sigma$ is known, and $b=1$ where UCBV and \ducbv are used.

\ifdoublecol
\ 
\else
\textbf{SVM on NewsGroup.} We evaluate the algorithms to tune hyper-parameters of SVM classifier on the NewsGroup dataset~\citep{newsweeder}. The hyper-parameters to tune are the regularization term $C$, ranging from $[e^{-5}, e^{5}]$, and the kernel temperature $\gamma$ from the range $[e^{-5}, e^5]$. Both are accessed in log scale.
We set the delay $\tau$ to four seconds and the fidelity range $\mathcal{Z}=[0, 1]$ is mapped to $[100,7000]$. The fidelity range represents the number of samples used to train the SVM classifier with the chosen parameters.  We plot the 5-fold cross-validation accuracy in Figure~\ref{fig:experiement}(d).

\textbf{XGB on MNIST.} We tune hyperparameters of XGBOOST~\citep{xgboost} on the MNIST dataset~\citep{mnist}, where the hyperparameters are: 
\begin{enumerate*}[label=(\roman*)]
    \item \texttt{max\_depth} in $[2, 13]$,
    \item \texttt{n\_estimators} in $[10, 400]$,
    \item \texttt{colsample\_bytree} in $[0.2, 0.9]$,
    \item \texttt{gamma} in $[0, 0.7]$, and
    \item  \texttt{learning\_rate} ranging from $[0.05, 0.3]$. 
\end{enumerate*} 
The delay $\tau$ is set to ten seconds and the fidelity range $\mathcal{Z} = [0,1]$ is mapped to the training sample range $[500,20000]$. We plot the 3-fold cross-validation accuracy in Figure~\ref{fig:experiement}(e). 


\textbf{NN on MNIST.} We also apply the algorithms for tuning the hyper-parameters of a three layer multi layer perceptron (MLP) neural network (NN) classifier on the MNIST dataset~\citep{mnist}. Here, the hyper-parameters being tuned are:
\begin{enumerate*}[label=(\roman*)]
    \item number of neurons of the first, second, and third layers, which belong to the ranges $[32, 128]$, $[128, 256]$, and $[256, 512]$, respectively,
    \item initial learning rate of optimizer in $[e^{-1}, e^{-5}]$ (accessed in log-scale),
    \item optimizers from (`lbfgs', `sgd', `adam'),
    \item activation function from (`tanh', `relu', `logistic'),
    \item early\_stopping from (`enable', `disable').
\end{enumerate*}
The delay $\tau$ for this experiment is 20 seconds. The number of training samples corresponding to the fidelities $z=0$ and $1$ are 1000 and 60000 respectively.  We plot the 3-fold cross-validation accuracy in Figure~\ref{fig:experiement}(f).
\fi
\textbf{Summary of Results.} \textit{In all of the experiments, we observe that either \frameworkucb{} or \frameworkucbv{} outperforms the competing algorithms in terms of convergence speed.} Also, in case of synthetic functions, they achieve approximately $1$ to $3$ order lower simple regret. These results empirically validate the efficiency of \framework{} to adapt to DNF feedback.
Due to lack of space, further implementation details, results on tree depth, performance for stochastic delays, and error statistics for other synthetic functions and hyperparamter tuning experiments are deferred to appendix.

\section{Discussion and Future Work}
We propose a generic tree search approach \framework{} for black-box optimization problems with DNF feedbacks. We provide a generic analysis to bound the expected simple regret of \framework{} given a horizon $T$.
We instantiate \framework{} with delayed-UCB1 and delayed-UCBV for observable stochastic delays, and known and unknown noises respectively. Our analysis shows that the expected simple regret for \framework{} worsens by a constant factor and $T^{\frac{1-\alpha}{d+2}}$ for expected delay of $O(\log T)$ and $O(T^{1-\alpha})$ respectively. 
We also experimentally show that \framework{} outperforms other global optimizers incompatible or individually tolerant to noise, delay, or multi-fidelity on both synthetic and real-world functions.
In addition, our work extends UCB-V to stochastic delays.

In the future, we plan to consider anonymous delay feedbacks in order to develop tree search optimizers that respect privacy. It also shows the need for proving a problem-independent lower bound for hierarchical tree search with stochastic delay. The other possible direction is to deploy \framework{} for planning in Markov Decision Processes with delay, where tree search algorithms have been successful.

\section*{Acknowledgment}
This research project is supported by NSF grant IIS-1910830 (“Regret-Bounded Query Evaluation via Reinforcement Learning”).

\bibliographystyle{apalike}
\bibliography{reference}

\begin{thebibliography}{}

\bibitem[Agarwal and Duchi, 2012]{agarwal2012distributed}
Agarwal, A. and Duchi, J.~C. (2012).
\newblock Distributed delayed stochastic optimization.
\newblock In {\em 2012 IEEE 51st IEEE Conference on Decision and Control
  (CDC)}, pages 5451--5452. IEEE.

\bibitem[Agrawal, 1995]{agrawal1995continuum}
Agrawal, R. (1995).
\newblock The continuum-armed bandit problem.
\newblock {\em SIAM journal on control and optimization}, 33(6):1926--1951.

\bibitem[Audibert et~al., 2007]{ucbv}
Audibert, J.-Y., Munos, R., and Szepesv{\'a}ri, C. (2007).
\newblock Tuning bandit algorithms in stochastic environments.
\newblock In Hutter, M., Servedio, R.~A., and Takimoto, E., editors, {\em
  Algorithmic Learning Theory}, pages 150--165, Berlin, Heidelberg. Springer
  Berlin Heidelberg.

\bibitem[Auer et~al., 2002]{Auer2002}
Auer, P., Cesa-bianchi, N., and Fischer, P. (2002).
\newblock {Finite time analysis of the multiarmed bandit problem}.
\newblock {\em Machine Learning}, 47(2-3):235--256.

\bibitem[Auer et~al., 2007]{auer2007improved}
Auer, P., Ortner, R., and Szepesv{\'a}ri, C. (2007).
\newblock Improved rates for the stochastic continuum-armed bandit problem.
\newblock In {\em International Conference on Computational Learning Theory},
  pages 454--468. Springer.

\bibitem[Azar et~al., 2014]{lazaric2014online}
Azar, M.~G., Lazaric, A., and Brunskill, E. (2014).
\newblock Online stochastic optimization under correlated bandit feedback.
\newblock In {\em International Conference on Machine Learning}, pages
  1557--1565. PMLR.

\bibitem[B{\"a}ck and Schwefel, 1993]{schwefel}
B{\"a}ck, T. and Schwefel, H.-P. (1993).
\newblock An overview of evolutionary algorithms for parameter optimization.
\newblock {\em Evolutionary computation}, 1(1):1--23.

\bibitem[Bubeck et~al., 2011]{bubeck2011a}
Bubeck, S., Munos, R., Stoltz, G., and Szepesv{\'a}ri, C. (2011).
\newblock X-armed bandits.
\newblock {\em Journal of Machine Learning Research}, 12(5).

\bibitem[Buitinck et~al., 2013]{sklearn}
Buitinck, L., Louppe, G., Blondel, M., Pedregosa, F., Mueller, A., Grisel, O.,
  Niculae, V., Prettenhofer, P., Gramfort, A., Grobler, J., Layton, R.,
  VanderPlas, J., Joly, A., Holt, B., and Varoquaux, G. (2013).
\newblock {API} design for machine learning software: experiences from the
  scikit-learn project.
\newblock In {\em ECML PKDD Workshop: Languages for Data Mining and Machine
  Learning}, pages 108--122.

\bibitem[Cesa-Bianchi et~al., 2016]{cesa2016delay}
Cesa-Bianchi, N., Gentile, C., Mansour, Y., and Minora, A. (2016).
\newblock Delay and cooperation in nonstochastic bandits.
\newblock In {\em Conference on Learning Theory}, pages 605--622. PMLR.

\bibitem[Chen, 1988]{chen1988lower}
Chen, H. (1988).
\newblock Lower rate of convergence for locating a maximum of a function.
\newblock {\em The Annals of Statistics}, pages 1330--1334.

\bibitem[Chen and Guestrin, 2016]{xgboost}
Chen, T. and Guestrin, C. (2016).
\newblock Xgboost: A scalable tree boosting system.
\newblock In {\em Proceedings of the 22nd acm sigkdd international conference
  on knowledge discovery and data mining}, pages 785--794.

\bibitem[Currin et~al., 1988]{currin}
Currin, C., Mitchell, T., Morris, M., and Ylvisaker, D. (1988).
\newblock A bayesian approach to the design and analysis of computer
  experiments.
\newblock Technical report, Oak Ridge National Lab., TN (USA).

\bibitem[Eick, 1988]{eick1988two}
Eick, S.~G. (1988).
\newblock The two-armed bandit with delayed responses.
\newblock {\em The Annals of Statistics}, pages 254--264.

\bibitem[Fischer et~al., 2015]{fischer2015machines}
Fischer, L., Gao, S., and Bernstein, A. (2015).
\newblock Machines tuning machines: Configuring distributed stream processors
  with bayesian optimization.
\newblock In {\em 2015 IEEE International Conference on Cluster Computing},
  pages 22--31. IEEE.

\bibitem[Gael et~al., 2020]{gael2020stochastic}
Gael, M.~A., Vernade, C., Carpentier, A., and Valko, M. (2020).
\newblock Stochastic bandits with arm-dependent delays.
\newblock In {\em International Conference on Machine Learning}, pages
  3348--3356. PMLR.

\bibitem[Goldstein, 1977]{goldstein1977optimization}
Goldstein, A. (1977).
\newblock Optimization of lipschitz continuous functions.
\newblock {\em Mathematical Programming}, 13(1):14--22.

\bibitem[Grill et~al., 2020]{grill2020monte}
Grill, J.-B., Altch{\'e}, F., Tang, Y., Hubert, T., Valko, M., Antonoglou, I.,
  and Munos, R. (2020).
\newblock Monte-carlo tree search as regularized policy optimization.
\newblock In {\em International Conference on Machine Learning}, pages
  3769--3778. PMLR.

\bibitem[Grill et~al., 2015]{poo_first}
Grill, J.-B., Valko, M., Munos, R., and Munos, R. (2015).
\newblock Black-box optimization of noisy functions with unknown smoothness.
\newblock In Cortes, C., Lawrence, N., Lee, D., Sugiyama, M., and Garnett, R.,
  editors, {\em Advances in Neural Information Processing Systems}, volume~28.
  Curran Associates, Inc.

\bibitem[Hey, 1979]{branin}
Hey, A.~M. (1979).
\newblock Towards global optimisation 2.
\newblock {\em Journal of the Operational Research Society}, 30(9):844--844.

\bibitem[Huang et~al., 2006]{sko}
Huang, D., Allen, T.~T., Notz, W.~I., and Miller, R.~A. (2006).
\newblock Sequential kriging optimization using multiple-fidelity evaluations.
\newblock {\em Structural and Multidisciplinary Optimization}, 32(5):369--382.

\bibitem[Jamieson et~al., 2012]{jamieson2012query}
Jamieson, K.~G., Nowak, R.~D., and Recht, B. (2012).
\newblock Query complexity of derivative-free optimization.
\newblock {\em arXiv preprint arXiv:1209.2434}.

\bibitem[Jones et~al., 1998]{gpei}
Jones, D.~R., Schonlau, M., and Welch, W.~J. (1998).
\newblock Efficient global optimization of expensive black-box functions.
\newblock {\em J. of Global Optimization}, 13(4):455–492.

\bibitem[Joulani et~al., 2013]{Joulani2013a}
Joulani, P., Gyorgy, A., and Szepesv{\'a}ri, C. (2013).
\newblock Online learning under delayed feedback.
\newblock In {\em International Conference on Machine Learning}, pages
  1453--1461. PMLR.

\bibitem[Joulani et~al., 2016]{joulani2016delay}
Joulani, P., Gyorgy, A., and Szepesv{\'a}ri, C. (2016).
\newblock Delay-tolerant online convex optimization: Unified analysis and
  adaptive-gradient algorithms.
\newblock In {\em Proceedings of the AAAI Conference on Artificial
  Intelligence}, volume~30.

\bibitem[Kajita et~al., 2020]{kajita2020autonomous}
Kajita, S., Kinjo, T., and Nishi, T. (2020).
\newblock Autonomous molecular design by monte-carlo tree search and rapid
  evaluations using molecular dynamics simulations.
\newblock {\em Communications Physics}, 3(1):1--11.

\bibitem[Kandasamy et~al., 2016]{mfgpucb}
Kandasamy, K., Dasarathy, G., Oliva, J.~B., Schneider, J., and Poczos, B.
  (2016).
\newblock Gaussian process bandit optimisation with multi-fidelity evaluations.
\newblock In Lee, D., Sugiyama, M., Luxburg, U., Guyon, I., and Garnett, R.,
  editors, {\em Advances in Neural Information Processing Systems}, volume~29.
  Curran Associates, Inc.

\bibitem[Kandasamy et~al., 2017]{boca}
Kandasamy, K., Dasarathy, G., Schneider, J., and P{\'o}czos, B. (2017).
\newblock Multi-fidelity bayesian optimisation with continuous approximations.
\newblock In {\em International Conference on Machine Learning}, pages
  1799--1808. PMLR.

\bibitem[Kleinberg et~al., 2008a]{10.1145/1374376.1374475}
Kleinberg, R., Slivkins, A., and Upfal, E. (2008a).
\newblock Multi-armed bandits in metric spaces.
\newblock In {\em Proceedings of the Fortieth Annual ACM Symposium on Theory of
  Computing}, STOC '08, page 681–690, New York, NY, USA. Association for
  Computing Machinery.

\bibitem[Kleinberg et~al., 2008b]{kleinberg2008multi}
Kleinberg, R., Slivkins, A., and Upfal, E. (2008b).
\newblock Multi-armed bandits in metric spaces.
\newblock In {\em Proceedings of the fortieth annual ACM symposium on Theory of
  computing}, pages 681--690.

\bibitem[Kleinberg et~al., 2019]{kleinberg2019bandits}
Kleinberg, R., Slivkins, A., and Upfal, E. (2019).
\newblock Bandits and experts in metric spaces.
\newblock {\em Journal of the ACM (JACM)}, 66(4):1--77.

\bibitem[Kumagai, 2017]{kumagai2017regret}
Kumagai, W. (2017).
\newblock Regret analysis for continuous dueling bandit.
\newblock {\em arXiv preprint arXiv:1711.07693}.

\bibitem[Lang, 1995]{newsweeder}
Lang, K. (1995).
\newblock Newsweeder: Learning to filter netnews.
\newblock In {\em Machine Learning Proceedings 1995}, pages 331--339. Elsevier.

\bibitem[Langford et~al., 2009]{langford2009slow}
Langford, J., Smola, A., and Zinkevich, M. (2009).
\newblock Slow learners are fast.
\newblock {\em arXiv preprint arXiv:0911.0491}.

\bibitem[Lattimore and Szepesv{\'a}ri, 2020]{lattimore2020bandit}
Lattimore, T. and Szepesv{\'a}ri, C. (2020).
\newblock {\em Bandit algorithms}.
\newblock Cambridge University Press.

\bibitem[LeCun et~al., 1998]{mnist}
LeCun, Y., Bottou, L., Bengio, Y., and Haffner, P. (1998).
\newblock Gradient-based learning applied to document recognition.
\newblock {\em Proceedings of the IEEE}, 86(11):2278--2324.

\bibitem[Li et~al., 2019]{li2019bandit}
Li, B., Chen, T., and Giannakis, G.~B. (2019).
\newblock Bandit online learning with unknown delays.
\newblock In {\em The 22nd International Conference on Artificial Intelligence
  and Statistics}, pages 993--1002. PMLR.

\bibitem[Liu et~al., 2018]{liu2018zeroth}
Liu, S., Li, X., Chen, P.-Y., Haupt, J., and Amini, L. (2018).
\newblock Zeroth-order stochastic projected gradient descent for nonconvex
  optimization.
\newblock In {\em 2018 IEEE Global Conference on Signal and Information
  Processing (GlobalSIP)}, pages 1179--1183. IEEE.

\bibitem[Locatelli and Carpentier, 2018]{locatelli2018}
Locatelli, A. and Carpentier, A. (2018).
\newblock Adaptivity to smoothness in x-armed bandits.
\newblock In {\em Conference on Learning Theory}, pages 1463--1492. PMLR.

\bibitem[Martinez-Cantin, 2017]{martinez2017bayesian}
Martinez-Cantin, R. (2017).
\newblock Bayesian optimization with adaptive kernels for robot control.
\newblock In {\em 2017 IEEE International Conference on Robotics and Automation
  (ICRA)}, pages 3350--3356. IEEE.

\bibitem[Munos, 2014]{mcts}
Munos, R. (2014).
\newblock From bandits to monte-carlo tree search: The optimistic principle
  applied to optimization and planning.
\newblock {\em Foundations and Trends{\textregistered} in Machine Learning},
  7(1):1--129.

\bibitem[Nguyen et~al., 2019]{nguyen2019filtering}
Nguyen, V., Gupta, S., Rana, S., Li, C., and Venkatesh, S. (2019).
\newblock Filtering bayesian optimization approach in weakly specified search
  space.
\newblock {\em Knowledge and Information Systems}, 60(1):385--413.

\bibitem[Oh et~al., 2018]{oh2018bock}
Oh, C., Gavves, E., and Welling, M. (2018).
\newblock Bock: Bayesian optimization with cylindrical kernels.
\newblock In {\em International Conference on Machine Learning}, pages
  3868--3877. PMLR.

\bibitem[Pavlo et~al., 2017]{pavlo2017self}
Pavlo, A., Angulo, G., Arulraj, J., Lin, H., Lin, J., Ma, L., Menon, P., Mowry,
  T.~C., Perron, M., Quah, I., et~al. (2017).
\newblock Self-driving database management systems.
\newblock In {\em CIDR}, volume~4, page~1.

\bibitem[Pike-Burke et~al., 2018]{pikeburke2018bandits}
Pike-Burke, C., Agrawal, S., Szepesvari, C., and Grunewalder, S. (2018).
\newblock Bandits with delayed, aggregated anonymous feedback.

\bibitem[Sen et~al., 2018]{sen18_mfdoo}
Sen, R., Kandasamy, K., and Shakkottai, S. (2018).
\newblock Multi-fidelity black-box optimization with hierarchical partitions.
\newblock In {\em International conference on machine learning}, pages
  4538--4547. PMLR.

\bibitem[Sen et~al., 2019]{sen19_MFHOO}
Sen, R., Kandasamy, K., and Shakkottai, S. (2019).
\newblock Noisy blackbox optimization using multi-fidelity queries: A tree
  search approach.
\newblock In {\em The 22nd international conference on artificial intelligence
  and statistics}, pages 2096--2105. PMLR.

\bibitem[Shang et~al., 2018]{Shang2018AdaptiveBO}
Shang, X., Kaufmann, E., and Valko, M. (2018).
\newblock Adaptive black-box optimization got easier: Hct only needs local
  smoothness.
\newblock In {\em EWRL 2018}.

\bibitem[Shang et~al., 2019]{poo}
Shang, X., Kaufmann, E., and Valko, M. (2019).
\newblock General parallel optimization without a metric.
\newblock In {\em Algorithmic Learning Theory}, pages 762--788. PMLR.

\bibitem[Springenberg et~al., 2016]{NIPS2016_a96d3afe}
Springenberg, J.~T., Klein, A., Falkner, S., and Hutter, F. (2016).
\newblock Bayesian optimization with robust bayesian neural networks.
\newblock In Lee, D., Sugiyama, M., Luxburg, U., Guyon, I., and Garnett, R.,
  editors, {\em Advances in Neural Information Processing Systems}, volume~29.
  Curran Associates, Inc.

\bibitem[Sra et~al., 2015]{sra2015adadelay}
Sra, S., Yu, A.~W., Li, M., and Smola, A.~J. (2015).
\newblock Adadelay: Delay adaptive distributed stochastic convex optimization.
\newblock {\em arXiv preprint arXiv:1508.05003}.

\bibitem[Srinivas et~al., 2010]{gpucb}
Srinivas, N., Krause, A., Kakade, S., and Seeger, M. (2010).
\newblock Gaussian process optimization in the bandit setting: No regret and
  experimental design.
\newblock In {\em Proceedings of the 27th International Conference on
  International Conference on Machine Learning}, ICML'10, page 1015–1022,
  Madison, WI, USA. Omnipress.

\bibitem[van~der Vlerk, 1996]{hartmann3}
van~der Vlerk, M.~H. (1996).
\newblock Stochastic programming bibliography.
\newblock {\em World Wide Web, http://mally. eco. rug. nl/spbib. html}, 2003.

\bibitem[Vernade et~al., 2017]{vernade2017stochastic}
Vernade, C., Capp{\'e}, O., and Perchet, V. (2017).
\newblock Stochastic bandit models for delayed conversions.
\newblock {\em arXiv preprint arXiv:1706.09186}.

\bibitem[Wang et~al., 2021]{Wang2021}
Wang, J., Trummer, I., and Basu, D. (2021).
\newblock {UDO: universal database optimization using reinforcement learning}.
\newblock In {\em arXiv:2104.01744}, pages 1--13.

\bibitem[Wang et~al., 2019a]{wang2019alphax}
Wang, L., Zhao, Y., Jinnai, Y., Tian, Y., and Fonseca, R. (2019a).
\newblock Alphax: exploring neural architectures with deep neural networks and
  monte carlo tree search.
\newblock {\em arXiv preprint arXiv:1903.11059}.

\bibitem[Wang et~al., 2019b]{wang2019optimization}
Wang, Y., Balakrishnan, S., and Singh, A. (2019b).
\newblock Optimization of smooth functions with noisy observations: Local
  minimax rates.
\newblock {\em IEEE Transactions on Information Theory}, 65(11):7350--7366.

\bibitem[Weinberger and Ordentlich, 2002]{weinberger2002delayed}
Weinberger, M.~J. and Ordentlich, E. (2002).
\newblock On delayed prediction of individual sequences.
\newblock {\em IEEE Transactions on Information Theory}, 48(7):1959--1976.

\bibitem[Xiong et~al., 2013]{borehole}
Xiong, S., Qian, P.~Z., and Wu, C.~J. (2013).
\newblock Sequential design and analysis of high-accuracy and low-accuracy
  computer codes.
\newblock {\em Technometrics}, 55(1):37--46.

\bibitem[Xu et~al., 2020]{xu2020zeroth}
Xu, Y., Joshi, A., Singh, A., and Dubrawski, A. (2020).
\newblock Zeroth order non-convex optimization with dueling-choice bandits.
\newblock In {\em Conference on Uncertainty in Artificial Intelligence}, pages
  899--908. PMLR.

\bibitem[Xue et~al., 2016]{xue2016accelerated}
Xue, D., Balachandran, P.~V., Hogden, J., Theiler, J., Xue, D., and Lookman, T.
  (2016).
\newblock Accelerated search for materials with targeted properties by adaptive
  design.
\newblock {\em Nature communications}, 7(1):1--9.

\end{thebibliography}
\newpage
\appendix
\part{Appendix}
\parttoc
\section{Additional Background Details}
\subsection{Regret}
Typically, in a multi-armed bandit problem~\citep{lattimore2020bandit}, an algorithm encounters $K$ unknown probability distributions of rewards. The algorithm can only known more about it by sampling the distribution (or often referred as \textit{arm}). Now, the goal of a bandit algorithm is to maximise total sum of accumulated rewards, i.e. $\sum_{t=1}^T R_t$, given a time horizon $T$. 
Typically, what we aim to maximize is the expectation of accumulated rewards, i.e. $\expect[\sum_{t=1}^T R_t]$. 

If we want to maximize the value of $f$ for a point $x$ in the subdomain covered by a node $(h,i)$, it becomes equivalent to maximising the total obtained reward through its children. Thus, multi-armed bandit algorithms are deployed in HOO in order to maximize the value of an objective function, while the tree-based partition is given. 

There is an alternative way of formulating the goal of a bandit, i.e. minimizing deviation of the expected accumulated reward $\expect[\sum_{t=1}^T R_t]$ from the maximal achievable reward $T f^*$. This is called \textit{expected cumulative regret} or simply \textit{regret}.
\begin{align*}
        \expect[\reg_T]
        &= T \mu^* - \sum_{a=1}^K \mathbb{E}_{\pi}\left[N^a_T\right] \mu_a\\
        &= \sum_{a=1}^K \mathbb{E}_{\pi}\left[N^a_T\right] (\mu^* - \mu_a), \qquad \text{since, }T = \sum_{a=1}^K \mathbb{E}_{\pi}\left[N^a_T\right]. 
\end{align*}
Here, $K$ is the number of arms, $\mathbb{E}_{\pi}\left[N^a_T\right]$ is the expected number of times the arm $a$ is drawn, and $\mu^*-\mu_a$ is the expected suboptimality of arm $a$.
Following the traditional analysis of hierarchical tree search algorithms~\citep{mcts}, we are going to use regret as the measure of performance.
Less is the regret better is the performance of the algorithm. If the upper bound on regret grows sublinearly with horizon $T$, it means that the error incurred by corresponding algorithm asymptotically vanishes. 

The other performance metric relevant to  a black-box optimization algorithm is the \textit{error} incurred at any time $t$: 
\begin{equation}
    r_t = \max_x f(x) - f(x_t) = f^* - f(x_t).
\end{equation}
$r_t$ is also termed as \textit{simple regret} in bandit literature. This metric is different than regret but in case of HOO, their expected values are closely related.
If we choose the state $x_t$ uniformly randomly from all the states observed till time $T$, the expected value of error (or simple regret) becomes 
\begin{equation}\label{eq:two_regrets}
    \expect[r_t] = \expect[f^* - f(x_t)] = \frac{1}{T}\sum_{i=1}^t[f^* - f(x_i)] = \frac{1}{T} \reg_T
\end{equation}
Later, we will use this relation to convert the traditional regret bound originating from the bandit algorithms used in \framework{} to the expected error or the expected simple regret.

\subsection{Optimistic Algorithms for Multi-armed Bandits}
In both finite and continuous armed bandits, one of the successful paradigm is to design algorithms with Optimism in Front of Uncertainty (OFU) principle. OFU-type of algorithms, such as Upper Confidence Bound (UCB), UCB-$\sigma$, UCB-V etc., at each step $t$ computes an optimistic value $B_{i,t}$ for each arm $i \in \mathcal{A}$ and chooses the arm with the maximum optimistic value.
This optimistic value depends on the mean reward obtained from the arm, the number of pulls on the arm, and sometimes other carefully chosen statistics (e.g. variance and range) of rewards generated by the arm. 

For example, for UCB~\cite{Auer2002}, $B_{i,t}^{\mathrm{UCB}} \triangleq \hat{\mu}_{i,t} + \sqrt{\frac{2\log t}{T_i(t-1)}}$. Here, $T_i(t-1)$ is the number of times arm $i$ is played till time step $t$ and $\hat{\mu}_{i,t}$ is the empirical mean of rewards obtained from arm $i$ till time $t$. UCB does not consider any noise added to the rewards. 

If the reward arrives with a noise of known variance $\sigma^2$, the UCB bound can be adapted to create UCB-$\sigma$~\cite{auer2007improved}. For UCB-$\sigma$, $B_{i,t}^{\mathrm{UCB}-\sigma} \triangleq \hat{\mu}_{i,t} + \sqrt{\frac{2\sigma^2\log t}{T_i(t-1)}}$. Since in most of the cases the noise variance $\sigma^2$ is not known, UCB-V is designed to adapt for noise with unknown variance and bounded range $[0,b]$. In this case, the noise variance for arm $i$ is estimated at each step $t$ as $\hat{\sigma}_{i,t}^2 \triangleq \sum_{t=1}^T \mathds{1}[a_t = i](f(x_t) - \hat{\mu}^i_t)^2$. Following that, the effective noise variance is used to define the optimistic value of UCB-V~\cite{ucbv} as $B_{i,t}^{\mathrm{UCB-V}} \triangleq \hat{\mu}_{i,t} + \sqrt{\frac{2\hat{\sigma}_{i,t}^2\log t}{T_i(t-1)}} + c\frac{3b\log t}{T_i(t-1)}$. Here, $c>0$ is some tunable parameter, which we consider $1$ throughout our analysis. Interested practitioner may like to experiment with it.

The triumph of the OFU-type of algorithms is due to the fact that they achieve $O(\log T)$ regret bounds after $T$ time steps, which is the optimal achievable regret according to the problem-dependent lower bound on regret of stochastic bandits~\cite[Chapter 16]{lattimore2020bandit}. For further details on OFU algorithms, we refer to~\cite[Chapters 7-10]{lattimore2020bandit}. 

In last decade, the bandit community has been interested to adapt these OFU-type algorithms to delayed setup. \cite{Joulani2013a,joulani2016delay,vernade2017stochastic,pikeburke2018bandits,li2019bandit} have extended some OFU-type algorithms, such as UCB and KL-UCB, to different delayed feedback settings, such as observable delay, arm-dependent delay etc.
\newpage
\section{Proof Details}
\subsection{Assumptions: Structural Requirements of Optimistic Tree Search}
In order to proof convergence of UDO, we oblige by the assumptions regarding theoretical analysis of HOO~\citep{bubeck2011a}. In this section, we elaborate them.


\paragraph{Contracting Hierarchical Partition $\tree$.} The hierarchical optimistic tree search (HOO) or $\domain$-armed bandits rely on existence of a hierarchical partitioning $\tree$ of the domain $\domain$. Let us represent the interval covered by the $l$-th node at depth $h$ as $X_{h,i}$, where $l \in \setof{1,\ldots,2^h}$ and $h \in \setof{0,\ldots,H}$. Then, we can define the corresponding hierarchical tree inducing the partition as $\tree \triangleq \setof{X_{h,l}}_{h,l=0,1}^{H, 2^h}$.
We observe that
\begin{align*}
    X_{(0,1)} &= \domain,\\
    X_{(h,l)} &= \cup_{j=0}^{K-1} X_{(h+1,Kl-j)} \cup X_{(h+1,Kl+j)},
\end{align*}
where $K$ is the maximum number of children of a node in this tree.

The specific value obtained at that node is denoted as $x_{h,i}$. Let us also assume that the domain of $f$, say $\domain \subset \real^D$, has a dissimilarity measure or semi-metric $\ell$ that can quantify difference in output due to two inputs. 

\begin{repassumption}{ass:partition}[Hierarchical Partition with Decreasing Diameter and Shape]\
\begin{enumerate}
    \item[1.1] \emph{Decreasing diameters.} There exists a decreasing sequence $\delta(h) > 0$ and constant $\nu_1 > 0$ such that
    \begin{equation}
        \mathrm{diam}(X_{h,i}) \triangleq \max_{x \in X_{h,i}} \ell(x_{h,i}, x) \leq \nu_1 \delta(h),\label{assum:1.1}
    \end{equation}
    for any depth $h \geq 0$, for any interval $X_{h,i}$, and for all $i=1,\ldots,2^h.$ For simplicity, we consider that $\delta(h) = \rho^h$ for some $\rho \in (0,1)$.
    \item[1.2] \emph{Regularity of the intervals.} There exists a constant $\nu_2>0$ such that for any depth $h\geq 0$, every interval $X_{h,i}$ contains at least a ball $\ball_{h,i}$ of radius $\nu_2 \rho^h$ and center $x_{h,i}$ in it. Since the tree creates a partition at any given depth $h$, $\ball_{h,i} \cap \ball_{h,i} = \emptyset$ for all $1\leq i < j \leq 2^h$.
\end{enumerate}
\end{repassumption}

\paragraph{Global Smoothness of $f$.} The other condition that we need to prove convergence of HOO to a global optimum is smoothness of $f$ around the optimum, say $x^*$. This is often referred as weak Lipschitz property.
\begin{repassumption}{ass:lipschitz}[Weak Lipschitzness of $f$]
For all $x, y \in \domain$, $f$ satisfies
\begin{equation}
    f^* - f(y) \leq f^* - f(x) + \max\{ f^* - f(x), \ell(x,y)\},
\end{equation}
where $f^*$ is the optimal value of $f$ achieved at a global optimum $x^*$.
\end{repassumption}
This assumption holds true if
\begin{enumerate}[leftmargin=*]
    \item either $f(x) - f(y) \leq \ell(x,y)$ and $f^* - f(x) \leq \max_y \ell(x,y)$,
    \item or $f(x) - f(y) \leq f^* - f(x)$ and $f^* - f(x) \geq \max_y \ell(x,y)$.
\end{enumerate}
This property basically implies that there is no sudden drop or jump in performance $f$ around the optimal point $x^*$.
Weak Lipschitzness can hold even for discontinuous functions.
Thus, it widens applicability of HOO's analysis to more general performance metrics and configuration spaces in comparison with algorithms that explicitly need gradients or smoothness in some form.

\newpage
\subsection{Regret of \framework{} with Delayed Feedback}
Let us define a few quantities before proceeding to the proofs. 
\begin{enumerate}
    \item $G_t = \sum_{s=1}^{t-1} \mathds{1}\lbrace s + \tau_s \geq t\rbrace$ i.e. the number of missing feedbacks when the forecaster chooses the next action at time $t$.
    \item $G_t^* = \max_{1\leq s\leq t} G_t$ the maximum number of feedbacks not observed till time $t$. Note that it is $\tau_{const}$ for constant delay.
    \item $G_{i,t}$ which is the number of missing feedbacks for action $i$ at time $t$.
    \item $T_i(t)$ which is the number of reward samples observed from arm $i$ at time $t$ in a non-delayed setting.
    \item $S_i(t)$ which is the number of reward samples observed from arm $i$ at time $t$ in a delayed setting. Note that $T_i(t) =  S_i(t) + G_{i,t}$.
\end{enumerate}
On the other hand, we can describe the action selection process of any UCB-like optimistic bandit algorithm as:
\begin{align*}
    a_t = \argmax_{i \in \mathcal{A}} B_{i,s,t}.
\end{align*}
Here, $B_{i,s,t}$ is the optimistic upper confidence bound for action $i$ at time $t$, and $s$ is the number of reward samples obtained from arm $i$ till time $t$. For example, in case of UCB1~\citep{Auer2002}, 
\begin{equation}\label{eqn:ucb1}
    B_{i,s,t} = \hat{\mu}_{i,s} + \sqrt{\frac{2 \log t}{s}} = \frac{1}{s}\sum_{j=1}^s r_{i,j} + \sqrt{\frac{2 \log t}{s}}.
\end{equation}
For non-delayed setting, $s= T_i(t-1)$ and for delayed setting $s=S_i(t-1)$.

Thus, given an aforementioned UCB-like optimistic bandit algorithm, the leaf node $(h_t, j_t)$ selected by \framework{} algorithm at time $t$ is
\begin{align}
     (h_t, l_t) &\triangleq \argmax_{(h,l) \in \tree_t} B^{\min}_{(h,l)}(t) \triangleq \argmax_{(h,l) \in \tree_t} \min\lbrace B_{(h,l),s,t} + \nu_1\rho^h, \max_{(h',l')\in C(h,l)} B^{\min}_{(h,l)}(t)\rbrace.
\end{align}
Given these notations and definitions, now we elaborate the proof sketch of \framework{} following that of HOO and show the generic technique to incorporate regret in it. The analysis described in this section, we assume that the smoothness parameters $\nu_1$ and $\rho$ are known. We loosen this assumption in the following section.

\subsubsection{Generic Proof Sketch: From Regret to the Number of Visits to a Suboptimal Node}
\begin{theorem}\label{thm:regret_decomp}
Let us consider that the expected objective function $f$ satisfies Assumption~\ref{ass:lipschitz}, and its $4\nu_1/\nu_2$-near-optimality dimension is $d>0$. Then, under Assumption~\ref{ass:partition} and for any $d'>d$, \framework{} algorithm uses a bandit algorithm MAB for node selection will achieve expected regret
\begin{align}
    \expect[\reg_T] &\leq 4 C \nu_1\nu_2^{-d'} \sum_{h=0}^{H-1} (\delta(h))^{(1-d')} + 4 \nu_1 \delta(H) T\notag\\
    &+ 8C \nu_1\nu_2^{-d'} \sum_{h=1}^H (\delta(h-1))^{(1-d')} \times (U(T,\tau,\delta(h))).
\end{align}
Here, $U(T,\tau,\delta(h))$ is the upper bound on number of visits to the $2\nu_1\delta(h)$-suboptimal nodes at depth $h>0$ by \bandit.
\end{theorem}
\begin{proof}
Now, we proceed with the regret analysis of \framework{} which is essentially similar with the analysis of the HOO while we try to accommodate the delayed feedbacks in it. The proof can be divided into three steps: a) regret decomposition to suboptimal and optimal node exploration, b) bounding the number of times the suboptimal nodes and the optimal nodes reached through the suboptimal nodes, c) bringing in the effect of delay and UCB-like algorithms in above too and merging them.

\paragraph{Step 1: Regret Decomposition.} Let us divide the nodes of the Monte Carlo tree $\tree$, which can possibly grow infinitely, in three categories such that $\tree = \tree_1 \cup \tree_2 \cup \tree_3$.

In order to define, these subcategories first let us define the `optimal' nodes $I$ and `suboptimal' nodes reached by playing optimal nodes $J$.

Let us denote the set of $2\nu_1\delta(h)$-optimal nodes at depth $h$ ($h \in \mathbb{Z}_{\geq 0}$) as $I_h$, i.e.
$$ I_{h} \triangleq \lbrace  (h,i) | f^*_{h,i} \geq f^* - 2\nu_1\delta(h) \rbrace.$$
Thus, the set of all optimal nodes in the MC-tree is $I = \bigcup_h I_h$. Note that, the root node has $(h,i)=(0,1)$ and $I_0 = \{(0,1)\}$.

Let us denote the nodes $J = \bigcup_h J_h$ who are not essentially in $I$  but whose parents are in $I$. These are the suboptimal nodes to be reached through playing the optimal nodes. As the root node is in $I_0$, we can observe that all the $2\nu_1\delta(h)$-suboptimal nodes (for all $h>0$) in the tree are children of either nodes in $I$ or those in $J$.

Given these definitions, we can define now the three subcategories of nodes that we mentioned earlier.
\begin{enumerate}
    \item[$\tree_1$:] Set of all optimal nodes in the tree $I_H = \bigcup_{h=0}^H I_h$.
    \item[$\tree_2$:] All the nodes which are descendants\footnote{Descendants of a node include the node itself.} of nodes in $I_H$.
    \item[$\tree_3$:] All the nodes which are descendants of nodes in $J_H = \bigcup_{h=1}^H J_h$.
\end{enumerate}
Now, we decompose the expected regret in three components corresponding to each of these three categories:
\begin{align}\label{eq:reg_Decomp}
    \expect[\reg_T] &= \expect[\sum_{t=1}^T (f^* - f(X_t))] \notag\\ 
    &= \expect[\sum_{t=1}^T (f^* - f(X_t))]\mathds{1}[(H_t, I_t) \in \tree_1] + \expect[\sum_{t=1}^T (f^* - f(X_t))]\mathds{1}[(H_t, I_t) \in \tree_2] \notag\\
    &\qquad\qquad\qquad\qquad\qquad\qquad\qquad\qquad+ \expect[\sum_{t=1}^T (f^* - f(X_t))]\mathds{1}[(H_t, I_t) \in \tree_3]\notag\\
    &\triangleq \expect[\reg_{T,1}] + \expect[\reg_{T,2}] + \expect[\reg_{T,3}]
\end{align}

By Lemma 3 in~\citep{bubeck2011a}, we obtain that
\begin{align}\label{eq:reg1}
    \expect[\reg_{T,1}] \leq \sum_{h=0}^{H-1} 4 \nu_1 \delta(h) |I_h|.
\end{align}
and, 
\begin{align}\label{eq:reg2}
    \expect[\reg_{T,2}] \leq 4 \nu_1 \delta(H) T.
\end{align}

Now, for $\tree_3$, we observe that parent of any element in $J_h$ is in $I_{h-1}$. Thus, the region covered by these nodes is a subset of the region covered by $I_{h-1}$ (Assumption~\ref{ass:partition} and~\ref{ass:lipschitz}). Thus, we obtain
\begin{align}\label{eq:reg3}
    \expect[\reg_{T,3}] &\leq \sum_{h=1}^H 4 \nu_1 \delta(h-1) \sum_{i:(h,i) \in J_h}  \expect[S_{h,i}(T)]\\ 
    &\leq \sum_{h=1}^H 4 \nu_1 \delta(h-1) |J_h| \max_{(h,i) \in J_h} \expect[S_{h,i}(T)]\\
    &\leq \sum_{h=1}^H 8 \nu_1 \delta(h-1) |I_{h-1}| \max_{(h,i) \in J_h} \expect[S_{h,i}(T)].
\end{align}
The last inequality holds as the parents of nodes of $J_h$ are in $I_{h-1}$, and by the way the tree grows $|J_h| \leq 2 |I_{h-1}|$ for any $h \geq 1$. 
\paragraph{Step 2: Bounding the Number of Optimal Nodes and Direct Descendants of Optimal Nodes.}
Now, we want to bound two things. Firstly, the size of the optimal nodes at level $h$ of the MC-tree, i.e. $|I_h|$. Secondly, the expected number of times the nodes in $J_h$ are visited, i.e. $\expect[S_{h,i}(t) | (h,i) \in J_h]$.

From~\citep{bubeck2011a}, we observe that bounding the first quantity, i.e. $|I_h|$, is independent of the MC-tree dynamics under Assumption~\ref{ass:partition}. Thus, we get
\begin{align}\label{eq:subopt_node}
    |I_h| \leq C(\nu_2 \delta(h))^{-d'}~~\text{for some}~d' \geq d.
\end{align}

The second quantity depends on the UCB-like algorithm used for next action/node selection and also the delay model. In all these cases, we can show that
\begin{align}
    \expect[S_{h,i}(t) | (h,i) \in J_h] \leq U(t,\tau,\delta(h)).
\end{align}
In corresponding sections, we prove the specific forms of $U(t,\tau,\delta(h))$ for UCB-like algorithms and delay models.

\paragraph{Step 3: Merging the Effects of Delay and Suboptimal Node Plays.}
If we merge all the aforementioned results, we obtain an upper bound on the total regret of the \framework{} framework.
\begin{align}
    \expect[\reg_T] &\leq \sum_{h=0}^{H-1} 4 \nu_1 \delta(h) |I_h| + 4 \nu_1 \delta(H) T + \sum_{h=1}^H 8 \nu_1 \delta(h-1) |I_{h-1}| \max_{(h,i) \in J_h} \expect[S_{h,i}(t)] \notag\\
    &\leq \sum_{h=0}^{H-1} 4 \nu_1 \delta(h) \times C(\nu_2 \delta(h))^{-d'} \notag\\
    &+ 4 \nu_1 \delta(H) T\notag\\
    &+ \sum_{h=1}^H 8 \nu_1 \delta(h-1) \times C(\nu_2 \delta(h-1))^{-d'} \times (U(T,\tau,\delta(h)))\notag\\
    &\leq 4 C \nu_1\nu_2^{-d'} \sum_{h=0}^{H-1} (\delta(h))^{(1-d')} + 4 \nu_1 \delta(H) T\notag\\
    &+ 8C \nu_1\nu_2^{-d'} \sum_{h=1}^H (\delta(h-1))^{(1-d')} \times (U(T,\tau,\delta(h))).
\end{align}
Thus, we get the desired regret bound for \framework{} with observable stochastic delays in feedbacks and \framework{} algorithm with any efficient bandit algorithm MAB for node selection.
\end{proof}

\subsubsection{Expected Visits of a Node $(h,i)$ under Delay}
\begin{lemma}\label{lemma:visits}
Let us consider $(h,i)$ is a $2\nu_1\delta(h)$-suboptimal node in $J_h$ and there's an observable stochastic delay $G_{(h,i),t}$ while receiving the feedback. Then for any integer $u\geq 0$, 
\begin{align}
    \expect[S_{h,i}(T)] \leq u + \expect[G_{(h,i),T}^*] + \sum_{t=u+1}^T ( &\prob[S_{h,i}(t) > u \wedge B_{(h,i),S_{h,i},t} + \nu_1 \delta(h)> f^*]\notag \\ &+ \sum_{s=1}^t \prob[B_{(h,i),S_{h,i},t} + \nu_1 \delta(h) \leq f^*]  ).
\end{align}
\end{lemma}
\begin{proof}
For the undelayed feedback, the number of times a $2\nu_1\delta(h)$-suboptimal node $(h,i)$ is visited can be written as
\begin{align*}
    T_{h,i}(T) &= \sum_{t=1}^T \mathds{1}[(H_t, I_t) \in \mathrm{Descendant}(h,i) \wedge T_{h,i}(t) \leq u] \\
    &+ \sum_{t=1}^T \mathds{1}[(H_t, I_t) \in \mathrm{Descendant}(h,i) \wedge T_{h,i}(t) > u]\\
    &\leq u + \sum_{t=u+1}^T \mathds{1}[(H_t, I_t) \in \mathrm{Descendant}(h,i) \wedge T_{h,i}(t) > u].
\end{align*}
Now, to shift this analysis to the observable stochastic delayed feedback model, we replace $T_{h,i}(t)$ with $S_{h,i}(t) + G_{h,i}(t)$ and $u$ with $u'+G_{(h,i),T}^*$. Thus, we get
\begin{align*}
    S_{h,i}(T) \leq u' + G_{(h,i),T}^* + \sum_{t=u'+1}^T \mathds{1}[\underset{\text{Event}~ E_1}{\underbrace{(H_t, I_t) \in \mathrm{Descendant}(h,i)}} \wedge S_{h,i}(t) > u'].
\end{align*}

Now, in order to understand the event $E_1$, let us consider that the path from the root node $(0,1)$ to node $(h,i)$ passes through an optimal node $(k, i^*_k)$ last at depth $k \in [0,h-1]$ and does not visit an optimal node afterwards. Under this specification, the path toward the suboptimal node $(h,i)$ is chosen than the next optimal node $(k+1, i^*_{k+1})$ if $$B_{(h,i),S_{h,i},t} + \nu_1 \delta(h) \geq B_{(k+1,i^*_{k+1}),S_{k+1,i^*_{k+1}},t}.$$
This event is satisfied if $E_2: B_{(h,i),S_{h,i},t} + \nu_1 \delta(h) \geq f^*$ and $E'_{k+1}: f^*\geq B_{(k+1,i^*_{k+1}),S_{k+1,i^*_{k+1}},t}$ hold true.
Thus, $E_1 \subset E_2 \cup E'_{k+1}$.

We also observe that $E'_{k+1} \subset E_{k+1} \cup E'_{k+2}$, where $E_{k+1}: B_{(k+1,i^*_{k+1}),S_{(k+1,i^*_{k+1})},t} + \nu_1 \delta(k+1) \leq f^*$ and $E'_{k+2}: f^*\geq B_{(k+2,i^*_{k+1}),S_{k+1,i^*_{k+1}},t}$. Iterating similar argument from $k+1$ to $t$, we get $E'_{k+1} \subset \cup_{s=k+1}^{t-1} E_{s}$. The induction stops at $t-1$ because the node $(t,i^*_t)$ is yet not visited and thus, the upper confidence value assigned to it is $+\infty$, which is not bounded by $f^*$ for sure.

Hence, $E_1 \subset E_2 \cup (\cup_{s=k+1}^{t-1} E_{s})$, and number of visits to node $(h,i)$ under delayed feedback
\begin{align*}
    S_{h,i}(T) &\leq u' + G_{(h,i),T}^* + \sum_{t=u'+1}^T \mathds{1}[(E_2 \cup (\cup_{s=k+1}^{t-1} E_{s}))  \wedge S_{h,i}(t) > u']\\
    &\leq u' + G_{(h,i),T}^* + \sum_{t=u'+1}^T \mathds{1}[(E_2 \wedge S_{h,i}(t) > u') \cup (\cup_{s=k+1}^{t-1} E_{s})].
\end{align*}
Thus, 
\begin{align*}
    \expect[S_{h,i}(T)] &~~~~~\leq u' + \expect[G_{(h,i),T}^*] + \sum_{t=u'+1}^T \prob[(E_2 \wedge S_{h,i}(t) > u') \cup (\cup_{s=k+1}^{t-1} E_{s})]\\
    &\underset{Union~bound}{\leq}  u' + \expect[G_{(h,i),T}^*] + \sum_{t=u'+1}^T \left(\prob[E_2 \wedge S_{h,i}(t) > u'] + \sum_{s=k+1}^{t-1} \prob[E_{s}]\right)\\
    &~~~~~~\leq  u' + \expect[G_{(h,i),T}^*] + \sum_{t=u'+1}^T \left(\prob[E_2 \wedge S_{h,i}(t) > u'] + \sum_{s=1}^{t-1} \prob[E_{s}]\right).
\end{align*}
\end{proof}

\subsubsection{Specification for Delayed-UCB1}
Up to this point the analysis is independent of the choice of the bandit algorithm for node selection. For an example, let us choose the bandit algorithm to be Delayed-UCB1~\citep{Joulani2013a}.
From Delayed-UCB1, we obtain that
\begin{align}\label{eq:ducb1}
    B_{(h,i),S_{(h,i)}(t-1),t} &\triangleq \hat{\mu}_{(h,i),S_{(h,i)}(t-1)} + \sqrt{\frac{2 \log t}{S_{(h,i)}(t-1)}}
\end{align}
Under this decision rule, we prove the upper bound on the expected number of visits to any $2\nu_1\delta(h)$-suboptimal node $(h,i)$.

\begin{corollary}\label{cor:ducb1_visits}
Let us consider $(h,i)$ is a $2\nu_1\delta(h)$-suboptimal node in $J_h$ and there's an observable stochastic delay $G_{(h,i),t}$ in feedback with expectation bound $\tau$. If we use Delayed-UCB1 for node selection at any time $t$, we obtain that 
\begin{align}
    U(T, \tau, \delta(h)) = \frac{8 \ln T}{(\nu_1 \delta(h))^2} + \tau + 4.
\end{align}
\end{corollary}
\begin{proof}
Now, if we choose $u' = \frac{8 \ln T}{(\nu_1 \delta(h))^2}$, using Hoeffding's concentration inequalities, and Assumptions~\ref{ass:partition} and~\ref{ass:lipschitz}, we get from Lemma~\ref{lemma:visits}
\begin{align*}
    \expect[S_{h,i}(T)] &\leq \frac{8 \ln T}{(\nu_1 \delta(h))^2} + \expect[G_{(h,i),T}^*] + \sum_{t=u'+1}^T \left(\prob[E_2 \wedge S_{h,i}(t) > u'] + \sum_{s=1}^{t-1} \prob[E_{s}]\right)\\
    &\leq \frac{8 \ln T}{(\nu_1 \delta(h))^2} + \expect[G_{(h,i),T}^*] + \sum_{t=u'+1}^T \left(t T^{-4} + \sum_{s=1}^{t-1} t^{-3}\right)\\
    &\leq \frac{8 \ln T}{(\nu_1 \delta(h))^2} + \expect[G_{(h,i),T}^*] + 4\\
    &\leq \frac{8 \ln T}{(\nu_1 \delta(h))^2} + \tau + 4.
\end{align*}
\end{proof}

\begin{reptheorem}{thm:regret_delay}[Regret of \frameworkucb{}]
Let us consider that the expected objective function $f$ satisfies Assumption~\ref{ass:lipschitz}, and its $4\nu_1/\nu_2$-near-optimality dimension is $d>0$. Then, under Assumption~\ref{ass:partition} and for any $d'>d$, \framework{} algorithm using Delayed-UCB1 (\ducb{}) achieves expected regret
\begin{align}
    \expect[\reg_T]
    &= \bigO\left(T^{1-\frac{1}{d'+2}}(\ln T)^{\frac{1}{d'+2}} \left(1+ \frac{\tau}{\ln T}\right)^{\frac{1}{d'+2}}\right),
\end{align}
and expected simple regret
\begin{align}
    \epsilon_T
    &= \bigO\left(\left(\frac{\ln T}{T}\right)^{\frac{1}{d'+2}} \left(1+ \frac{\tau}{\ln T}\right)^{\frac{1}{d'+2}}\right),
\end{align}
where the expected delay is upper bounded by $\tau$.
\end{reptheorem}
\begin{proof}
From Theorem~\ref{thm:regret_decomp} and Corollary~\ref{cor:ducb1_visits}, we get
\begin{align*}
    \expect[\reg_T] &\leq 4 C \nu_1\nu_2^{-d'} \sum_{h=0}^{H-1} (\delta(h))^{(1-d')} + 4 \nu_1 \delta(H) T \\
    &+ 8C \nu_1\nu_2^{-d'} \sum_{h=1}^H (\delta(h-1))^{(1-d')} \times \left(\frac{8 \ln T}{(\nu_1 \delta(h))^2} + \tau + 4\right)\\
    &\leq 4 C \nu_1\nu_2^{-d'} \sum_{h=0}^{H-1} (\delta(h))^{(1-d')} (2\tau + 9) + 4 \nu_1 \delta(H) T\\
    &+ 64C \nu_1^{-1}\nu_2^{-d'} \sum_{h=1}^H (\delta(h-1))^{(1-d')}\delta(h)^{-2} \times \ln T.
\end{align*}
For simplicity, we represent the decreasing diameter as $\delta(h) = \rho^h$ for some $\rho \in (0,1)$. Thus, 
\begin{align}
    \expect[\reg_T] &\leq 4 C \nu_1\nu_2^{-d'} \sum_{h=0}^{H-1} \rho^{h(1-d')} (2\tau + 9) + 4 \nu_1 \rho^H T\notag\\
    &+ 64C \nu_1^{-1}\nu_2^{-d'} \sum_{h=1}^H \rho^{h(1-d')- (1-d')}\rho^{-2h} \times \ln T\notag\\
    &\leq 4 C \nu_1\nu_2^{-d'} \sum_{h=0}^{H-1} \rho^{h(1-d')} (2\tau + 9) + 4 \nu_1 \rho^H T\notag\\
    &+ 64C \nu_1^{-1}\nu_2^{-d'} \sum_{h=1}^H \rho^{-h(1+d')- (1-d')} \times \ln T\notag\\
    &= \bigO(\sum_{h=0}^{H-1} \rho^{h(1-d')} \tau + \sum_{h=0}^{H-1} \rho^{-h(1+d')} \ln T + \rho^H T)\label{eq:40}
\end{align}
Since $\rho \in (0,1)$, we get
\begin{align*}
    \expect[\reg_T] &= \bigO(\sum_{h=0}^{H-1} \rho^{-h(1+d')} (\tau + \ln T) + \rho^H T) \\
     &= \bigO(\rho^{-H(1+d')} (\tau + \ln T) + \rho^H T)
\end{align*}
By choosing a $\rho$ such that $\rho^{-H(d'+2)} = \frac{T}{\tau+\ln T}$, we obtain
\begin{align*}
    \expect[\reg_T] &= \bigO\left(\left(\frac{T}{\tau+\ln T}\right)^{\frac{d'+1}{d'+2}} (\tau + \ln T) + \left(\frac{\tau+\ln T}{T}\right)^{\frac{1}{d'+2}} T\right)\\
    &= \bigO\left(T^{1-\frac{1}{d'+2}}(\tau+\ln T)^{\frac{1}{d'+2}}\right)\\
    &= \bigO\left(T^{1-\frac{1}{d'+2}}(\ln T)^{\frac{1}{d'+2}} \left(1+ \frac{\tau}{\ln T}\right)^{\frac{1}{d'+2}}\right).
\end{align*}
This trivially leads to the bound on simple regret, i.e. the expected error at each step, as $\epsilon_T = \frac{1}{T}\expect[\reg_T]$.
\end{proof}

\begin{repcorollary}{cor:const_delay}[Regret of \frameworkucb{} under Constant Delay]
If the delay in feedback is constant, i.e. $\tau_{const} > 0$, the expected regret of the \framework{} algorithm using Delayed-UCB1 (\ducb)
\begin{align}
    \expect[\reg_T]
    &= \bigO\left(T^{1-\frac{1}{d'+2}}(\ln T)^{\frac{1}{d'+2}} \left(1+ \frac{\tau_{const}}{\ln T}\right)^{\frac{1}{d'+2}}\right),
\end{align}
and expected simple regret
\begin{align}
    \epsilon_T
    &= \bigO\left(\left(\frac{\ln T}{T}\right)^{\frac{1}{d'+2}} \left(1+ \frac{\tau_{const}}{\ln T}\right)^{\frac{1}{d'+2}}\right).
\end{align}
\end{repcorollary}
\begin{proof}
We don't need an upper bound on expected maximum delay in this setting. Since all the delays are $\tau_{const}$, by default the expectation is exactly $\tau_{const}$. Putting that in the derivation trivially provides us this results.
\end{proof}

\subsection{Regret of \framework{} with Delayed and Noisy Feedback}
Typically, when we evaluate the objective function at any time step $t$, we obtain a noisy version of the function as feedback such that $\Tilde{f}(X_t) = f(X_t) + \epsilon_t$. Here, $\epsilon_t$ is a noise sampled independently generated from a noise distribution $\mathcal{N}$. Till now, we did not explicitly consider the noise for the action selection step. In this section, we provide analysis for that for both known and unknown variance cases.
In both of the cases, we assume that the noise has bounded mean and variance. This holds for the present setup as any noisy evaluation can be clipped in the range of the evaluations where we optimise the objective functions. And, we know that a bounded random variable is sub-Gaussian with bounded mean and variance.

\subsubsection{Noise with Known Variance} 
Let us assume that the variance of the noise is known, say $\sigma^2 > 0$. In this case, the optimistic bounds are computed using a simple variant of delayed-UCB1, say delayed-UCB-$\sigma^2$ (Equation~\eqref{eq:ducb1}), where
\begin{align}
    B_{(h,i),S_{(h,i)}(t-1),t} &\triangleq \hat{\mu}_{(h,i),S_{(h,i)}(t-1)} + \sqrt{\frac{2 \sigma^2 \log t}{S_{(h,i)}(t-1)}}.
\end{align}
Here, $\hat{\mu}_{(h,i),S_{(h,i)}(t-1)}$ is the empirical mean computed using noisy evaluations.

Using confidence bound of Equation~\eqref{eq:noisy_ducb1}, we obtain a modified version of the Corollary~\ref{cor:ducb1_visits}.
\begin{corollary}\label{cor:noisy_ducb1_visits}
Let us consider $(h,i)$ is a $2\nu_1\delta(h)$-suboptimal node in $J_h$, there's an observable stochastic delay $G_{(h,i),t}$ in feedback with expectation bound $\tau$, and the variance of the added noise in evaluations is $\sigma^2$. If we use delayed-UCB-$\sigma^2$ (Equation~\eqref{eq:noisy_ducb1}) for node selection at any time $t$, we obtain that 
\begin{align}
    U(T, \tau, \delta(h)) = \frac{8 \sigma^2 \ln T}{(\nu_1 \delta(h))^2} + \tau + 4.
\end{align}
\end{corollary}
\begin{proof}
Now, if we choose $u' = \frac{8 \sigma^2 \ln T}{(\nu_1 \delta(h))^2}$, similarly using Hoeffding's concentration inequalities, and Assumptions~\ref{ass:partition} and~\ref{ass:lipschitz}, we get from Lemma~\ref{lemma:visits}
\begin{align*}
    \expect[S_{h,i}(T)] &\leq \frac{8 \sigma^2 \ln T}{(\nu_1 \delta(h))^2} + \expect[G_{(h,i),T}^*] + \sum_{t=u'+1}^T \left(\prob[E_2 \wedge S_{h,i}(t) > u'] + \sum_{s=1}^{t-1} \prob[E_{s}]\right)\\
    &\leq \frac{8 \sigma^2 \ln T}{(\nu_1 \delta(h))^2} + \expect[G_{(h,i),T}^*] + \sum_{t=u'+1}^T \left(t T^{-4} + \sum_{s=1}^{t-1} t^{-3}\right)\\
    &\leq \frac{8 \sigma^2 \ln T}{(\nu_1 \delta(h))^2} + \tau + 4.
\end{align*}
\end{proof}
\begin{reptheorem}{thm:regret_delay_noise_known}[Regret of \frameworkucbs{}]
Let us assume that the variance of the noise in evaluations is $\sigma^2$ and is available to the algorithm. Then, under the same assumptions as Theorem~\ref{thm:HOO} and upper bound on expected delay $\tau$, \framework{} using \ducbs{} achieves
expected simple regret
\begin{align}
    \epsilon_T
    &= \bigO\left(T^{-\frac{1}{d'+2}} \left((\sigma/\nu_1)^2 \ln T+ {\tau}{}\right)^{\frac{1}{d+2}}\right).
\end{align}
\end{reptheorem}
\begin{proof}
Using the result of Corollary~\ref{cor:noisy_ducb1_visits} in the proof schematic of Theorem~\ref{thm:regret_delay}, we obtain that \framework{} algorithm using Equation~\eqref{eq:noisy_ducb1} for node selection will achieve 
\begin{align*}
    \expect[\reg_T] 
     &= \bigO(\rho^{-H(1+d')} (\tau + (\sigma/\nu_1)^2\ln T) + \rho^H T)
\end{align*}
By choosing a $\rho$ such that $\rho^{-H(d'+2)} = \frac{T}{\tau+(\sigma/\nu_1)^2 \ln T}$, we obtain
\begin{align*}
    \expect[\reg_T] &= \bigO\left(\left(\frac{T}{\tau+(\sigma/\nu_1)^2 \ln T}\right)^{\frac{d'+1}{d'+2}} (\tau + (\sigma/\nu_1)^2 \ln T) + \left(\frac{\tau+(\sigma/\nu_1)^2 \ln T}{T}\right)^{\frac{1}{d'+2}} T\right)\\
    &= \bigO\left(T^{1-\frac{1}{d'+2}}(\tau +(\sigma/\nu_1)^2 \ln T)^{\frac{1}{d'+2}}\right)
\end{align*}

Thus, for \framework{} with noisy evaluations and confidence bounds of Equation~\eqref{eq:noisy_ducb1}, i.e. \ducbs, achieves a simple regret
\begin{align}
    \epsilon_T &= \bigO\left(T^{1-\frac{1}{d'+2}}(\tau +(\sigma/\nu_1)^2 \ln T)^{\frac{1}{d'+2}}\right)\notag
\end{align}
where the expected delay is upper bounded by $\tau$ and $\sigma^2$ is the noise variance.
\end{proof}

\subsubsection{Noise with Unknown Variance} 
If the variance of the noise is unknown, we have to estimate the noise variance empirically from evaluations. Given the evaluations $\Tilde{f}(X_1), \ldots, \Tilde{f}(X_T)$ and the delayed statistics $S_{(h,i)}(t-1)$ of node $(h,i)$, the empirical noise variance at time $t$ is
\begin{align*}
    \hat{\sigma}_{(h,i),S_{(h,i)}(t-1)}^2 \triangleq \frac{1}{S_{(h,i)}(t-1)} \sum_{j=1}^{S_{(h,i)}(t-1)} \left( \Tilde{f}(X_j)\mathds{1}[(H_j, I_j)=(h,i)] - \hat{\mu}_{(h,i),S_{(h,i)}(t-1)}\right)^2,
\end{align*}
where empirical mean $\hat{\mu}_{(h,i),S_{(h,i)}(t-1)} \triangleq \frac{1}{S_{(h,i)}(t-1)} \sum_{j=1}^{S_{(h,i)}(t-1)} \Tilde{f}(X_j)\mathds{1}[(H_j, I_j)=(h,i)]$.

Using the empirical mean and variance of functional evaluations for each node $(h,i)$, we now define a delayed-UCBV\footnote{UCB-V without delay is proposed in~\citep{ucbv}.} confidence bound:
\begin{align}
    B_{(h,i),S_{(h,i)}(t-1),t} &\triangleq \hat{\mu}_{(h,i),S_{(h,i)}(t-1)} + \sqrt{\frac{2 \hat{\sigma}_{(h,i),S_{(h,i)}(t-1)}^2 \log t}{S_{(h,i)}(t-1)}} + \frac{3b \log t}{S_{(h,i)}(t-1)}.
\end{align}

\begin{corollary}\label{cor:ducbv_visits}
Let us consider $(h,i)$ is a $2\nu_1\delta(h)$-suboptimal node in $J_h$, there's an observable stochastic delay $G_{(h,i),t}$ in feedback with expectation bound $\tau$, and the variance of the added noise in evaluations is $\sigma^2$. If we use Delayed-UCBV (Equation~\eqref{eq:ducbv}) for node selection at any time $t$, we obtain that 
\begin{align}
    U(T, \tau, \delta(h)) = c\left(\frac{\sigma^2}{(\nu_1 \delta(h))^2} + \frac{2b}{(\nu_1 \delta(h))} \right)\ln T + \tau.
\end{align}
\end{corollary}
\begin{proof}
Now, if we choose $u' = 1+8c_1\left(\frac{\sigma^2}{(\nu_1 \delta(h))^2} + \frac{2b}{(\nu_1 \delta(h))} \right)\ln T$,\footnote{We fix $c_1=1.2$.} we get from Lemma~\ref{lemma:visits}
\begin{align*}
    \expect[S_{h,i}(T)] &\leq 1+8c_1\left(\frac{\sigma^2}{(\nu_1 \delta(h))^2} + \frac{2b}{(\nu_1 \delta(h))} \right)\ln T + \expect[G_{(h,i),T}^*] + \sum_{t=u'+1}^T \left(\prob[E_2 \wedge S_{h,i}(t) > u'] + \sum_{s=1}^{t-1} \prob[E_{s}]\right)
\end{align*}
By Theorem 3 in~\citep{ucbv}, we get $\sum_{s=1}^{t-1} \prob[E_{s}] \leq \exp[-c_1 \ln T]\left(\frac{24\sigma^2}{(\nu_1 \delta(h))^2} + \frac{4b}{(\nu_1 \delta(h))} \right)$. This is a consequence of the empirical Bernstein's inequality. Thus, $$\sum_{t=u'+1}^T\sum_{s=1}^{t-1} \prob[E_{s}] \leq T e^{-c_1 \ln T}\left(\frac{24\sigma^2}{(\nu_1 \delta(h))^2} + \frac{4b}{(\nu_1 \delta(h))} \right) \leq 0.21 \left(\frac{24\sigma^2}{(\nu_1 \delta(h))^2} + \frac{4b}{(\nu_1 \delta(h))} \right) \ln T,$$ for $c_1 = 1.2$.
By Theorem 1 in~\citep{ucbv}, we get $$\sum_{t=u'+1}^T \prob[E_2 \wedge S_{h,i}(t) > u'] \leq \sum_{t=u'+1}^T \beta(c_1\ln s, s) \leq 0.07 \frac{2b}{\nu_1 \delta(h)}\ln T,$$
for $c_1=1.2$. Combining these three equations together yield the desired result with $c=10$.
\end{proof}

\begin{reptheorem}{thm:regret_delay_noise}[Regret of \frameworkucbv{} with Unknown Noise]
Let us consider that the expected objective function $f$ satisfies Assumption~\ref{ass:lipschitz}, and its $4\nu_1/\nu_2$-near-optimality dimension is $d>0$. Let us also assume that the variance of the noise in evaluations is $\sigma^2$ which is unknown to the algorithm. Then, under Assumption~\ref{ass:partition} and for any $d'>d$, \framework{} algorithm using \ducbv{} achieves expected regret
\begin{align}
    \expect[\reg_T]
    &= \bigO\left(T^{1-\frac{1}{d'+2}}(\tau+((\sigma/\nu_1)^2+2b/\nu_1) \ln T)^{\frac{1}{d'+2}}\right) 
\end{align}
and expected simple regret
\begin{align}
    \epsilon_T &= \bigO\left(T^{-\frac{1}{d'+2}}(\tau+((\sigma/\nu_1)^2+2b/\nu_1) \ln T)^{\frac{1}{d'+2}}\right)
\end{align}
where the expected delay is upper bounded by $\tau$.
\end{reptheorem}
\begin{proof}
From Theorem~\ref{thm:regret_decomp} and Corollary~\ref{cor:ducb1_visits}, we get
\begin{align*}
    \expect[\reg_T] &\leq 4 C \nu_1\nu_2^{-d'} \sum_{h=0}^{H-1} (\delta(h))^{(1-d')} + 4 \nu_1 \delta(H) T \\
    &+ 8C \nu_1\nu_2^{-d'} \sum_{h=1}^H (\delta(h-1))^{(1-d')} \times \left(\frac{10 \sigma^2 \ln T}{(\nu_1 \delta(h))^2} + \frac{20b \ln T}{(\nu_1 \delta(h))} + \tau \right)\\
    &\leq 4 C \nu_1\nu_2^{-d'} \sum_{h=0}^{H-1} (\delta(h))^{(1-d')} (2\tau) + 4 \nu_1 \delta(H) T\\
    &+ (80\sigma^2)C \nu_1^{-1}\nu_2^{-d'} \sum_{h=1}^H (\delta(h-1))^{(1-d')}\delta(h)^{-2} \times \ln T\\
    &+ (160b)C \nu_2^{-d'} \sum_{h=1}^H (\delta(h-1))^{(1-d')}\delta(h)^{-1} \times \ln T.
\end{align*}
For simplicity, we represent the decreasing diameter as $\delta(h) = \rho^h$ for some $\rho \in (0,1)$. Thus, 
\begin{align*}
    \expect[\reg_T] &\leq 4 C \nu_1\nu_2^{-d'} \sum_{h=0}^{H-1} \rho^{h(1-d')} (2\tau) + 4 \nu_1 \rho^H T\\
    &+ (80\sigma^2)C \nu_1^{-1}\nu_2^{-d'} \sum_{h=1}^H \rho^{h(1-d')- (1-d')}\rho^{-2h} \times \ln T
    + (160b)C \nu_2^{-d'} \sum_{h=1}^H \rho^{h(1-d')- (1-d')}\rho^{-h} \times \ln T\\
    &\leq 4 C \nu_1\nu_2^{-d'} \sum_{h=0}^{H-1} \rho^{h(1-d')} (2\tau) + 4 \nu_1 \rho^H T\\
    &+ (80\sigma^2) C \nu_1^{-1}\nu_2^{-d'} \sum_{h=1}^H \rho^{-h(1+d')- (1-d')} \times \ln T
    + (160b)C \nu_2^{-d'} \sum_{h=1}^H \rho^{-hd'- (1-d')}\times \ln T\\
    &= \bigO\left(\sum_{h=0}^{H-1} \rho^{h(1-d')} \tau + (\sigma/\nu_1)^2 \sum_{h=0}^{H-1} \rho^{-h(1+d')} \ln T + 2b/\nu_1 \sum_{h=0}^{H-1} \rho^{-hd'} \ln T + \rho^H T\right)\\
     &= \bigO\left(\sum_{h=0}^{H-1} \rho^{-h(1+d')} (\tau + (\sigma/\nu_1)^2 \ln T + 2b/\nu_1 \ln T) + \rho^H T\right) \text{,      since $\rho \in (0,1)$}\\
     &= \bigO\left(\rho^{-H(1+d')} (\tau + ((\sigma/\nu_1)^2 + 2b/\nu_1) \ln T) + \rho^H T\right)
\end{align*}
By choosing a $\rho$ such that $\rho^{-H(d'+2)} = \frac{T}{\tau+((\sigma/\nu_1)^2 + 2b/\nu_1) \ln T}$, we obtain
\begin{align*}
    \expect[\reg_T] &= \bigO\Bigg(\left(\frac{T}{\tau+((\sigma/\nu_1)^2 + 2b/\nu_1) \ln T}\right)^{\frac{d'+1}{d'+2}} (\tau + (2b/\nu_1+ (\sigma/\nu_1)^2) \ln T)\\
    &\qquad\quad+ \left(\frac{\tau+((\sigma/\nu_1)^2+2b/\nu_1) \ln T}{T}\right)^{\frac{1}{d'+2}} T\Bigg)\\
    &= \bigO\left(T^{1-\frac{1}{d'+2}}(\tau+((\sigma/\nu_1)^2+2b/\nu_1) \ln T)^{\frac{1}{d'+2}}\right) 
\end{align*}
This trivially leads to the bound on simple regret, i.e. the expected error at each step, as $\epsilon_T = \frac{1}{T}\expect[\reg_T]$

\end{proof}
\subsection{Regret of \framework{} with Delayed, Noisy, and Multi-fidelity (DNF) Feedback}
Now, let us consider that we don't only have a delayed and noisy functional evaluator at each step but also the evaluator has different fidelity at each level $h$ of the MC tree. This setup of multi-fidelity MCTS without unknown noise and delay was first considered in~\citep{sen19_MFHOO}. We extend their schematic to the version with delayed and noisy feedback with unknown delays and noise.

Let us consider the cost of selecting a new node at level $h > 0$ is $\lambda(Z_h)$ and the bias added in the decision due to the limited evaluation is $\zeta(Z_h)$. Here, $Z_h$ is the internal state of the multi-fidelity evaluator which influences both the cost of evaluation and the bias in evaluation.

Thus, the multi-fidelity delayed-UCBV selection rule with unknown noise and delay becomes
\begin{align}\label{eq:ducbv-fidelity}
    B_{(h,i),S_{(h,i)}(t-1),t} &\triangleq \hat{\mu}_{(h,i),S_{(h,i)}(t-1)} + \sqrt{\frac{2 \hat{\sigma}_{(h,i),S_{(h,i)}(t-1)}^2 \log t}{S_{(h,i)}(t-1)}} + \frac{3b \log t}{S_{(h,i)}(t-1)} + \zeta(Z_h).
\end{align}

Given this update rule and the multi-fidelity model, we observe that the Lemma 1 of~\citep{sen19_MFHOO} holds.
\begin{lemma}[Lemma 1~\citep{sen19_MFHOO}]\label{lemma:fidel_iter}
When the \framework{} algorithm runs with a total budget $\Lambda$ and the multi-fidelity selection rule (Equation~\eqref{eq:ducbv-fidelity}), then the total number of iterations that the algorithms runs for
\begin{equation}
    T(\Lambda) \geq H(\Lambda) + 1,
\end{equation}
where
\begin{equation}
    H(\Lambda) \triangleq \max \lbrace H: \sum_{h=1}^H \lambda(Z_h) \leq \Lambda \rbrace.
\end{equation}
\end{lemma}

Given Lemma~\ref{lemma:fidel_iter}, we can retain the bounds of Theorem~\ref{thm:regret_delay} and~\ref{thm:regret_delay_noise} by substituting $T= H(\Lambda)$.

\begin{repcorollary}{thm:regret_multifidel}[Multi-fidelity \framework{} with stochastic delays and noise]
Let us consider that the expected objective function $f$ satisfies Assumption~\ref{ass:lipschitz}, and its $4\nu_1/\nu_2$-near-optimality dimension is $d>0$. The function evaluation has $h$-dependent fidelity such that $ H(\Lambda) \triangleq \max \lbrace H: \sum_{h=1}^H \lambda(Z_h) \leq \Lambda \rbrace$ and the induced bias $\zeta(Z_h) = \nu_1 \rho^h$.

Then, under Assumption~\ref{ass:partition} and for any $d'>d$, \framework{} algorithm using \ducb{} achieves expected simple regret
\begin{align}
    \epsilon_\Lambda
    &= \bigO\left(\left({H(\Lambda)}\right)^{-\frac{1}{d'+2}} \left(\ln H(\Lambda)+ \tau\right)^{\frac{1}{d'+2}}\right),
\end{align}
where the expected delay is upper bounded by $\tau$, and
\framework{} algorithm using \ducbv{} achieves expected simple regret
\begin{align}
    \epsilon_\Lambda
    &= \bigO\left(\left({H(\Lambda)}\right)^{-\frac{1}{d'+2}} \left(((\sigma/\nu_1)^2 +2b/\nu_1)\ln H(\Lambda)+ \tau\right)^{\frac{1}{d'+2}}\right),
\end{align}
where the maximum variance of the noise in evaluations is $\sigma^2$.
\end{repcorollary}
\begin{proof}
The proof relies on a simple observation by~\citep{mcts} and~\citep{sen19_MFHOO} that at each iteration of the HOO schematic only one point is selected randomly. Thus, we observe that the simple regret for \framework{} with delayed-UCB1 is
\begin{align*}
    \epsilon_\Lambda &\leq \expect\left[\expect\left[\frac{\reg_T}{T} \mid T(\Lambda)=T\right]\right]\\
    &\leq \expect\left[\left(\frac{\ln T}{T}\right)^{\frac{1}{d'+2}} \left(1+ \frac{\tau}{\ln T}\right)^{\frac{1}{d'+2}} \mid T(\Lambda)=T\right]
\end{align*}
In the first line, the inner expectation is over the randomness of a new node selection and the outer expectation is over the randomness of the computational budget due to inherent state of the multi-fidelity evaluator.

Now, we observe that the inner quantity in the last conditional expectation decreases with $T$ and by Lemma~\ref{lemma:fidel_iter}, $T(\Lambda) \geq H(\Lambda)$ almost surely. Thus, we get
\begin{align}
    \epsilon_\Lambda
    &= \bigO\left(\left({H(\Lambda)}\right)^{-\frac{1}{d'+2}} \left(\ln H(\Lambda)+ \tau\right)^{\frac{1}{d'+2}}\right).\notag
\end{align}
The proof for the \framework{} with delayed-UCBV node selection proceeds similarly.
\end{proof}

\subsubsection{Some Models of Cost Budget} 
Depending on the evaluation problem, we may have different cost functions. Here, we instantiate three such models of cost budget.

Let us denote the bias and cost functions of a multi-fidelity evaluators as $\zeta(.)$ and $\lambda(.)$, and the internal state of the evaluator is $Z_h = \zeta^{-1}(\nu_1\rho^h)$. Under this specification, we show four possible cost models:
\begin{enumerate}[leftmargin=*]
    \item \textbf{Linearly increasing cost:} $\lambda(Z_h) \leq \min\{\beta h, \lambda(1)\}$. This cost model is observed for hyperparameter tuning of deep neural networks.
    
    Under \textbf{cost model 1}, $$\Lambda \leq \sum_{h=1}^H \lambda(Z_h) \leq \sum_{h=1}^H \min\{\beta h, \lambda(1)\} \leq 1/2({\sum_{h=1}^H \beta h + \lambda(1)}) \leq \frac{1}{4}(\beta H^2) + \lambda(1)/2).$$ Thus, $$H(\Lambda) \geq \sqrt{2(2\Lambda - \lambda(1))/\beta}.$$
    
    \item \textbf{Constant cost:} $\lambda(Z_h) \leq \min\{\beta, \lambda(1)\}$, where $\beta > 0$. This cost model is observed for hyperparameter tuning of deep neural networks.
    
    Under \textbf{cost model 2}, \[\Lambda \leq \sum_{h=1}^H \lambda(Z_h) \leq \sum_{h=1}^H \min\{\beta, \lambda(1)\} \leq \frac{\sum_{h=1}^H \beta + \lambda(1)}{2} \leq \frac{1}{2}(H\beta + \lambda(1)).\] Thus, $$H(\Lambda) \geq (2\Lambda - \lambda(1))/\beta.$$

    \item \textbf{Polynomially decaying cost.} $\lambda(Z_h) \leq \min\{h^{-\beta}, \lambda(1)\}$, where $\beta>0$ and $\beta \neq 1$. This cost model is observed for tuning database parameters in UDO.
    
    Under \textbf{cost model 3}, 
    \begin{align*}
        \Lambda \leq \sum_{h=1}^H \lambda(Z_h) &\leq \sum_{h=1}^H \min\{h^{-\beta}, \lambda(1)\}\\ &\leq \frac{\sum_{h=1}^H h^{-\beta} + \lambda(1)}{2} \leq \frac{1}{2}(\mathrm{Har}_\beta(H) + \lambda(1))\leq \frac{1}{2}(\frac{H^{1-\beta}-1}{1-\beta} + \lambda(1)).
    \end{align*} Here, $\mathrm{Har}_\beta(H)$ is the generalised harmonic function of $H$ of order $\beta$ and is upper bounded by $\mathrm{RiemannZeta}(\beta)$. Thus, $$H(\Lambda) \geq (1+(1-\beta)(2\Lambda-\lambda(1)))^{1/(1-\beta)}.$$

    \item \textbf{Exponentially decaying cost.} $\lambda(Z_h) \leq \min\{\beta^{-h}, \lambda(1)\}$, where $\beta \in (\rho, 1)$. This cost model is observed in tuning strongly convex functions with accelerated gradient descent.
    
    Under \textbf{cost model 4}, \[\Lambda \leq \sum_{h=1}^H \lambda(Z_h) \leq \sum_{h=1}^H \min\{\beta^{-h}, \lambda(1)\} \leq \frac{\sum_{h=1}^H \beta^{-h} + \lambda(1)}{2} \leq \frac{1}{2}(\frac{\beta^{- H}-1}{1-\beta} + \lambda(1)).\] Thus, $$H(\Lambda) \geq \log_{1/\beta}\left(1+(1-\beta)(2\Lambda - \lambda(1))\right).$$
\end{enumerate}

\newpage
\section{Relaxing the Smoothness Assumptions}
\subsection{Local Smoothness with respect to the $\tree$}
Performing a global black-box optimization without any regularity assumption on the function is too good to be true.
In tree search literature, initially, \cite{agrawal1995continuum,auer2007improved,kleinberg2008multi,kleinberg2019bandits} have proposed to use global smoothness conditions on the objective functions.
\cite{bubeck2011a} proposes to relax the pointwise global assumption over the whole domain $\domain$ of $f$, for example global continuity or Lipschitzness~\citep{goldstein1977optimization}, to the continuity condition around the optimum $f(x^*)$.
In Section~\ref{sec:back}, we present the weak Lipschitzness assumption (Assumption~\ref{ass:lipschitz}) to elaborate the \framework{} framework and later on to derive the corresponding regret bounds.
We use this assumption to keep the results directly comparable with the original HOO algorithm~\citep{bubeck2011a}.
Now, we show that the assumption of weak Lipschitzness (Assumption~\ref{ass:lipschitz}) can be relaxed for \framework{} to a completely \textit{local} smoothness assumption (Assumption~\ref{ass:local_lip}).

\cite{poo_first} shows that the only smoothness condition required is local smoothness with respect to the partition constructed by the tree $\tree$.
\begin{assumption}[Local Smoothness with respect to the $\tree$]\label{ass:local_lip}
Given the global maximizer $x^* \in \domain$, we denote the index of the node at depth $h$ containing $x^*$ as $l_{h}^*$. Then, we assume that there exists a global maximizer $x^*$ and smoothness parameters $\nu>0$, $\rho \in (0,1)$ such that,
\begin{align}
    \forall h \geq 0, \forall x \in \domain_{(h,l_{h}^*)}, \quad f(x) \geq f^* - \nu \rho^h.
\end{align}
\end{assumption}
This assumption shows that the only constrain we need on $f$ is that along the optimal path of the covering tree. This is a plausible property in an optimization problem and also increases applicability of the algorithm significantly~\citep{poo,sen18_mfdoo,sen19_MFHOO}.
The local assumption allows us to redefine the near-optimality dimension of Definition~\ref{def:NOpt_d}.
\begin{definition}[Near-optimality Dimension with respect to the $\tree$]\label{def:NOpt_local}
For any $\nu > 0$, and $\rho \in (0,1)$, 
the near-optimality dimension is defined as
\begin{align*}
    d(\nu, \rho) \triangleq \inf \lbrace d'\in \real^+: \exists C(\nu, \rho), \forall h \geq 0, \mathcal{N}_h(2\nu\rho^h) \leq C(\nu, \rho) \rho^{-d'h} \rbrace.
\end{align*}
Here, $\mathcal{N}_h(\epsilon)$ is the number of nodes $(h,l)$ such that $\sup_{x \in (h,l)} f(x) \geq f^* - \epsilon$. In other words, $\mathcal{N}_h(\epsilon)$ is the number of nodes at depth $h$ that covers the $\epsilon$-optimal region $\domain_{\epsilon}$.
\end{definition}
Given the corresponding smoothness assumptions, Definition~\ref{def:NOpt_d} and~\ref{def:NOpt_local} are analogous in intuition . They show the dependence of the global optimization problem on the volume of the near-optimal regions and their rate of shrinking with the depth of the tree~\citep{auer2007improved}. 
Specifically, $\mathcal{N}_h(2\nu\rho^h)$ is the number of nodes that any algorithm has to sample in order to find the optimum, and the optimization problem gets easier as the near-optimality dimension $d(\nu, \rho)$ decreases. Another interesting observation is that $d(\nu, \rho)$ depends only on $f$ and $\tree$, and not on the choice of the dissimilarity metric like Definition~\ref{def:NOpt_d}.

Given this new assumption, we restate the regret bound of Theorem~\ref{thm:regret_decomp}.
\begin{theorem}\label{thm:regret_decomp_local}
Let us consider that the expected objective function $f$ satisfies Assumption~\ref{ass:lipschitz}, and its $4\nu_1/\nu_2$-near-optimality dimension is $d>0$. Then, under Assumption~\ref{ass:partition} and for any $d'>d$, \framework{} algorithm uses a bandit algorithm MAB for node selection will achieve expected regret
\begin{align}
    \expect[\reg_T] &\leq 3 \nu C(\nu, \rho) \sum_{h=0}^{H-1} \rho^{h(1-d(\nu, \rho))} + 3 \nu \rho^H T
    + 6 \nu C(\nu, \rho) \sum_{h=1}^{H} \rho^{(h-1)(1-d(\nu, \rho))} \times (U(T,\tau,\nu\rho^h)).
\end{align}
Here, $U(T,\tau,\nu\rho^h)$ is the upper bound on number of visits to the $2\nu\rho^h$-suboptimal nodes at depth $h>0$ by \bandit.
\end{theorem}
\begin{proof}
From Equation~\eqref{eq:reg_Decomp}, we obtain the regret decomposition given three subtrees $\tree_1$, $\tree_2$, and $\tree_3$.
\begin{align}
    \expect[\reg_T] = \expect[\reg_{T,1}] + \expect[\reg_{T,2}] + \expect[\reg_{T,3}]
\end{align}

Case 1: All nodes in $I_h$ are $2\nu\rho^h$-optimal for all $h\geq 0$. Thus, all points in the corresponding subdomains are $3\nu\rho^H$-optimal (Assumption~\ref{ass:local_lip}).
\begin{align}\label{eq:reg1_ll}
    \expect[\reg_{T,1}] \leq \sum_{h=0}^{H-1} 3 \nu \rho^h |I_h| \leq \sum_{h=0}^{H-1} 3 \nu \rho^h C(\nu,\rho) \rho^{-d(\nu,\rho) h} = 3 \nu C(\nu,\rho) \sum_{h=0}^{H-1} \rho^{h(1-d(\nu,\rho))}.
\end{align}
The second inequality is a consequence of the Definition~\ref{def:NOpt_local} of the near-optimality dimension $d(\nu, \rho)$.

Case 2: Since the nodes in $\tree_2$ are $2\nu\rho^H$-optimal, all points in the corresponding subdomains are $3\nu\rho^H$-optimal (Assumption~\ref{ass:local_lip}). Thus,
\begin{align}\label{eq:reg2_ll}
    \expect[\reg_{T,2}] \leq 3 \nu_1 \rho^H T.
\end{align}

Case 3: In $\tree_3$, any node of $J_h$ has a parent in $I_{h-1}$. Thus, the subdomains covered by nodes in $J_h$ are at least $3\nu\rho^{h-1}$-optimal (Assumption~\ref{ass:partition} and~\ref{ass:local_lip}). Thus, we obtain
\begin{align}
    \expect[\reg_{T,3}] &\leq \sum_{h=1}^H 3\nu \rho^{h-1} \sum_{l:(h,l) \in J_h}  \expect[S_{h,l}(T)]\notag\\ 
    &\leq \sum_{h=1}^H 3\nu \rho^{h-1} |J_h| \max_{(h,l) \in J_h} \expect[S_{h,l}(T)]\label{eq:a}\\
    &\leq \sum_{h=1}^H 6\nu \rho^{h-1} |I_{h-1}| \max_{(h,l) \in J_h} \expect[S_{h,l}(T)]\label{eq:b}\\
    &\leq \sum_{h=1}^H 6\nu \rho^{h-1} C(\nu,\rho) \rho^{-d(\nu,\rho) (h-1)} \max_{(h,l) \in J_h} \expect[S_{h,l}(T)]\label{eq:c}\\
    &\leq 6\nu C(\nu,\rho) \sum_{h=1}^H  \rho^{(1-d(\nu,\rho))(h-1)} U(T,\tau,\nu\rho^h)\label{eq:reg3_ll}
\end{align}
Equation~\eqref{eq:a} is obtained from the fact $\sum_{i=1}^K a_i \leq K \max_i a_i$.
Equation~\eqref{eq:b} holds true as the parents of nodes of $J_h$ are in $I_{h-1}$, and by the way the tree grows $|J_h| \leq 2 |I_{h-1}|$ for any $h \geq 1$ (Branching factor = 2).
Equation~\eqref{eq:c} is a direct consequence of the near-optimality dimension in Defintion~\ref{def:NOpt_local}.
The last inequality is obtained as we denote the upper bound on number of visits to the $2\nu\rho^h$-suboptimal nodes at depth $h>0$, i.e. nodes in $J_h$, by \bandit as $U(T,\tau,\nu\rho^h)$.

Combining Equations~\eqref{eq:reg1_ll},~\eqref{eq:reg2_ll}, and~\eqref{eq:reg3_ll} concludes the proof.
\end{proof}

If we consider $\nu_1 = \nu$, $d' = d(\nu, \rho)$, $C\nu_2^{-d'} = C(\nu, \rho)$, and $\delta(h) = \rho^h$, we obtain that the results of Theorem~\ref{thm:regret_decomp} and~\ref{thm:regret_decomp_local} only differ by constant factors. They also reduce the problem of bounding regret of \framework{} to finding an upper bound on $U(T,\tau,\nu\rho^h)$ for a given \bandit algorithm. Thus, we confirm that if we restate the regret bounds proved using Theorem~\ref{thm:regret_decomp} with the local smoothness assumption, they will have same dependency on $T$, $\tau$, $\sigma^2$ and $\Lambda$, while $d'$ changes to $d(\nu, \rho)$.

\subsection{Unknown Smoothness}
In Algorithm~\ref{alg:pcts}, we present a simplistic version of \framework{} that takes the smoothness parameters $(\nu_1, \rho)$ as input. But in practice, we do not need to know the smoothness parameter. We take the Parallel Optimistic Optimizaton (POO) approach proposed by~\cite{poo_first} and later on extensively used in hierarchical tree search literature~\citep{poo,lazaric2014online,Shang2018AdaptiveBO,sen18_mfdoo,sen19_MFHOO}. 

Given a hierarchical tree search algorithm $\mathcal{A}$, POO($\mathcal{A}$) takes maximum values of the smoothness parameters $(\nu_{1_{\max}}, \rho_{\max})$ as input.\footnote{Both POO and $\mathcal{A}$ also take the branching factor of the tree, say $K$, as input. In \framework, we fix $K$ to $2$. Thus, we omit mentioning it.}  
At first, POO($\mathcal{A}$) chooses $N = \frac{0.5\ln 2}{\ln(1/\rho_{\max})} \log(T/\log T)$ points $\{\rho_i\}_{i=1}^N$ in the interval $[0,\rho_{\max}]$, such that $\rho_i \triangleq \rho_{\max}^{\frac{2N}{i+1}}$.
Then, it parallely spawns $N$ instances of $\mathcal{A}$ with smoothness parameters $(\nu_{1_{\max}}, \rho_i)$ as input.
Finally, POO outputs the maximum of the optimal values computed by these $N$ instances of $\mathcal{A}$. We denote the smoothness parameters corresponding to that tree as $(\nu^*, \rho^*)$. 
In brief, POO($\mathcal{A}$) performs a geometric line search for the parameter $\rho^*$ in the interval $[0,\rho_{\max}]$ that maximizes the optimal values achieved by $\mathcal{A}(\nu_{1_{\max}}, \rho)$. Here, $\rho \in [0,\rho_{\max}]$.

Due to the multi-fidelity feedback, we adopt a specific version of POO, i.e. MFPOO~\citep[Algorithm 2]{sen19_MFHOO}, which is designed to be compatible with fixed budget and multi-fidelity feedback.

\begin{theorem}[\framework{} with Unknown Smoothness and DNF Feedback]\label{thm:final}
If \framework{} with DNF feedback is executed with parameters $\nu_{max} (\geq \nu^*)$, $\rho_{max}(\geq \rho^*)$, and a total cost budget
$\Lambda$, then under Assumptions~\ref{ass:local_lip} and~\ref{ass:partition}, the expected simple regret of at least one of the \frameworkucb{} instances is bounded by, 
\begin{align}
    \epsilon_\Lambda
    &= \bigO\left((\nu_{max}/\nu^*)^{D_{max}}\left({H(\Lambda/\log \Lambda)}\right)^{-\frac{1}{d'+2}} \left(\tau + \ln H(\Lambda/\log \Lambda)\right)^{\frac{1}{d'+2}}\right).
\end{align}
and expected simple regret of at least one of the \frameworkucbv{} instances is bounded by,
\begin{align}
    \epsilon_\Lambda
    &= \bigO\left((\nu_{max}/\nu^*)^{D_{max}}\left({H(\Lambda/\log \Lambda)}\right)^{-\frac{1}{d'+2}} \left(\tau + (\sigma^2 +2b)\ln H(\Lambda/\log \Lambda)\right)^{\frac{1}{d'+2}}\right).
\end{align}
Here, $D_{max} = \log 2/\log(1/\rho_{max})$, $\sigma^2$ is the maximum noise variance, $\tau$ is the upper bound on expected delay, and $b$ is the range of optimization domain.
\end{theorem}
Given the upper bound on the simple regret of the base algorithm $\mathcal{A}$ of $MFPOO(\mathcal{A})$ under multi-fidelity feedback, the  proof technique of Theorem 2 in~\citep{sen19_MFHOO} can be reproduced. We merge Theorem 2 in~\citep{sen19_MFHOO} with the results of Theorem~\ref{thm:regret_multifidel} to get the Theorem~\ref{thm:final}.
Another interesting thing to note is that for MFPOO$(\mathcal{A})$ to work Assumption~\ref{ass:local_lip} has to hold for one of the optimizers only, while it may spawn multiple maximizers~\citep{poo_first,poo}.

\newpage
\section{Additional Experimental Results}
\textit{For the experimental analysis, we implement both \frameworkucbs{} and \frameworkucbv{} with the local smoothness assumption and unknown smoothness parameters.}\footnote{Link to our code: \url{https://github.com/jxiw/PCTS}}

\subsection{Details of Multi-fidelity Evaluations}
We compare the performance of \framework{}  with: BO algorithms (BOCA \citep{boca}, GP-UCB \citep{gpucb}, MF-GP-UCB \citep{mfgpucb}, GP-EI \citep{gpei}, MF-SKO \citep{sko}), tree search algorithms (MFPOO \citep{sen18_mfdoo}, MFPOO with UCB-V~\citep{ucbv}), zeroth-order GD algorithms (OGD, DBGD~\citep{li2019bandit}).\footnote{We use the implementations in \url{https://github.com/rajatsen91/MFTreeSearchCV} for baselines except OGD, DBGD, and MFPOO-UCBV.}

In the section D.2, we show experiments of those algorithms on synthetic functions with DNF feedbacks. In the section D.3, we show the experiments for hyperparameter tuning of different machine learning models using those algorithms with DNF feedbacks. In the section D.4. we consider the stochastic delay instead of constant delay, and show experiment results for different algorithms under the stochastic delay. 



\subsection{Details of Optimizing Synthetic Functions}
\label{se:ex_sythetic}
We evaluate the performance of aforementioned algorithm on those benchmark functions which are widely used in the black-box optimization literature \citep{sen18_mfdoo, sen19_MFHOO}. Those functions have been modified to incorporate the fidelity space $\mathcal{Z} = [0, 1]$ as \citep{sen18_mfdoo, sen19_MFHOO} suggest.

\subsubsection{Experiment Setup of Synthetic Functions}
We run each experiment ten times for 600s on a MacBook Pro with a 6-core Intel(R) Xeon(R)@2.60GHz CPU and plot the median value of simple regret, shown in Figure ~\ref{fig:additional_experiement_sythetic}. The delay time $\tau$ for all synthetic functions is set to four seconds. The noise is added from Gaussian distributions with corresponding variance $\sigma^2$. This $\sigma$ is passed to UCB1-$\sigma$ and DUCB1-$\sigma$ in MFPOO and \framework{} as it assumes the noise variance is known~\citep{sen19_MFHOO}.For \frameworkucbv{} and MFPOO-UCBV, we do not pass the exact upper bound $b$ of the function rather a loose upper bound on it, i.e. 5. For MFPOO and \framework{}, we set the $\nu_{\max} = 0.95$ and $\rho_{\max} = 1.0$. The final optimal values of different algorithms on different synthetic functions are shown in Table ~\ref{tb:experiment_sythetic}. In all those experiments, either \frameworkucb1{} or \frameworkucbv{} achieve the lowest simple regret. 


\subsubsection{Description of Synthetic Functions}

\paragraph{Hartmann functions \citep{hartmann3}} We use two Hartmann functions in 3 and 6 dimensions, named as Hartmann3 and Hartmann6. The multi-fidelity object is 

\begin{align*}
f_z(x) = \sum_{i=1}^4 (\alpha_i - \alpha^{\prime}(z)) \exp{(- \sum_{j=1}^3 A_{ij} (x_j - P_{ij})^2)}
\end{align*}

where $\alpha = [1.0,1.2,3.0,3.2]$ and $\alpha^{\prime}(z) = 0.1 (1 - z)$.

For the hartmann3, the cost function is $\lambda(z) = 0.05 + (1 - 0.05) z^3$ and noise variance $\sigma^2 = 0.01$. The delay time $\tau$ is set to four seconds. The matrix $A$ and $P$ are,
\begin{align*}
    A = \begin{bmatrix}
    3 & 10 & 30 \\
    0.1 & 10 & 35 \\
    3 & 10 & 30 \\
    0.1 & 10 & 35 \\
    \end{bmatrix}
    \quad \quad \quad
    P = 10^{-4} \times \begin{bmatrix}
    3689 & 1170 & 2673 \\
    4699 & 4387 & 7470 \\
    1091 & 8732 & 5547 \\
    381 & 5743 & 8828 \\
    \end{bmatrix}
\end{align*}

For the hartmann6, the cost function is $\lambda(z) = 0.05 + (1 - 0.05) z^3$ and noise variance $\sigma^2 = 0.05$. The delay time $\tau$ is set to four seconds. The matrix $A$ and $P$ are,
\begin{align*}
    A = \begin{bmatrix}
    10 & 3 & 17 & 3.5 & 1.7 & 8 \\
    0.05 & 10 & 17 & 0.1 & 8 & 14 \\
    3 & 3.5 & 1.7 & 10 & 17 & 8 \\
   17 & 8 & 0.05 & 10 & 0.1 & 14 \\
    \end{bmatrix}
    ~~P = 10^{-4} \times 
    \begin{bmatrix}
    1312 & 1696 & 5569 & 124 & 8283 & 5886 \\
    2329 & 4135 & 8307 & 3736 & 1004 & 9991 \\
    2348 & 1451 & 3522 & 2883 & 3047 & 6650 \\
    4047 & 8828 & 8732 & 5743 & 1091 & 381\\
    \end{bmatrix}
\end{align*}

\paragraph{Currin  exponential  function \citep{currin}.}
The input domain $\mathcal{X}=[0,1]^2$. The cost function for CurrinExp is $\lambda(z) = 0.1 + z^2$ and the noise variance $\sigma^2 = 0.05$. The multi-fidelity object as a function of $(x, z)$ is

\begin{align*}
    f_z(x) = \left( 1 - 0.1(1 - z) \exp{ \left( \frac{-1}{2 x_2}  \right)} \right)  \times
     \left(  \frac{2300 x^3_1 + 1900x^2_1 + 2092 x_1 + 60}{100x^3_1 +500x^2_1 +4x_1 +20}  \right) 
\end{align*}

\paragraph{Borehole function \citep{borehole}} The cost function is $\lambda(z) = 0.1 + z^{1.5}$ and noise variance $\sigma^2 = 0.01$. The multi-fidelity object as a function of $(x, z)$ is

\begin{align*}
    f_z(x) = \frac{2 z \pi T_u (H_u - H_l)}{\log(r/r_w) (1 + \frac{2 L T_u}{\log(r/r_w) r_w^2 K_w} + \frac{T_u}{T_l})} +
     \frac{5 (1 - z) T_u (H_u - H_l)}{\log(r/r_w) (1.5 + \frac{2 L T_u}{\log(r/r_w) r_w^2 K_w} + \frac{T_u}{T_l}) } 
\end{align*}
where $x = [r_w, r, T_u, H_u, T_l, H_l, L, H_w]$. The delay time $\tau$ is set to four seconds.

\paragraph{Branin function \citep{branin}} 
The input domain $\mathcal{X}=[[-5,10],[0,15]]^2$. The objective function is 

\begin{align*}
f_z(x)=a(x_2 - b(z)x^2_1 +c(z)x_1 - r)^2 + s (1 - t(z)) \cos{(x_1)} + s,
\end{align*}

where $a = 1, b(z) = \frac{5.1}{4 \pi^2} - 0.01(1 - z) c(z) = \frac{5}{\pi} - 0.1(1 - z), r = 6, s = 10$ and $t(z) = \frac{1}{8 \pi} + 0.05 (1-z)$. When $z=1$, it becomes the standard Branin function. The cost function is $\lambda(z) = 0.05 + z^3$ and noise variance $\sigma^2 = 0.05$. The delay time $\tau$ is set to four seconds.

\paragraph{Schwefel function \citep{schwefel}} 
The objective function is
\begin{align*}
f(x) = - 418.9829 d + \sum_{i=1}^{d} x_i \sin{\sqrt{|x_i|}}
\end{align*}

We evaluate this function on the hypercube $x_i \in [0, 500]$, for $i = 1, \cdots, 20$ and $d = 20$. The noise variance $\sigma^2 = 0.1$ 

\subsubsection{Statistics of optimal values achieved by different optimizers}
\begin{figure}[h!]
    \centering
    \includegraphics[clip, width=\textwidth]{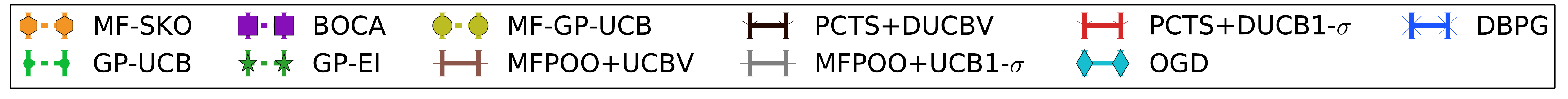}
    \includegraphics[clip, width=\textwidth]{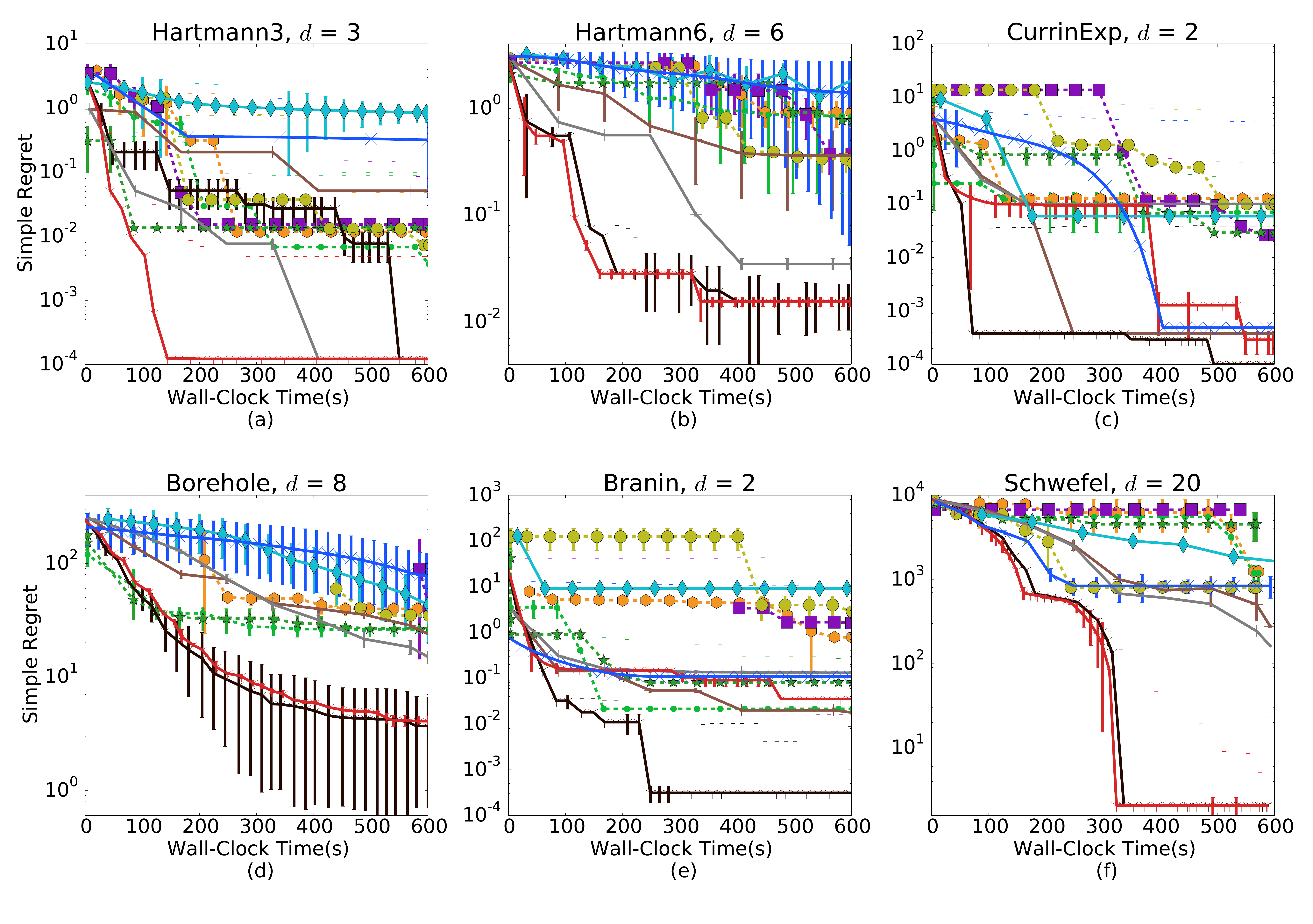}
    \vspace*{-1.6em}
    \caption{Figures (a) to (f) show simple regret (median of 10 runs) of different algorithms on synthetic functions with DNF feedbacks and the corresponding error bars $(y_{\text{median}} - y_{\min}, y_{\max} - y_{\text{median}})$ are plotted.}\label{fig:additional_experiement_sythetic}
    \vspace*{-1em}
\end{figure}

The median value of simple regret over ten runs are shown in Figure~\ref{fig:additional_experiement_sythetic} with the error bar. The error bar indicates the spread of the maximum and minimum values obtained over the ten runs.
The statistics of optimal values of different algorithms after 600s on different synthetic functions are shown in Table~\ref{tb:experiment_sythetic}. In all the cases, we observe that either \frameworkucbs{} or \frameworkucbv{} outperforms the competing optimization algorithms.

\begin{table}[p]
\centering
\caption{Maximum, Median, and Standard Deviation of optimal values over $10$ runs of different algorithms for different synthetic functions.}~\label{tb:experiment_sythetic}
\vspace*{1em}
\begin{tabular}{|l|l|c|c|c|}
\hline
Synthetic functions &
  Algorithms &
  \multicolumn{1}{c|}{Max value} &
  \multicolumn{1}{c|}{Median value} &
  \multicolumn{1}{c|}{Std. Dev.} \\ \hline
\multirow{11}{*}{\begin{tabular}[c]{@{}l@{}}Hartmann3 \\ \\ (optimal value = 3.86278)\end{tabular}} &
  MFPOO+UCB1-$\sigma$ &
  3.862658 &
  3.862658 &
  4.71E-07 \\ \cline{2-5} 
 & MFPOO+UCBV          & 3.811498              & 3.811498              & \textbf{6.84E-10}              \\ \cline{2-5} 
 & \frameworkucbs{} & 3.862658              & 3.862658              & 4.70E-03              \\ \cline{2-5} 
 & \frameworkucbv{}         & \textbf{3.862659688}           & \textbf{3.8626584}             & 6.44E-07              \\ \cline{2-5} 
 & GP-UCB              & 3.8591                & 3.8591                & 8.21E-02              \\ \cline{2-5} 
 & MF-GP-UCB           & 3.8555                & 3.8555                & 2.97E-01              \\ \cline{2-5} 
 & BOCA                & 3.85496942            & 3.8474                & 7.03E-02              \\ \cline{2-5} 
 & GP-EI               & 3.8492                & 3.8492                & 8.59E-02              \\ \cline{2-5} 
 & MF-SKO              & 3.8511                & 3.8511                & 3.67E-02              \\ \cline{2-5} 
 & OGD                 & 3.54411674            & 3.014917788           & 2.65E-01              \\ \cline{2-5} 
 & DBGD                & 3.54411674            & 3.54411674            & 2.78E-01              \\ \hline
\multirow{11}{*}{\begin{tabular}[c]{@{}l@{}}Hartmann6\\ \\ (optimal value = 3.32237)\end{tabular}} &
  MFPOO+UCB1-$\sigma$ &
  3.292949 &
  3.287516333 &
  4.38E-03 \\ \cline{2-5} 
 & MFPOO+UCBV          & 3.292949              & 2.958906875           & 2.55E-01              \\ \cline{2-5} 
 & \frameworkucbs{} & \textbf{3.306916432}          & \textbf{3.305830186}           & \textbf{1.54E-03}              \\ \cline{2-5} 
 & \frameworkucbv{}          & \textbf{3.306916432}           & 3.298213481           & 7.11E-03              \\ \cline{2-5} 
 & GP-UCB              & 3.16314701            & 2.4833599             & 6.80E-01              \\ \cline{2-5} 
 & MF-GP-UCB           & 3.07564567            & 3.01829366            & 5.74E-02              \\ \cline{2-5} 
 & BOCA                & 3.08446115            & 2.94932738            & 1.35E-01              \\ \cline{2-5} 
 & GP-EI               & 2.58371424            & 2.551397575           & 3.23E-02              \\ \cline{2-5} 
 & MF-SKO              & 2.6617437             & 2.4059598             & 2.56E-01              \\ \cline{2-5} 
 & OGD                 & 2.25086772            & 1.395420838           & 2.25E+00              \\ \cline{2-5} 
 & DBGD                & 2.954460562           & 1.916368356           & 1.35E+00              \\ \hline
\multirow{11}{*}{\begin{tabular}[c]{@{}l@{}}CurrinExp\\ \\ (optimal value = 13.798685)\end{tabular}} &
  MFPOO+UCB1-$\sigma$ &
  13.697782 &
  13.697675 &
  6.48E-05 \\ \cline{2-5} 
 & MFPOO+UCBV          & 13.798306             & 13.798306             & 8.85E-05              \\ \cline{2-5} 
 & \frameworkucbs{} & \textbf{13.798685}             & 13.798396             & 1.36E-04              \\ \cline{2-5} 
 & \frameworkucbv{}          & 13.798585             & \textbf{13.798585}             & 3.85E-02              \\ \cline{2-5} 
 & GP-UCB              & \textbf{13.798685 }            & 13.72859575           & 3.25E-02              \\ \cline{2-5} 
 & MF-GP-UCB           & 13.77983743           & 13.69857206           & 3.70E+00              \\ \cline{2-5} 
 & BOCA                & 13.798685             & 13.77240763           & 2.89E-02              \\ \cline{2-5} 
 & GP-EI               & 13.79599637           & 13.76940012           & 6.32E-02              \\ \cline{2-5} 
 & MF-SKO              & 13.67126373           & 13.67126373           & 1.86E-01              \\ \cline{2-5} 
 & OGD                 & 13.73885226           & 13.73885226           & \textbf{2.17E-11}              \\ \cline{2-5} 
 & DBGD                & 13.79819971           & 13.7891421            & 3.45E+00              \\ \hline
\multirow{11}{*}{\begin{tabular}[c]{@{}l@{}}Borehole\\ \\ (optimal value = 309.523221)\end{tabular}} &
  MFPOO+UCB1-$\sigma$ &
  297.097139 &
  294.6182492 &
  1.91E+00 \\ \cline{2-5} 
 & MFPOO+UCBV          & 290.466079            & 285.769891            & 4.14E+00              \\ \cline{2-5} 
 & \frameworkucbs{} & 305.9956224           & 305.4490009           & \textbf{4.57E-01}              \\ \cline{2-5} 
 & \frameworkucbv{}          & \textbf{308.3696046}           & \textbf{305.8342653}           & 2.99E+00              \\ \cline{2-5} 
 & GP-UCB              & 288.8612949           & 284.6253047           & 4.09E+00              \\ \cline{2-5} 
 & MF-GP-UCB           & 278.6624033           & 274.664564            & 4.00E+00              \\ \cline{2-5} 
 & BOCA                & 278.0044636           & 269.1741653           & 9.29E+00              \\ \cline{2-5} 
 & GP-EI               & 285.1259936           & 283.1443876           & 2.26E+00              \\ \cline{2-5} 
 & MF-SKO              & 276.8185156           & 269.7879994           & 5.39E+00              \\ \cline{2-5} 
 & OGD                 & 292.3678513           & 266.3082687           & 2.31E+01              \\ \cline{2-5} 
 & DBGD                & 296.0795646           & 231.7955644           & 4.55E+01              \\ \hline
\multirow{11}{*}{\begin{tabular}[c]{@{}l@{}}Branin\\ \\ (optimal value =  -0.3979)\end{tabular}} &
  MFPOO+UCB1-$\sigma$ &
  -0.519116 &
  -0.52918 &
  9.20E-03 \\ \cline{2-5} 
 & MFPOO+UCBV          & -0.415716             & -0.415716             & \textbf{1.54E-03}              \\ \cline{2-5} 
 & \frameworkucbs{} & -0.4331555716         & -0.4331555716         & 5.20E-02              \\ \cline{2-5} 
 & \frameworkucbv{}          & \textbf{-0.3987907502  }       & \textbf{-0.3988127406}         & 8.58E-03              \\ \cline{2-5} 
 & GP-UCB              & -0.41925899           & -0.41925899           & 1.23E-01              \\ \cline{2-5} 
 & MF-GP-UCB           & -3.24208941           & -3.24208941           & 1.93E-02              \\ \cline{2-5} 
 & BOCA                & -2.02091133           & -2.02091133           & 2.86E-01              \\ \cline{2-5} 
 & GP-EI               & -0.47900755           & -0.47900755           & 9.94E-02              \\ \cline{2-5} 
 & MF-SKO              & -1.1853606            & -1.1853606            & 1.12E+00              \\ \cline{2-5} 
 & OGD                 & -9.524725088          & -10.96088904          & 1.43E+01              \\ \cline{2-5} 
 & DBGD                & -0.5061438122         & -0.5205826043         & 1.02E+01              \\ \hline
\end{tabular}%
\end{table}

\newpage
\subsubsection{Key statistics of the trees constructed by MFPOO and \framework{}}
The detail comparisons between MFPOO and \framework{} are shown in Table~\ref{tb:tree_statistics_sythetic}. In general, the depth of the tree created using \framework{} algorithm is significantly larger than the depth of the tree created using MFPOO algorithm for both UCB1-$\sigma$ and UCBV policy.

\begin{table}[h!]
\centering
\caption{Statistics of the trees constructed by MFPOO and \framework{} based approaches for optimizing different synthetic functions.}~\label{tb:tree_statistics_sythetic}
\resizebox{\textwidth}{!}{%
\begin{tabular}{|l|l|r|r|r|r|r|}
\hline
Synthetic functions &
  Tree Search Algorithms &
  \multicolumn{1}{l|}{Tree Height} &
  \multicolumn{1}{l|}{Number of Tree Nodes} &
  \multicolumn{1}{l|}{Number of Iterations (T)} &
  \multicolumn{1}{l|}{Best $\rho$} &
  \multicolumn{1}{l|}{Best $\nu_1$} \\ \hline
\multirow{4}{*}{Hartmann3} & MFPOO+UCB1-$\sigma$ & 21 & 151 & 75  & 0.9259454628 & 0.006103201358 \\ \cline{2-7} 
                           & MFPOO+UCBV          & 10 & 151 & 75  & 0.9259454628 & 0.006103201358 \\ \cline{2-7} 
                           & \frameworkucbs{} & \textbf{27} & 603 & 301 & 0.95         & 0.006103201358 \\ \cline{2-7} 
                           & \frameworkucbv{}          & 18 & 681 & 340 & 0.95         & 0.006103201358 \\ \hline
\multirow{4}{*}{Hartmann6} & MFPOO+UCB1-$\sigma$ & 25 & 151 & 75  & 0.9259454628 & 0.00842254613  \\ \cline{2-7} 
                           & MFPOO+UCBV          & 25 & 151 & 75  & 0.9259454628 & 0.00842254613  \\ \cline{2-7} 
                           & \frameworkucbs{} & 36 & 593 & 296 & 0.95         & 0.00842254613  \\ \cline{2-7} 
                           & \frameworkucbv{}          & \textbf{39} & 427 & 213 & 0.857375     & 0.01179156458  \\ \hline
\multirow{4}{*}{CurrinExp} & MFPOO+UCB1-$\sigma$ & 15 & 151 & 75  & 0.9259454628 & 0.6094708386   \\ \cline{2-7} 
                           & MFPOO+UCBV          & 12 & 151 & 75  & 0.9259454628 & 0.6094708386   \\ \cline{2-7} 
                           & \frameworkucbs{} & \textbf{25} & 607 & 303 & 0.9259454628 & 0.6094708386   \\ \cline{2-7} 
                           & \frameworkucbv{}          & 18 & 915 & 457 & 0.95         & 0.6094708386   \\ \hline
\multirow{4}{*}{Borehole}  & MFPOO+UCB1-$\sigma$ & 42 & 151 & 75  & 0.9259454628 & 20.46960058    \\ \cline{2-7} 
                           & MFPOO+UCBV          & 35 & 151 & 75  & 0.857375     & 28.65744081    \\ \cline{2-7} 
                     & \frameworkucbs{} & \textbf{65} & 597 & 298 & 0.857375     & 28.65744081    \\ \cline{2-7} 
                           & \frameworkucbv{}          & 60 & 601 & 300 & 0.857375     & 28.65744081    \\ \hline
\multirow{4}{*}{Branin}    & MFPOO+UCB1-$\sigma$ & 20 & 151 & 75  & 0.9259454628 & 0.8313313704   \\ \cline{2-7} 
                           & MFPOO+UCBV          & 13 & 151 & 75  & 0.9259454628 & 0.8313313704   \\ \cline{2-7} 
                           & \frameworkucbs{} & \textbf{29} & 599 & 299 & 0.95         & 0.8313313704   \\ \cline{2-7} 
                           & \frameworkucbv{}          & 25 & 619 & 309 & 0.857375     & 1.163863919    \\ \hline
\end{tabular}%
}
\end{table}
\noindent\textbf{Remark.} Though we perform experiments for Gaussian noise, our analysis is valid for any noise with variance less than $\sigma^2$.
In order to validate the claim, we also ran experiments with Laplace noise of variance $\sigma^2 = 0.05$ for Hartmann6, PCTS achieves an optimal value 3.2942 whereas the second one achieves 3.2562. 


\newpage
\subsection{Optimizing Hyperparameters of Machine Learning Models}
In this section, we show experiments about evaluating the aforementioned algorithms on a 32-core Intel(R) Xeon(R)@2.3 GHz server for hyperparameter tuning of real machine models. We tune hyper-parameters of SVM on News Group dataset, and XGB and Neural Network on MNIST datasets. The runtime of each experiment for those three tuning tasks are 700s, 1700s, and 1800s respectively. The median value of cross-validation accuracy is shown in Figure \ref{fig:additional_experiement_real}. The table \ref{tb:experiment_real} shows the final cross-validation accuracy found by different algorithms on those three tuning tasks. Notice that we invoke an additional simulation for all algorithms at fidelity $z = 1$ to obtain the cost for optimal fidelity and we preclude this initialization time for all algorithms. We set $\nu_{\max}=1.0$, $\rho_{\max}=0.95$ for MFPOO and \framework{} algorithms. We set $\sigma^2=0.02$ for UCB1-$\sigma$ used in \frameworkucbs{} and MFPOO-UCB1. For \frameworkucbv{} and MFPOO-UCBV, we use $b=1$ as that is the maximum cross-validation accuracy achievable by any classification algorithm.

\subsubsection{Description of Datasets and Models}

\paragraph{News data on SVM.} We train SVM classifier using different hyper-parameter tuning algorithms in NewsGroup dataset ~\citep{newsweeder} The hyper-parameters to tune are the regularization term $C$, ranging from $[e^{-5}, e^{5}]$, and the kernel temperature $\gamma$ from the range $[e^{-5}, e^5]$. Both are accessed in log scale. We set the delay $\tau$ to four seconds and the fidelity range $\mathcal{Z}=[0, 1]$ is mapped to $[100,7000]$. The fidelity range represents the number of samples used to train the SVM classifier with the chosen parameters. The number of jobs specified for sklearn \citep{sklearn} is one. The 5-fold cross-validation accuracy is shown in Figure~\ref{fig:additional_experiement_real}(a). 

\paragraph{MNIST on XGB.}
We tune hyperparameters of XGBOOST~\citep{xgboost} on the MNIST dataset~\citep{mnist}, where the hyperparameters are: 
\begin{enumerate*}[label=(\roman*)]
    \item \texttt{max\_depth} in $[2, 13]$,
    \item \texttt{n\_estimators} in $[10, 400]$,
    \item \texttt{colsample\_bytree} in $[0.2, 0.9]$,
    \item \texttt{gamma} in $[0, 0.7]$, and
    \item  \texttt{learning\_rate} ranging from $[0.05, 0.3]$. 
\end{enumerate*} 
The delay $\tau$ is set to ten seconds and the fidelity range $\mathcal{Z} = [0,1]$ is mapped to the training sample range $[500,20000]$.  The number of jobs specified for sklearn \citep{sklearn} is three. The 3-fold cross-validation accuracy is shown in Figure~\ref{fig:experiement}(b). 

\paragraph{MNIST on Deep Neural Network.}
We also apply the algorithms for tuning the hyper-parameters of a three layer multi layer perceptron (MLP) neural network (NN) classifier on the MNIST dataset~\citep{mnist}. Here, the hyper-parameters being tuned are:
\begin{enumerate*}[label=(\roman*)]
    \item number of neurons of the first, second, and third layers, which belong to the ranges $[32, 128]$, $[128, 256]$, and $[256, 512]$, respectively,
    \item initial learning rate of optimizer in $[e^{-1}, e^{-5}]$ (accessed in log-scale),
    \item optimizers from (`lbfgs', `sgd', `adam'),
    \item activation function from (`tanh', `relu', `logistic'),
    \item early\_stopping from (`enable', `disable').
\end{enumerate*}
The delay $\tau$ for this experiment is 20 seconds. The number of training samples corresponding to the fidelities $z=0$ and $1$ are 1000 and 60000 respectively. The number of jobs specified for sklearn~\citep{sklearn} is ten. The 3-fold cross-validation accuracy is shown in Figure~\ref{fig:experiement}(c).

\subsubsection{Hyper-parameters found by \frameworkucbv{}}

SVM on NewsGroup: \{\texttt{kernel}: \texttt{rbf}, \texttt{C}: 5623.413251903499, \texttt{gamma}: 0.003162277660168379\}

XGB on MNIST: \{\texttt{n\_estimators}: 302, \texttt{learning\_rate}: 0.2375, \texttt{colsample\_bytree}: 0.55, \texttt{gamma}: 0.35, \texttt{hidden\_layer\_sizes}: 4\}

NN on MNIST:\{\texttt{solver}: \texttt{adam}, \texttt{learning\_rate\_init}: 0.001, \texttt{learning\_rate}: \texttt{invscaling}, \texttt{hidden\_layer\_sizes}: (104, 192, 320), \texttt{early\_stopping}: False, \texttt{activation}: \texttt{relu}\}

\subsubsection{Hyper-parameters found by \frameworkucbs{}}

SVM on NewsGroup: \{\texttt{kernel}: \texttt{rbf}, \texttt{C}: 316.2277660168377, \texttt{gamma}: 0.003162277660168379\}

XGB on MNIST: \{\texttt{n\_estimators}: 302, \texttt{learning\_rate}: 0.175, \texttt{colsample\_bytree}: 0.55, \texttt{gamma}: 0.35, \texttt{hidden\_layer\_sizes}: 10\}

NN on MNIST: \{\texttt{solver}: \texttt{adam}, \texttt{learning\_rate\_init}: 0.001, \texttt{learning\_rate}: \texttt{invscaling}, \texttt{hidden\_layer\_sizes}: (104, 192, 448), \texttt{early\_stopping}: False, \texttt{activation}: \texttt{relu}\}

\subsubsection{Statistics of optimal values achieved by different optimizers}
\begin{figure}[h!]
    \centering
    \includegraphics[clip, width=\textwidth]{figures/legend.pdf}
    \includegraphics[clip, width=\textwidth]{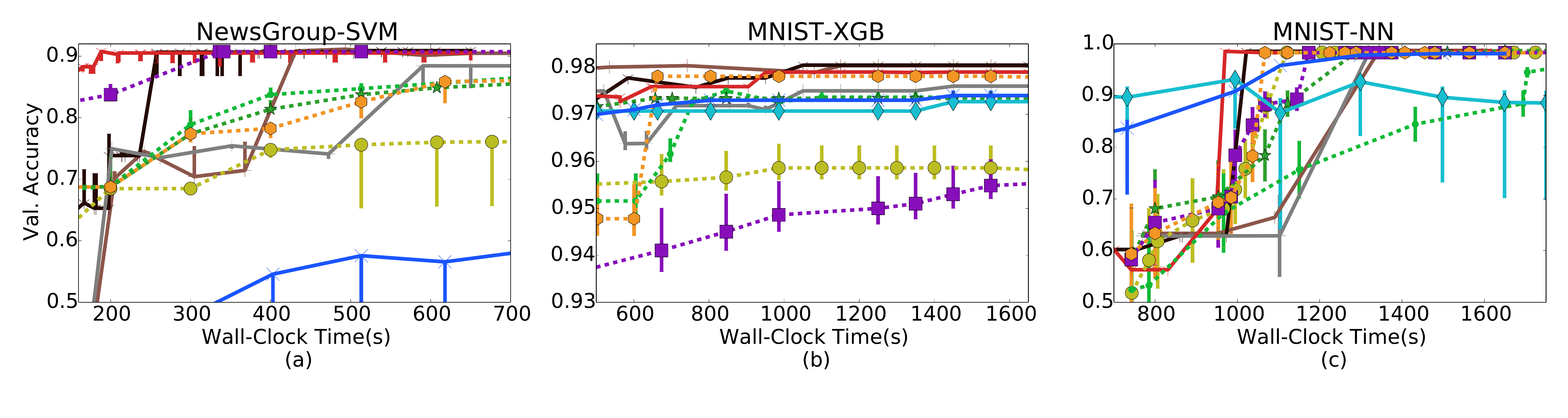}
    \vspace*{-1.6em}
    \caption{Figures (a) to (c) show the cross-validation accuracy (median of 5 runs) achieved on the hyperparameter tuning of classifiers on datasets with DNF feedbacks and the corresponding error bars  $(y_{\text{median}} - y_{\min}, y_{\max} - y_{\text{median}})$ are plotted.}\label{fig:additional_experiement_real}
    \vspace*{-1em}
\end{figure}

The median value of 3-fold cross-validation accuracy over five runs are shown in Figure~\ref{fig:additional_experiement_real} with the error bar. The error bar indicates the spread of the maximum and minimum values obtained over the ten runs.
The statistics of 3-fold cross-validation accuracy of three different classifiers tested on different real-world datasets after 700s, 1700s, and 1800s and over five runs are shown in Table~\ref{tb:experiment_real}. In all the cases, we observe that either \frameworkucbs{} or \frameworkucbv{} outperforms the competing optimization algorithms.

\begin{table}[p]
\centering
\caption{Maximum, median, and standard deviation of final cross-validation accuracy over 5 runs as achieved by different algorithms on different machine learning models tested on real datasets.}
\label{tb:experiment_real}
\begin{tabular}{|l|l|l|l|l|}
\hline
Tuning Task & Algorithms          & Max value    & Median value & Std      \\ \hline
\multirow{11}{*}{NewsGroup-SVM} & MFPOO+UCB1-$\sigma$ & 0.888050296  & 0.8842852451 & 6.76E-02 \\ \cline{2-5} 
 & MFPOO+UCBV          & 0.9047538     & 0.9045731818  & 7.44E-03 \\ \cline{2-5} 
 & \frameworkucbs{} & 0.9092364617  & 0.9081418236  & 2.37E-02 \\ \cline{2-5} 
 & \frameworkucbv{}          & \textbf{0.910979}      & \textbf{0.910269}      & 7.94E-03 \\ \cline{2-5} 
 & GP-UCB              & 0.8961291054  & 0.8821465296  & 1.75E-02 \\ \cline{2-5} 
 & MF-GP-UCB           & 0.7609988  & 0.76098258    & 2.09E-01 \\ \cline{2-5} 
 & BOCA                & 0.9034514049  & 0.9030594374  & \textbf{6.53E-04} \\ \cline{2-5} 
 & GP-EI               & 0.8774512931  & 0.871705938   & 1.15E-02 \\ \cline{2-5} 
 & MF-SKO              & 0.8792711233  & 0.858478645   & 6.93E-02 \\ \cline{2-5} 
 & OGD                 & 0.06817823396 & 0.06542826139 & 3.93E-03 \\ \cline{2-5} 
 & DBGD                & 0.565711969  & 0.565711969   & 2.53E-01 \\ \hline
\multirow{11}{*}{MNIST-XGB}     & MFPOO+UCB1-$\sigma$ & 0.9783142575 & 0.9758995517 & 3.45E-03 \\ \cline{2-5} 
 & MFPOO+UCBV          & 0.98021339    & 0.9797497668  & 7.17E-03 \\ \cline{2-5} 
 & \frameworkucbs{} & \textbf{0.9819297976}  & 0.9791489477  & 3.97E-03 \\ \cline{2-5} 
 & \frameworkucbv{}          & 0.981874905   & \textbf{0.9804613962}  & 2.36E-03 \\ \cline{2-5} 
 & GP-UCB              & 0.9748646302  & 0.9736495155  & 2.03E-03 \\ \cline{2-5} 
 & MF-GP-UCB           & 0.9583507463  & 0.9576005617  & 9.50E-03 \\ \cline{2-5} 
 & BOCA                & 0.9577324826  & 0.954150664   & 1.12E-02 \\ \cline{2-5} 
 & GP-EI               & 0.9775881491  & 0.9767505065  & 1.68E-03 \\ \cline{2-5} 
 & MF-SKO              & 0.9783595013  & 0.9780995824  & 6.50E-04 \\ \cline{2-5} 
 & OGD                 & 0.9729382575  & 0.972700653   & \textbf{4.75E-04} \\ \cline{2-5} 
 & DBGD                & 0.9787723997  & 0.9781500036  & 2.07E-03 \\ \hline
\multirow{11}{*}{MNIST-NN}      & MFPOO+UCB1-$\sigma$ & 0.9845752279 & 0.9845167421 & 2.92E-02 \\ \cline{2-5} 
 & MFPOO+UCBV          & 0.9841946274  & 0.9840167256  & 2.97E-02 \\ \cline{2-5} 
 & \frameworkucbs{} & \textbf{0.9876736106}  & \textbf{0.9872834306}  & 6.50E-02 \\ \cline{2-5} 
 & \frameworkucbv{}          & 0.9868071128  & 0.9865498949  & 4.29E-02 \\ \cline{2-5} 
 & GP-UCB              & 0.9446076809  & 0.9444498028  & 1.97E-02 \\ \cline{2-5} 
 & MF-GP-UCB           & 0.9832337681  & 0.9832332932  & \textbf{1.58E-04} \\ \cline{2-5} 
 & BOCA                & 0.9829225491  & 0.9828833355  & 9.80E-03 \\ \cline{2-5} 
 & GP-EI               & 0.9858101928  & 0.985799993   & 1.27E-03 \\ \cline{2-5} 
 & MF-SKO              & 0.9828856982  & 0.9828833355  & 5.91E-04 \\ \cline{2-5} 
 & OGD                 & 0.9083502708  & 0.8981166245  & 4.67E-01 \\ \cline{2-5} 
 & DBGD                & 0.9813198067  & 0.9810665814  & 4.22E-02 \\ \hline
\end{tabular}%
\end{table}

\subsubsection{Key statistics of the trees constructed by MFPOO and \framework{}}
The detail comparisons between MFPOO and \framework{} are shown in Table~\ref{tb:tree_real}. In general, the depth of the tree created using \framework{} algorithm is larger than the depth of the tree created using MFPOO algorithm for both UCB1-$\sigma$ and UCBV policy.

\begin{table}[]
\centering
\caption{Statistics of the trees constructed by MFPOO and \framework{} based approaches for hyper-parameter tuning of real machine learning models.}~\label{tb:tree_real}
\resizebox{\textwidth}{!}{%
\begin{tabular}{|l|l|r|r|r|r|r|}
\hline
Tuning Task &
  Tree Search Algorithms &
  \multicolumn{1}{l|}{Tree Height} &
  \multicolumn{1}{l|}{Number of Tree Nodes} &
  \multicolumn{1}{l|}{Number of Iterations (T)} &
  \multicolumn{1}{l|}{Best $\rho$} &
  \multicolumn{1}{l|}{Best $\nu_1$}\\ \hline
\multirow{4}{*}{NewsGroup-SVM} & MFPOO+UCB1-$\sigma$ & 11 & 21 & 10 & 9.50E-01 & 5.49E-01 \\ \cline{2-7} 
                               & MFPOO+UCBV          & 7  & 35 & 17 & 9.50E-01 & 5.49E-01 \\ \cline{2-7} 
                               & \frameworkucbs{}  & \textbf{21} & 81 & 40 & 9.50E-01 & 4.49E-01 \\ \cline{2-7} 
                               & \frameworkucbv{} & 16 & 81 & 40 & 9.50E-01 & 3.88E-01 \\ \hline
\multirow{4}{*}{MNIST-XGB}     & MFPOO+UCB1-$\sigma$ & 4  & 9  & 4  & 8.15E-01 & 1.25E-01 \\ \cline{2-7} 
                               & MFPOO+UCBV          & 4  & 9  & 4  & 8.57E-01 & 1.17E-01 \\ \cline{2-7} 
                               & \frameworkucbs{}  & 8  & 17 & 8  & 9.03E-01 & 1.10E-01 \\ \cline{2-7} 
                               & \frameworkucbv{} & \textbf{9}  & 17 & 8  & 9.03E-01 & 1.09E-01 \\ \hline
\multirow{4}{*}{MNIST-NN}      & MFPOO+UCB1-$\sigma$ & 5  & 9  & 4  & 9.03E-01 & 6.90E-01 \\ \cline{2-7} 
                               & MFPOO+UCBV          & 5  & 11 & 5  & 9.03E-01 & 6.39E-01 \\ \cline{2-7} 
                               & \frameworkucbs{}  & \textbf{7}  & 15 & 7  & 9.50E-01 & 8.23E-01 \\ \cline{2-7} 
                               & \frameworkucbv{} & \textbf{7}  & 15 & 7  & 9.50E-01 & 7.01E-01 \\ \hline
\end{tabular}%
}
\end{table}

\newpage
\subsection{Experiments with Stochastic Delays}
In this section, we show experiment results for stochastic delays instead of constant delays. Same as previous experiments, We run each experiment ten times for 600s on a MacBook Pro with a 6-core Intel(R) Xeon(R)@2.60GHz CPU. The noise variance for all synthetic functions remains same as before and the initialization value of $\nu_{\max}$ and $\rho_{\max}$ for MFPOO and \framework{} also keeps same. 

\noindent\textbf{Generating Stochastic Delays.} Delays are generated using a geometric distribution $\tau \sim$ Geo$(1 / \mean{\tau})$, and the expectation of the delay time $\mean{\tau}$ for all synthetic functions is set to ten seconds. For generating stochastic delays using the geometric distribution, we use the same tricks and motivation as in~\citep[Section 7]{vernade2017stochastic}.

\subsubsection{Statistics of Optimal Values Achieved by Different Delay Tolerant Optimizers}
\begin{figure}[h!]
    \centering
    \includegraphics[clip, width=0.9\textwidth]{figures/legend.pdf}
    \includegraphics[clip, width=0.9\textwidth]{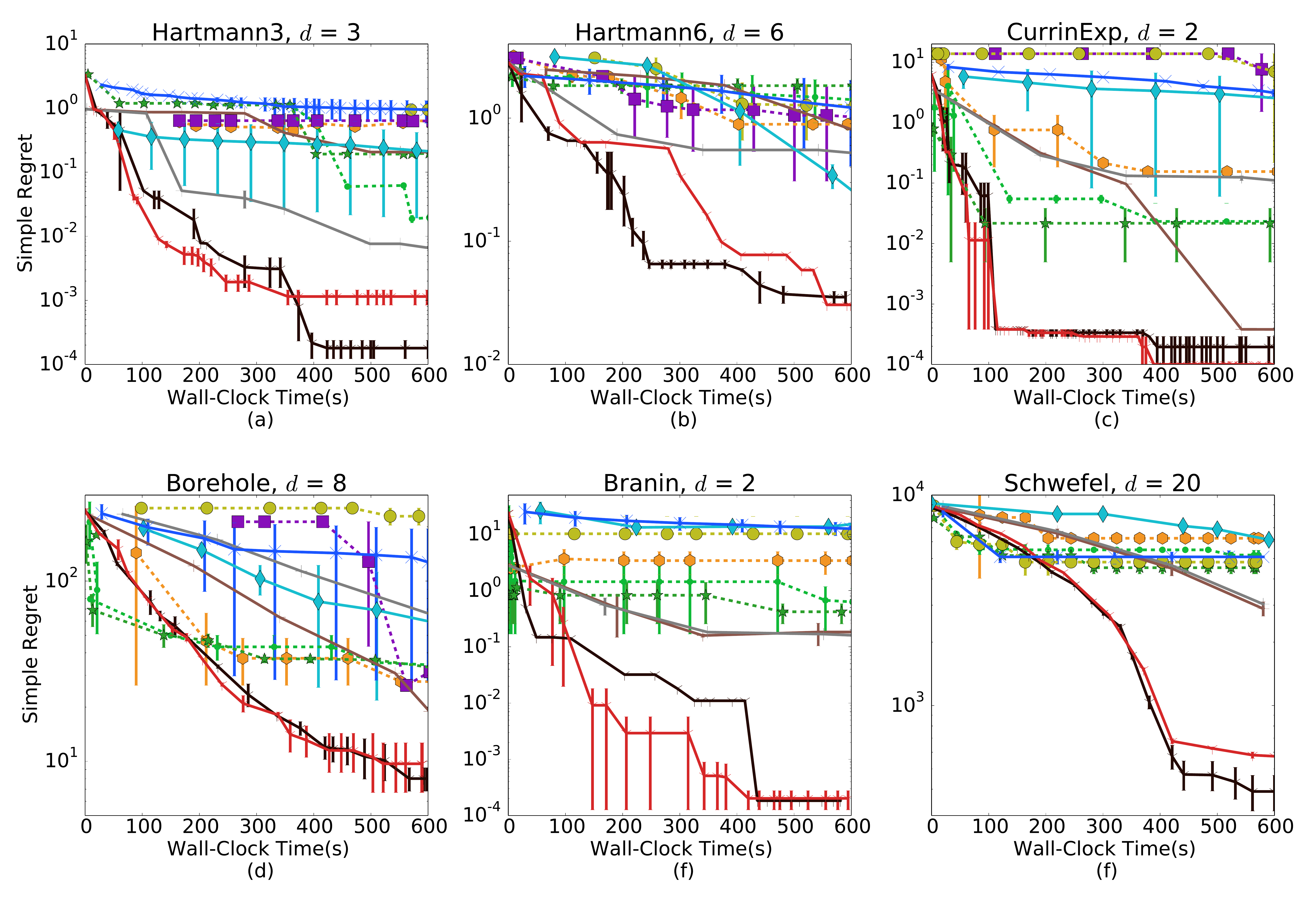}
    \vspace*{-1.6em}
    \caption{Figures (a) to (f) show simple regret (median of 10 runs) of different algorithms on synthetic functions under stochastic delay $\tau \sim$ Geo$(0.1)$ and the corresponding error bars  $(y_{\text{median}} - y_{\min}, y_{\max} - y_{\text{median}})$ are plotted.}\label{fig:additional_experiement_stochastic}
    \vspace*{-1em}
\end{figure}

The median value of simple regret over ten runs are shown in Figure~\ref{fig:additional_experiement_stochastic} with the error bar. The error bar indicates the spread of the maximum and minimum values obtained over the ten runs. The statistics of the optimal values of different algorithms tested on different synthetic functions for 600s are shown in Table~\ref{tb:experiment_stochastic}. In all the cases, we observe that either \frameworkucbs{} or \frameworkucbv{} outperforms the competing optimization algorithms.

\noindent\textbf{Improvement of \framework{} in constant and stochastic delays.} Under constant delay, for Hartmann6 and Branin, PCTS achieve the optimal values as 3.305830186 and -0.3988127406 respectively, while the second best achieves the optimal value of $f$ as 3.287516333 and -0.415716 respectively. Under stochastic delay, for Hartmann6 and Branin, PCTS has the optimal value of $f$ as 3.291825 and -0.398084 respectively, while the second best has the optimal value of $f$ as 3.147713 and $-0.554698$ respectively. \textit{We observe that the performance improvement led by \framework{} variants is even higher for stochastic delays than constant delays.} Similar conclusions can be drawn from other results.

\begin{table}[p]
\centering
\caption{Maximum, Median, and Standard Deviation of optimal values over $10$ runs of different algorithms for different synthetic functions under stochastic delay $\tau \sim$ Geo$(0.1)$.}~\label{tb:experiment_stochastic}
\vspace*{1em}
\begin{tabular}{|l|l|r|r|r|}
\hline
Synthetic functions & Algorithms & \multicolumn{1}{l|}{Max value} & \multicolumn{1}{l|}{Median value} & \multicolumn{1}{l|}{Std} \\ \hline
\multirow{11}{*}{\begin{tabular}[c]{@{}l@{}}Hartmann3 \\ \\ (optimal value = 3.86278)\end{tabular}} & MFPOO+UCB1-$\sigma$ & 3.858158 & 3.856659 & 5.90E-03 \\ \cline{2-5} 
 & MFPOO+UCBV & 3.656709 & 3.656619 & \textbf{2.99E-04} \\ \cline{2-5} 
 & \frameworkucbs{} & 3.862543 & 3.861633 & 9.55E-04 \\ \cline{2-5} 
 & \frameworkucbv{} & \textbf{3.862658} & \textbf{3.8626} & 1.03E-03 \\ \cline{2-5} 
 & GP-UCB & 3.846287 & 3.846113 & 6.28E-04 \\ \cline{2-5} 
 & MF-GP-UCB & 2.927849 & 2.925782 & 2.09E-03 \\ \cline{2-5} 
 & BOCA & 3.229559 & 3.229192 & 7.74E-04 \\ \cline{2-5} 
 & GP-EI & 3.672403 & 3.66852 & 4.81E-03 \\ \cline{2-5} 
 & MF-SKO & 3.229922 & 3.229182 & 1.24E-03 \\ \cline{2-5} 
 & OGD & 3.844178 & 3.660379 & 1.84E-01 \\ \cline{2-5} 
 & DBGD & 3.281405 & 2.922297 & 3.60E-01 \\ \hline
\multirow{11}{*}{\begin{tabular}[c]{@{}l@{}}Hartmann6\\ \\ (optimal value = 3.32237)\end{tabular}} & MFPOO+UCB1-$\sigma$ & 2.801362 & 2.801252 & 9.92E-04 \\ \cline{2-5} 
 & MFPOO+UCBV & 2.570945 & 2.570896 & \textbf{9.95E-05} \\ \cline{2-5} 
 & \frameworkucbs{} & \textbf{3.291905} & \textbf{3.291825} & 5.41E-04 \\ \cline{2-5} 
 & \frameworkucbv{} & 3.29112 & 3.287154 & 4.01E-03 \\ \cline{2-5} 
 & GP-UCB & 2.551277 & 1.91283 & 6.39E-01 \\ \cline{2-5} 
 & MF-GP-UCB & 2.661155 & 2.051153 & 6.11E-01 \\ \cline{2-5} 
 & BOCA & 3.014768 & 2.312637 & 7.03E-01 \\ \cline{2-5} 
 & GP-EI & 1.747662 & 1.488307 & 2.60E-01 \\ \cline{2-5} 
 & MF-SKO & 2.524356 & 2.430526 & 9.41E-02 \\ \cline{2-5} 
 & OGD & 3.251979 & 3.147713 & 1.05E-01 \\ \cline{2-5} 
 & DBGD & 2.914959 & 2.104141 & 8.11E-01 \\ \hline
\multirow{11}{*}{\begin{tabular}[c]{@{}l@{}}CurrinExp\\ \\ (optimal value = 13.798685)\end{tabular}} & MFPOO+UCB1-$\sigma$ & 13.688384 & 13.688384 & 5.70E-04 \\ \cline{2-5} 
 & MFPOO+UCBV & 13.798306 & 13.798306 & 4.95E-04 \\ \cline{2-5} 
 & \frameworkucbs{} & \textbf{13.798585} & \textbf{13.798491} & 8.82E-04 \\ \cline{2-5} 
 & \frameworkucbv{} & \textbf{13.798585} & 13.798398 & \textbf{3.12E-04} \\ \cline{2-5} 
 & GP-UCB & 13.78547 & 13.77547 & 1.03E-02 \\ \cline{2-5} 
 & MF-GP-UCB & 13.580857 & 6.790429 & 7.43E+00 \\ \cline{2-5} 
 & BOCA & 12.450548 & 6.225274 & 7.22E+00 \\ \cline{2-5} 
 & GP-EI & 13.793733 & 13.77704 & 1.73E-02 \\ \cline{2-5} 
 & MF-SKO & 13.671398 & 13.643552 & 2.83E-02 \\ \cline{2-5} 
 & OGD & 13.738852 & 11.310591 & 3.34E+00 \\ \cline{2-5} 
 & DBGD & 11.091881 & 10.673992 & 4.21E-01 \\ \hline
\multirow{11}{*}{\begin{tabular}[c]{@{}l@{}}Borehole\\ \\ (optimal value = 309.523221)\end{tabular}} & MFPOO+UCB1-$\sigma$ & 276.52646 & 246.52646 & 3.02E+01 \\ \cline{2-5}
 & MFPOO+UCBV & 293.597767 & 290.597767 & 3.42E+00 \\ \cline{2-5} 
 & \frameworkucbs{} & \textbf{302.808514} & 299.836555 & 3.52E+00 \\ \cline{2-5} 
 & \frameworkucbv{} & 302.694521 & \textbf{301.506202} & \textbf{1.80E+00} \\ \cline{2-5} 
 & GP-UCB & 277.294312 & 275.288875 & 3.00E+00 \\ \cline{2-5} 
 & MF-GP-UCB & 105.501624 & 80.501624 & 3.38E+01 \\ \cline{2-5} 
 & BOCA & 283.150628 & 278.150628 & 1.35E+01 \\ \cline{2-5} 
 & GP-EI & 281.252433 & 276.426655 & 8.53E+00 \\ \cline{2-5} 
 & MF-SKO & 283.150628 & 281.830216 & 2.25E+00 \\ \cline{2-5} 
 & OGD & 297.752136 & 250.768674 & 5.65E+01 \\ \cline{2-5} 
 & DBGD & 281.503929 & 182.224534 & 1.08E+02 \\ \hline
\multirow{11}{*}{\begin{tabular}[c]{@{}l@{}}Branin\\ \\ (optimal value =  -0.3979)\end{tabular}} & MFPOO+UCB1-$\sigma$ & -0.50178 & -0.5814 & 8.04E-02 \\ \cline{2-5} 
 & MFPOO+UCBV & -0.554698 & -0.554698 & 9.84E-04 \\ \cline{2-5} 
 & \frameworkucbs{} & \textbf{-0.398027} & -0.398102 & 1.03E-03 \\ \cline{2-5} 
 & \frameworkucbv{} & -0.398084 & \textbf{-0.398084} & \textbf{1.34E-04} \\ \cline{2-5} 
 & GP-UCB & -0.566961 & -1.028222 & 4.61E-01 \\ \cline{2-5} 
 & MF-GP-UCB & -9.96478 & -10.621964 & 6.58E-01 \\ \cline{2-5} 
 & BOCA & -19.24392808 & -20.621964 & 1.04E+01 \\ \cline{2-5} 
 & GP-EI & -0.654828 & -0.819048 & 1.65E-01 \\ \cline{2-5} 
 & MF-SKO & -2.315798 & -3.820579 & 2.62E+00 \\ \cline{2-5} 
 & OGD & -15.050285 & -16.180975 & 6.92E+00 \\ \cline{2-5} 
 & DBGD & -9.56788 & -13.007573 & 4.02E+00 \\ \hline
\end{tabular}%
\end{table}

\subsubsection{Key statistics of the trees constructed by MFPOO and \framework{}}
The detail comparisons between MFPOO and \framework{} under the  stochastic delay $\tau \sim$ Geo$(0.1)$ are shown in Table~\ref{tb:tree_statistics_stochastic}. In general, the depth of the tree created using \framework{} algorithm is larger than the depth of the tree created using MFPOO algorithm for both UCB1-$\sigma$ and UCBV policy.

\begin{table}[t!]
\centering
\caption{Statistics of the trees constructed by MFPOO and \framework{} based approaches for optimizing different synthetic functions under the stochastic delay $\tau \sim$ Geo$(0.1)$.}~\label{tb:tree_statistics_stochastic}
\resizebox{\textwidth}{!}{%
\begin{tabular}{|l|l|r|r|l|r|r|}
\hline
  Synthetic functions &
  Tree Search Algorithms &
  \multicolumn{1}{l|}{Tree Height} &
  \multicolumn{1}{l|}{Number of Tree Nodes} &
  \multicolumn{1}{l|}{Number of Iterations (T)} &
  \multicolumn{1}{l|}{Best $\rho$} &
  \multicolumn{1}{l|}{Best $\nu_1$}\\ \hline
\multirow{4}{*}{Hartmann3} & MFPOO+UCB1-$\sigma$ & 17 & 67 & 33 & 0.9259454628 & 0.006103201358 \\ \cline{2-7} 
 & MFPOO+UCBV & 12 & 67 & 33 & 0.95 & 0.006103201358 \\ \cline{2-7} 
 & PCTS+DUCB1-$\sigma$ & 26 & 407 & 203 & 0.95 & 0.006103201358 \\ \cline{2-7} 
 & PCTS+DUCBV & \textbf{28} & 339 & 169 & 0.95 & 0.006103201358 \\ \hline
\multirow{4}{*}{Hartmann6} & MFPOO+UCB1-$\sigma$ & 16 & 69 & 34 & 0.9259454628 & 0.00842254613 \\ \cline{2-7} 
 & MFPOO+UCBV & 9 & 63 & 31 & 0.857375 & 0.01179156458 \\ \cline{2-7} 
 & PCTS+DUCB1-$\sigma$ & 30 & 323 & 161 & 0.9259454628 & 0.01179156458 \\ \cline{2-7} 
 & PCTS+DUCBV & \textbf{33} & 327 & 163 & 0.95 & 0.00842254613 \\ \hline
\multirow{4}{*}{CurrinExp} & MFPOO+UCB1-$\sigma$ & 15 & 69 & 34 & 0.857375 & 0.6094708386 \\ \cline{2-7} 
 & MFPOO+UCBV & 10 & 69 & 34 & 0.9259454628 & 0.6094708386 \\ \cline{2-7} 
 & PCTS+DUCB1-$\sigma$ & 19 & 407 & 203 & 0.95 & 0.6094708386 \\ \cline{2-7} 
 & PCTS+DUCBV & 21 & 385 & 192 & 0.857375 & 0.853259174 \\ \hline
\multirow{4}{*}{Borehole} & MFPOO+UCB1-$\sigma$ & 21 & 63 & 31 & 0.857375 & 28.65744081 \\ \cline{2-7} 
 & MFPOO+UCBV & 34 & 69 & 34 & 0.9259454628 & 20.46960058 \\ \cline{2-7} 
 & PCTS+DUCB1-$\sigma$ & 53 & 399 & 199 & 0.857375 & 28.65744081 \\ \cline{2-7} 
 & PCTS+DUCBV & \textbf{73} & 321 & 160 & 0.9259454628 & 20.46960058 \\ \hline
\multirow{4}{*}{Branin} & MFPOO+UCB1-$\sigma$ & 18 & 67 & 33 & 0.95 & 0.8313313704 \\ \cline{2-7} 
 & MFPOO+UCBV & 10 & 69 & 34 & 0.9259454628 & 0.8313313704 \\ \cline{2-7} 
 & PCTS+DUCB1-$\sigma$ & \textbf{23} & 389 & 194 & 0.9259454628 & 0.8313313704 \\ \cline{2-7} 
 & PCTS+DUCBV & 22 & 385 & 192 & 0.9259454628 & 0.8313313704 \\ \hline
\end{tabular}
}\vspace*{-1em}
\end{table}

\subsection{Experimental comparison with Successive Rejections}
\begin{figure}[h!]
    \centering
    \includegraphics[clip, width=0.3\textwidth]{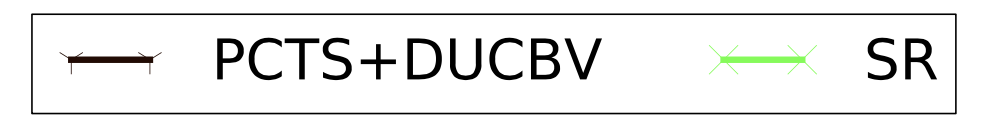}
    \includegraphics[clip, width=0.8\textwidth]{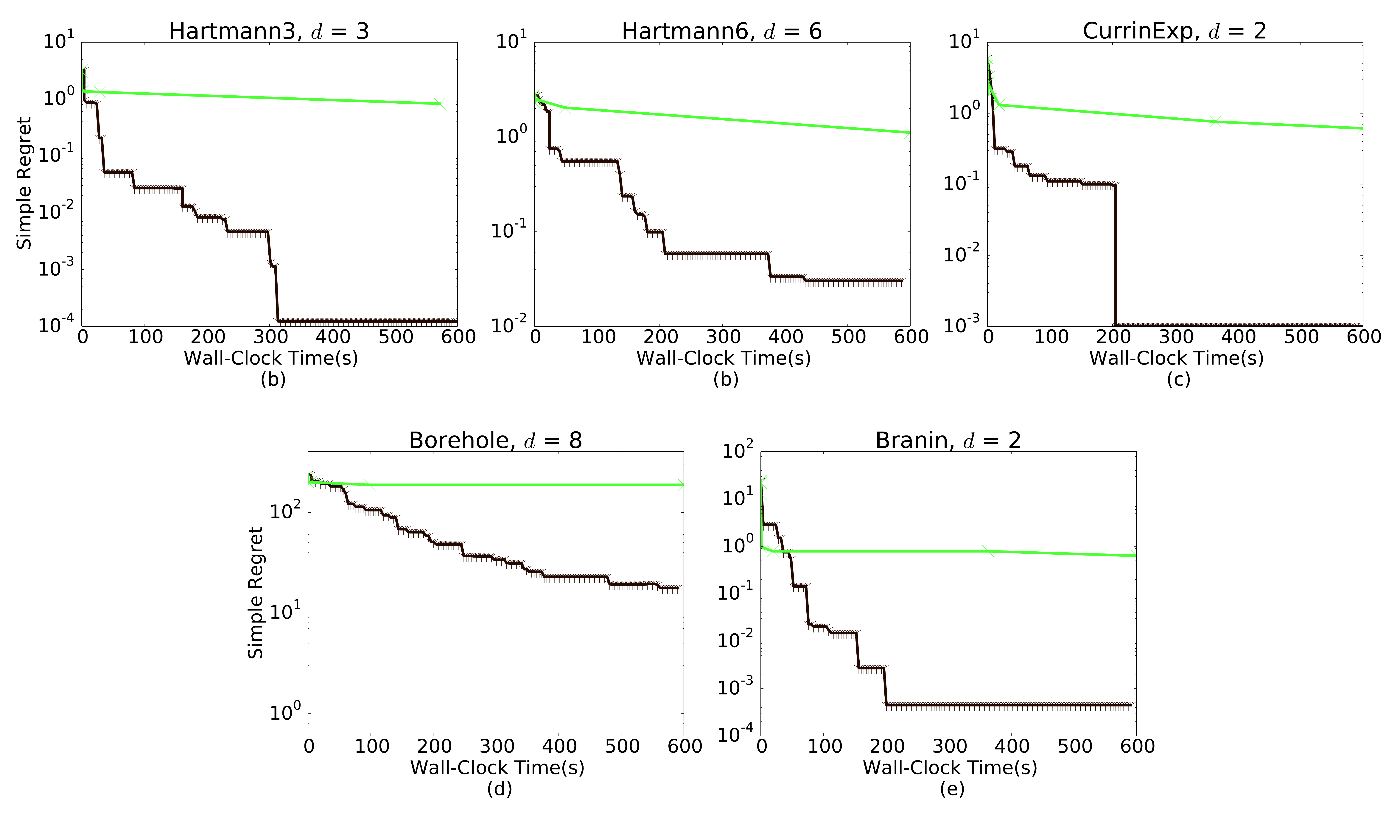}
    \vspace*{-2em}
    \caption{Figures (a) to (f) show simple regret (median of 5 runs) of \frameworkucbv{} and SR (successive rejection) on synthetic functions with delay feedbacks. We obtained almost same results for both \frameworkucbv{} and SR among those 5 experiments.
    }\label{fig:additional_sr_experiement}
    \vspace*{-1em}
\end{figure}


In order to compare with the wait-and-act version of the Successive Reject (SR) algorithm~\cite[Algorithm 2]{locatelli2018} with \frameworkucbv{}, we run experiments on five synthetic functions described in Appendix D.2. As the SR algorithm does not support multi-fidelity feedback, we set the feedbacks to have perfect fidelity. For comparison, also we set the delay $\tau$ to a constant, i.e. 4. The experimental results are shown in Figure~\ref{fig:additional_sr_experiement}.
We observe for all the synthetic functions the simple regret of \frameworkucbv{} is at least 10X less than that of the modified SR algorithm handling constant delay.
This shows that the modified SR algorithm enabled to handle constant delay does not yield lower simple regret than the proposed algorithm \frameworkucbv{}.

\end{document}